\documentclass[twoside,leqno]{article} 
\usepackage[accepted]{aistats2017}
\usepackage{natbib} 
\usepackage{amssymb} 
\usepackage{amsmath} 

\usepackage{graphicx} 

\usepackage{caption} 
\usepackage{subcaption} 

\usepackage{booktabs} 

\newcommand{\R}{\boldsymbol R}
\newcommand{\C}{\boldsymbol C}
\newcommand{\E}{\boldsymbol E}
\newcommand{\F}{\boldsymbol F}
\renewcommand{\S}{\boldsymbol S}
\newcommand{\G}{\boldsymbol G}
\newcommand{\D}{\boldsymbol D}
\newcommand{\btheta}{\boldsymbol \theta}

\newcommand{\lambdaF}{\lambda_F}
\newcommand{\lambdaS}{\lambda_S}
\newcommand{\lambdaG}{\lambda_G}

\newcommand{\muFik}{\mu_{ik}^F}

\newcommand{\muGjl}{\mu_{jl}^G}
\newcommand{\muGjk}{\mu_{jk}^G}

\newcommand{\tauFik}{\tau_{ik}^F}

\newcommand{\tauGjk}{\tau_{jk}^G}

\begin{document}

%

%

\twocolumn[

\aistatstitle{Bayesian Hybrid Matrix Factorisation for Data Integration}

\aistatsauthor{ Thomas Brouwer \And Pietro Li\'{o} }
\aistatsaddress{ University of Cambridge \And University of Cambridge } 
]

\begin{abstract}
	We introduce a novel Bayesian hybrid matrix factorisation model (HMF) for data integration, based on combining multiple matrix factorisation methods, that can be used for in- and out-of-matrix prediction of missing values. The model is very general and can be used to integrate many datasets across different entity types, including repeated experiments, similarity matrices, and very sparse datasets. 
	We apply our method on two biological applications, and extensively compare it to state-of-the-art machine learning and matrix factorisation models.
	For in-matrix predictions on drug sensitivity datasets we obtain consistently better performances than existing methods. This is especially the case when we increase the sparsity of the datasets.
	Furthermore, we perform out-of-matrix predictions on methylation and gene expression datasets, and obtain the best results on two of the three datasets, especially when the predictivity of datasets is high.
\end{abstract}

\section{INTRODUCTION}
\begin{figure*}[t]
		\centering
		\includegraphics[height=65pt]{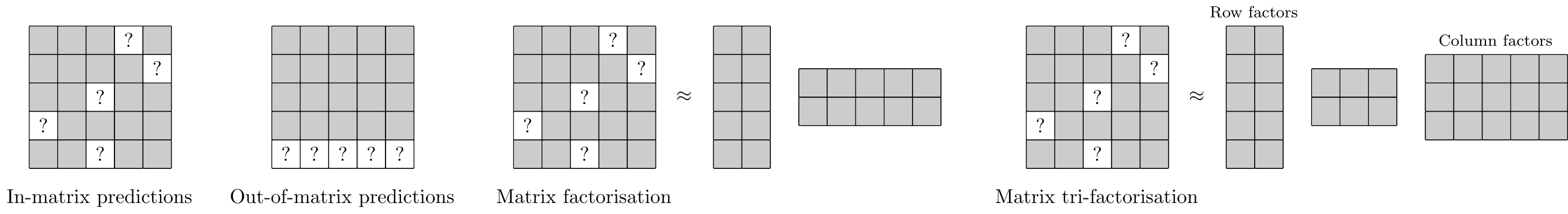}
		\captionsetup{width=2\columnwidth}
		\caption{Difference between in- and out-of-matrix predictions for missing values in matrices; and the difference between matrix factorisation and matrix tri-factorisation.}
		\label{in_out_mf_mtf}
	\end{figure*}

	Matrix factorisation methods offer an elegant way of analysing datasets. Here, a matrix relating two entity types is decomposed into two smaller matrices (so-called latent factors) so that their product approximates the original one. This extracts hidden structure in the data, and allows the prediction of missing values. Non-negativity constraints are often imposed on the matrices (\cite{Lee1999}) as this makes the results easier to interpret, and it is often inherent to the problem -- such as in image processing (\cite{Lee1999}) or bioinformatics (\cite{Brunet2004}). Non-negative matrix tri-factorisation is an extension of these methods, first introduced by \cite{Ding2006a}, where the matrix is decomposed into three smaller matrices, which again are constrained to be non-negative. 
	Both methods are shown in Figure \ref{in_out_mf_mtf}.
	
	A key question is how to best predict missing values in these datasets. 
	There are two different settings for this problem. Firstly, \textbf{in-matrix predictions}, where 
	if we are trying to predict an unknown value for a pair of drug D1 and cancer type C1, we will have at least one known value for D1 with another cancer type C2, and for C1 with another drug D1.
	The other setting is \textbf{out-of-matrix predictions}, where we predict values for entirely unseen rows or columns, such as a new drug for which we have no observed values inside the matrix. 
	This is illustrated in Figure \ref{in_out_mf_mtf}.
	
	
\begin{figure}[t]
		\centering
		\resizebox{0.475\textwidth}{!}{
			\includegraphics{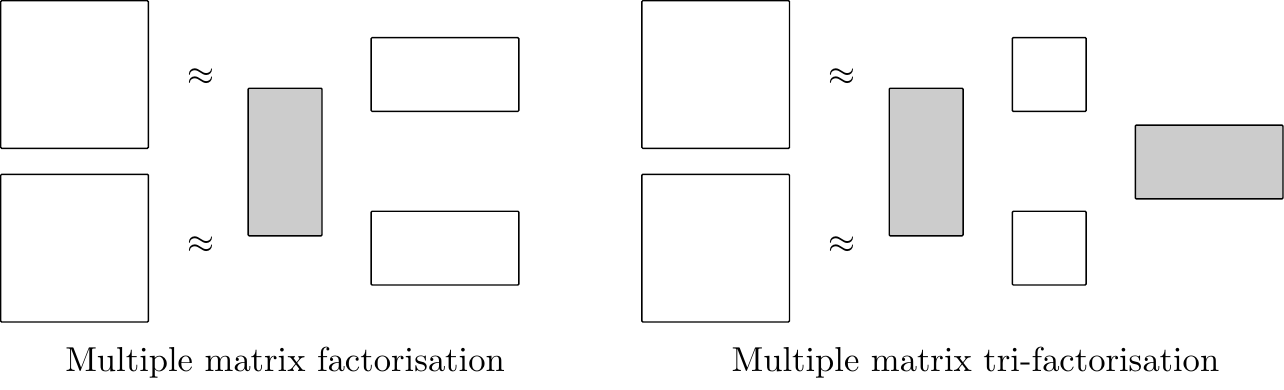}
		}
		\captionsetup{width=0.95\columnwidth}
		\caption{Difference between multiple matrix factorisation and multiple matrix tri-factorisation. The shared factor matrices are highlighted in grey.}
		\label{multiple_mf_mtf}
	\end{figure}

	In practice we often have many different datasets, relating different entity types.
	Matrix factorisation methods can be effectively used for data integration, by jointly decomposing multiple datasets and sharing the latent matrices (\cite{Zhang2005}). 
	This can improve our matrix factorisations, and hence our in-matrix predictions, and also allows us to do out-of-matrix predictions. 
	Another approach, based on multiple matrix tri-factorisation, was introduced by \cite{FeiWangTaoLi2008}, where they shared two of the three latent matrices. 
	By sharing more factors than the multiple matrix factorisation method, and hence having a much smaller dataset-specific matrix in the middle, we can more effectively integrate similar datasets. This is particularly interesting for integrating repeated experiments, where different biological labs perform similar experiments between the same two entity types, such as gene expression profiles and methylation levels. 
	Both approaches are illustrated in Figure \ref{multiple_mf_mtf}.
	
	We propose a novel Bayesian model for data integration, which combines multiple matrix factorisation and tri-factorisation. Our method can integrate many datasets across different entity types, including repeated experiments, similarity matrices, and very sparse datasets.
	In our method, the user can specify for each dataset whether it should be decomposed into two matrices, in which case only the row factor matrices are shared, or into three, in which case the row and column matrices are shared. This gives a hybrid between matrix factorisation and tri-factorisation. Additionally, the user can also specify for each of the latent matrices whether the factors should be nonnegative or real-valued, giving a hybrid between nonnegative, semi-nonnegative, and real-valued factorisations.
	By using a probabilistic approach, our method can effectively handle missing values and predict them, both for in- and out-of-matrix predictions, and the Bayesian approach is much less prone to overfitting than non-probabilistic models.
	Furthermore, the rank of each matrix is automatically chosen using Automatic Relevance Determination, eliminating the need to perform model selection.
	Related work is discussed in Section \ref{Related Work}.
	
	To demonstrate our method, we apply it to two different settings. 
	Firstly, we consider four drug sensitivity datasets, where the matrices are similar and hence have high predictivity. We measure the in-matrix predictive performance of our method, as well as Bayesian and non-probabilistic matrix factorisation methods, and several state-of-the-art machine learning methods. Our model consistently outperforms all other methods, especially when the sparsity of the data increases. Secondly, we integrate gene expression, promoter region methylation, and gene body methylation profiles for breast cancer patients. These datasets are much more dissimilar, hence predicting one dataset given the others is much harder. However, in out-of-matrix prediction experiments our method achieves better performance than state-of-the-art machine learning methods on two of the three combinations.
	
\section{MATRIX FACTORISATION} \label{MF}
	The problem of non-negative matrix factorisation (NMF) can be formulated as decomposing a matrix $ \R \in \mathbb{R}^{I \times J} $ into two latent (unobserved) factor matrices $ \F \in \mathbb{R}_+^{I \times K} $, $ \G \in \mathbb{R}_+^{J \times K} $. In other words, solving $ \R = \F \G^T + \E $, where noise is captured by matrix $ \E \in \mathbb{R}^{I \times J} $. Some entries in the dataset $ \R $ may not be known -- we represent the indices of observed entries by the set $ \Omega = \left\{ (i,j) \text{ } \vert \text{ } R_{ij} \text{ observed} \right\} $. Similarly, non-negative matrix tri-factorisation (NMTF) can be formulated as finding three latent factor matrices $ \F \in \mathbb{R}_+^{I \times K} $, $ \S \in \mathbb{R}_+^{K \times L} $, $ \G \in \mathbb{R}_+^{J \times K} $, such that $ \R = \F \S \G^T + \E $.
	
	Some NMF methods such as \citet{Lee2001} rely on optimisation-based techniques, where a cost function between the observed matrix $\R$ and the predicted matrix $\F \G^T$ is minimised, like the mean squared error or $I$-divergence, using multiplicative updates.
	Alternatively, probabilistic models formulate the problem of NMF by treating the entries in $\F, \G$ as unobserved or latent variables, and the entries in $\R$ as observed datapoints. Bayesian approaches furthermore place prior distributions over the latent variables. The problem then involves finding the distribution over $\F, \G$ after observing $\R$, $p(\F, \G | \R)$. This Bayesian approach has several benefits: it is less prone to overfitting, especially on small or sparse datasets; a distribution over the factors is obtained, rather than just a point estimate; it allows for flexible and elegant models (such as automatic model selection using Automatic Relevance Determination); and missing entries are easily handled (we simply do not include them in the observed data, through the $\Omega$ set introduced earlier). However, finding this posterior distribution can be very inefficient.
	
	\citet{Schmidt2009a} introduced a Bayesian model for non-negative matrix factorisation, by using Exponential priors and a Gaussian likelihood. For the precision $ \tau $ of the likelihood they used a Gamma distribution with shape $ \alpha > 0 $ and rate $ \beta > 0 $. The full set of parameters for this model is denoted $ \btheta = \left\{ \F, \G, \tau \right\} $.
	\begin{alignat*}{1}
		& R_{ij} \sim \mathcal{N} (R_{ij} | \boldsymbol F_i \cdot \boldsymbol G_j, \tau^{-1} ) \\
		& F_{ik} \sim \mathcal{E} ( F_{ik} | \lambdaF)		\quad		G_{jk} \sim \mathcal{E} ( G_{jk} | \lambdaG)		\quad		\tau \sim \mathcal{G} (\tau | \alpha, \beta )
	\end{alignat*}
	Inference to find the posterior $p(\F, \G | \R)$ can be efficiently performed using Gibbs sampling. This method works by sampling new values for each parameter $ \theta_i $ from its marginal $ p(\theta_i | \btheta_{-i}, D ) $ given the current values of the other parameters $ \btheta_{-i} $, and the observed data $ D $. If we sample new values in turn for each parameter $ \theta_i $ from $ p(\theta_i | \btheta_{-i}, D ) $, we will eventually converge to draws from the posterior $ p(\btheta|D) $, which can be used to approximate it. When doing so we have to discard the first $n$ draws because it takes a while to converge (\textit{burn-in}), and since consecutive draws are correlated we only use every $i$th value (\textit{thinning}).

	For this model we draw from the following distributions: 
	\begin{alignat*}{1}
		&p(F_{ik}|\tau,\F_{-ik},\G,D)	\quad\quad 	p(G_{jk}|\tau,\F,\G_{-jk},D)		\\
		&p(\tau|\F,\G,D)
	\end{alignat*}
	where $\F_{-ik}$ denotes all elements in $\F$ except $F_{ik}$, and similarly for $\G_{-jl}$. Using Bayes' theorem we obtain the following posterior distributions:
	\begin{alignat*}{1}
		p(\tau|\F,\G,D) &= \mathcal{G} (\tau | \alpha^*, \beta^* ) \\
		p(F_{ik}|\tau,\F_{-ik},\G,D) &= \mathcal{TN} ( F_{ik} | \muFik, \tauFik ) \\
		p(G_{jl}|\tau,\F,\G_{-jk},D) &= \mathcal{TN} ( G_{jk} | \muGjl, \tauGjk ),
	\end{alignat*}
	where
	\begin{equation*}
		\mathcal{TN} ( x | \mu, \tau ) = \left\{
		\begin{array}{ll}
			\displaystyle \frac{ \sqrt{ \frac{\tau}{2\pi} } \exp \left\{ -\frac{\tau}{2} (x - \mu)^2 \right\} }{ 1 - \Phi ( - \mu \sqrt{\tau} )}  & \mbox{if } x \geq 0 \\
			0 & \mbox{if } x < 0
		\end{array}
		\right.
	\end{equation*}
	is a truncated normal: a normal distribution with zero density below $ x = 0 $ and renormalised to integrate to one. $ \Phi(\cdot) $ is the cumulative distribution function of $ \mathcal{N}(0,1) $. 
	
	The extension of this model to non-negative matrix tri-factorisation is straightforward. 
	We can also choose to remove the nonnegativity constraint, by instead using a Gaussian prior for the factor matrices. This results in a Gaussian posterior in the Gibbs sampling algorithm, with slightly different parameters. A semi-nonnegative model, with only one real-valued matrix ($\G$ for MF, and $\S$ for MTF), is illustrated below.
	Gibbs samplers for all mentioned models are given in the Supplementary Materials (Section 1).
	%
	\begin{alignat*}{2}
		& \text{Prior: }		&& \text{Posterior: } \\
		&F_{ik} \sim \mathcal{E} ( F_{ik} | \lambdaF)	&&F_{ik} \sim \mathcal{TN} ( F_{ik} | \muFik, \tauFik ) \\
		&G_{jk} \sim \mathcal{N} ( G_{jk} | 0, \lambdaG^{-1})	\quad	 &&G_{jk} \sim \mathcal{N} ( G_{jk} | \muGjk, (\tauGjk)^{-1} )
	\end{alignat*}

\section{HYBRID MATRIX FACTORISATION} \label{HMF}
\begin{figure}[!t]
		\centering
		\includegraphics[width=0.76\columnwidth]{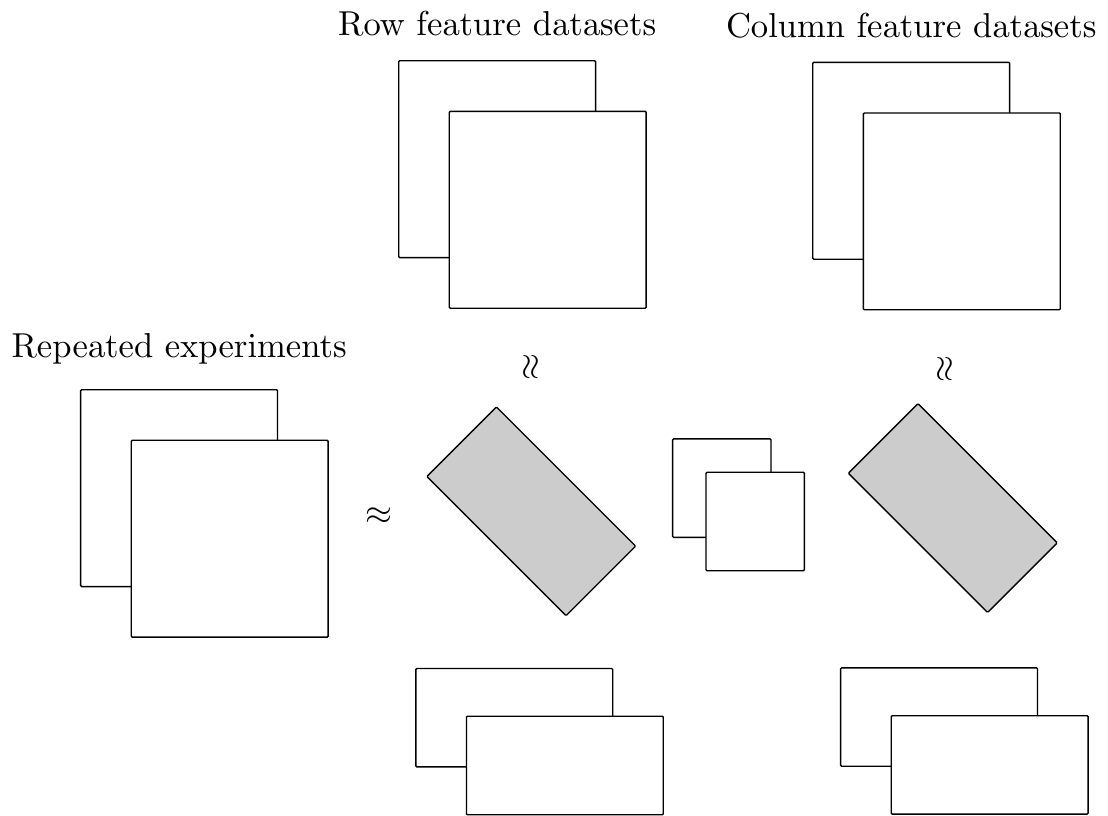}
		\captionsetup{width=0.98\columnwidth}
		\caption{Overview of HMF, combining the multiple matrix tri-factorisation of two repeated experiments with multiple matrix factorisations of row and column feature datasets. Shared factor matrices are grey.}
		\label{hmf_overview}
	\end{figure}

	The idea behind Hybrid Matrix Factorisation (HMF) is to integrate multiple datasets by jointly decomposing them, and sharing their latent factors. 
	Formally, we are given a number of datasets spanning $T$ different entity types $E_1, .., E_T$. Each entity type $E_t$ has $ I_t $ instances, $K_t$ factors, and a factor matrix $ \F^t \in \mathbb{R}^{I_t \times K_t} $, which is shared across the matrix factorisations of datasets that relate this entity type.
	We consider three dataset types, which we decompose in different ways (see Figure \ref{hmf_3_factorisations}):
	\begin{figure}[!t]
		\centering
		\includegraphics[width=\columnwidth]{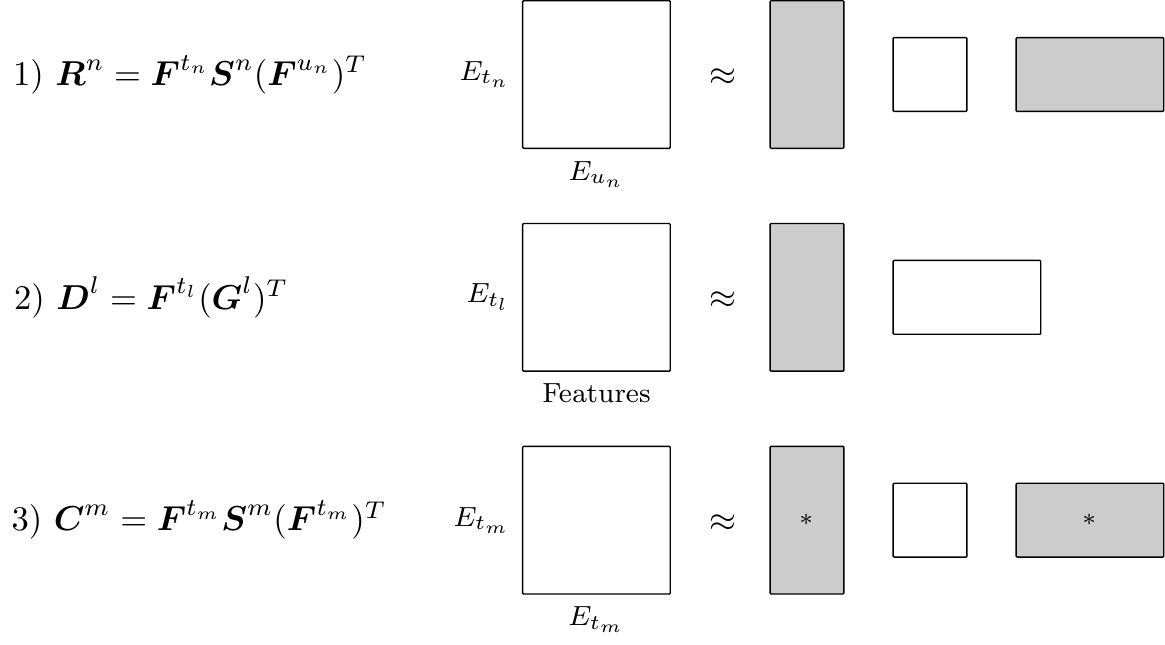}
		\captionsetup{width=0.98\columnwidth}
		\caption{The three different types of datasets and factorisations. Shared factor matrices are grey, and dataset-specific ones are white. The two grey matrices for the third factorisation type are the same (*).}
		\label{hmf_3_factorisations}
	\end{figure}
	\begin{enumerate}
		\item Main datasets $ \R = \lbrace \R^1, .., \R^N \rbrace $, relating two entity types, both of which we have other datasets for (such as features or repeated experiments). 
		Each dataset $\R^n \in \mathbb{R}^{I_{t_n} \times I_{u_n}} $ relates entity types $E_{t_n}$, $E_{u_n}$. We use matrix tri-factorisation to decompose it into two entity type factor matrices $\F^{t_n}, \F^{u_n}$, and a dataset-specific matrix $ \S^n \in \mathbb{R}^{K_{t_n} \times K_{u_n}} $.
		%
		\item Feature datasets $ \D = \lbrace \D^1, .., \D^L \rbrace $, giving feature values for an entity type. 
		Each dataset $\D^l \in \mathbb{R}^{I_{t_l} \times J_l} $ relates an entity type $E_{t_l}$ to $J_l$ features. We use matrix factorisation to decompose it into one entity type factor matrix $\F^{t_l}$, and a dataset-specific matrix $ \G^l \in \mathbb{R}^{J_l \times K_{t_l}} $.
		%
		\item Similarity datasets $ \C = \lbrace \C^1, .., \C^M \rbrace $, giving similarities between entities of the same entity type (such as Jaccard kernels).
		Each dataset $ \C^m \in \mathbb{R}^{I_{t_m} \times I_{t_m}} $ relates an entity type $ E_{t_m} $ to itself. We use matrix tri-factorisation to decompose it into a entity type factor matrix $\F^{t_m}$, a dataset-specific matrix $ \S^m \in \mathbb{R}^{K_{t_m} \times K_{t_m}} $, and $\F^{t_m}$ again.
		\begin{align*}
			\R^n &= \F^{t_n} \S^n (\F^{u_n})^T + \E^n \\
			\D^l &= \F^{t_l} (\G^l)^T + \E^l \\
			\C^m &= \F^{t_m} \S^m (\F^{t_m})^T + \E^m
		\end{align*}
	\end{enumerate}
	The above formulation allows the user to very easily choose the kind of joint factorisation. 
	By passing a set of matrices as $D_1, .., D_L$, multiple matrix factorisation is performed. Instead, passing them as $R_1, .., R_N$ gives multiple matrix tri-factorisation. A hybrid combination is also possible, as illustrated in Figure \ref{hmf_overview}.
	Furthermore, each of the factor matrices can either be nonnegative (using an exponential prior), or real-valued (using a Gaussian prior), 
	additionally giving a hybrid of nonnegative, semi-nonnegative and real-valued matrix factorisation.
	The model likelihood functions are
	\begin{align*}
		& R^n_{ij} \sim \mathcal{N} (R^n_{ij} | \F^{t_n}_i \cdot \S^n \cdot \F^{u_n}_j, (\tau^n)^{-1} )	\\
		& D^m_{ij} \sim \mathcal{N} (D^l_{ij} | \F^{t_l}_i \cdot \G^{l}_j, (\tau^l)^{-1} ) 		\\
		& C^m_{ij} \sim \mathcal{N} (C^m_{ij} | \F^{t_m}_i \cdot \S^m \cdot \F^{t_m}_j, (\tau^m)^{-1} ),
	\end{align*}
	with Bayesian priors
	\begin{align*}
	 	& \omit\rlap{ $ \tau^n, \tau^l, \tau^m \sim \mathcal{G} (\tau^* | \alpha_{\tau}, \beta_{\tau} ) $ } \\
		& F^t_{ik} \sim \mathcal{E} ( F^t_{ik} | \lambda_k^t)		 		 && \text{ or } \quad 		F^t_{ik} \sim \mathcal{N} ( F^t_{ik} | 0, (\lambda_k^t)^{-1}) \\
		& G^l_{jk} \sim \mathcal{E}( G^l_{jl} | \lambda_k^{t_l})		 		 && \text{ or } \quad 		G^l_{jk} \sim \mathcal{N}( G^l_{jl} | 0, (\lambda_k^{t_l})^{-1} ) \\
		& S^n_{kl} \sim \mathcal{E}( S^n_{kl} | \lambdaS^n)		 		 && \text{ or } \quad		 S^n_{kl} \sim \mathcal{N}( S^n_{kl} | 0, (\lambdaS^n)^{-1}) \\
		& S^m_{kl} \sim \mathcal{E}( S^m_{kl} | \lambdaS^m)		 		 && \text{ or } \quad		 S^m_{kl} \sim \mathcal{N}( S^m_{kl} | 0, (\lambdaS^m)^{-1}).
	\end{align*}
	%
	\paragraph{Automatic Relevance Determination (ARD)} We employ a Bayesian ARD prior, which helps perform automatic model selection. Note the $\lambda_k^t$ parameters in the prior of $F^t_{ik}$ and $G^l_{jk}$. This parameter is shared by all entities of type $E_t$, and hence the entire factor $k$ is either activated (if $\lambda_k^t$ has a low value) or ``turned off'' (if $\lambda_k^t$ has a high value). The ARD works by placing a Gamma prior over each of these variables, 
	\begin{equation*}
		\lambda_k^t \sim \mathcal{G} (\lambda_k^t | \alpha_0, \beta_0 ).
	\end{equation*}
	Through this construction, factors that are active for only a few entities will be pushed further to zero, turning the factor off. This prior has been used extensively for model selection in \citet{Virtanen2011,Virtanen2012} for real-valued matrix factorisation, and \cite{Tan2013} for nonnegative matrix factorisation. Instead of having to choose the correct values for the $K_t$, we can give an upper bound and our model will automatically determine the number of factors to use.
	
	%
	%
	\paragraph{Dataset importance} One challenge with multiple matrix factorisation is that it relies on finding common patterns in multiple datasets. If two datasets are very different, the methods may end up finding a solution that fits one dataset much better, resulting in poor predictions for the other one. To address this, we add an importance value for each of the $\R^n, \D^l, \C^m$ datasets, respectively $\alpha_n, \alpha_l, \alpha_m$, to ensure that the method will converge to a solution that better fits datasets with higher importance values. We modify the likelihood of the model by using these importance values,
	\begin{alignat*}{1}
		& p( \boldsymbol \theta | \R, \D, \C ) \propto 
		p(\boldsymbol \theta) \times \prod_{n=1}^N p(\R^n | \F^{t_n}, \S^n, \F^{u_n}, \tau^n )^{\alpha^n} \\
		& \times \prod_{l=1}^L p(\D^l | \F^{t_l}, \G^l, \tau^l )^{\alpha^l} \times \prod_{m=1}^M p(\C^m | \F^{t_m}, \S^m, \tau^m )^{\alpha^m} 
	\end{alignat*}
	where $\boldsymbol \theta$ is the set of model parameters.
	This technique was used by \citet{Remes2015} to ensure their model fits the binary training labels. This technique can be interpreted as repeating each of the values in the dataset $\D^l$ $\alpha^l$ times, hence forcing the model to fit better to that dataset.
	
	\paragraph{Inference}
	An efficient Gibbs sampling algorithm can be used for inference due to the model's conjugacy. For details see Supplementary Materials (Section 1). 

\section{RELATED WORK} \label{Related Work}
	The idea of using matrix factorisation and tri-factorisation to integrate multiple datasets can be traced back to the CANDECOMP/PARAFAC (CP) and PARAFAC2 tensor decompositions (\cite{Harshman1970,Harshman1972}). These models are in fact a less general version of multiple matrix tri-factorisation. If we are given multiple datasets for the same two entity types and concatenate them to form a tensor, the CP method will perform multiple matrix tri-factorisation, where the dataset-specific middle matrices $\S$ are restricted to being diagonal.
	
	Multiple matrix factorisation models for integrating datasets between two entity types (such as multiple gene expression profiles), by sharing one of the two factor matrices, can be found amongst others in
	\citet{Zhang2005} and \citet{Lee2012}, 
	with Bayesian models given by \citet{Virtanen2012} and \citet{Chatzis2014}.
	Some approaches focus on jointly decomposing two datasets spanning three entity types and sharing two latent matrices 
	(\cite{Shi2010}), 
	sometimes using supervised labels for learning 
	(\cite{Zhu2007}). 
	Others do not explicitly share the latent matrices but instead add a penalisation term based on the consensus between the matrices 
	(\cite{Seichepine2013}). 
	
	More general matrix factorisation methods are presented by \citet{MatthiasSchubert2008} and \citet{Singh2008},  
	where each entity type has its own latent matrix, with a Bayesian version given in \cite{Klami2014}. 
	However, these approaches cannot integrate multiple datasets between the same two entity types, since all matrices are shared. We would require a third, dataset-specific matrix to solve this problem -- which is exactly what matrix tri-factorisation allows us to do.
	Models for multiple non-negative matrix tri-factorisation are given by \cite{FeiWangTaoLi2008} and \cite{Zitnik2015}, which can also handle constraint matrices, but require all given datasets to be fully observed. As a result, missing values inside each matrix need to be imputed. For binary datasets a missing association can easily be imputed as a zero, but for real-valued datasets this is not a viable option.
		
	
	Overall, our method is novel in several aspects. 
	Firstly, it is the first general hybrid model between matrix factorisation and tri-factorisation. A non-probabilistic version can be found in \cite{Zhu2007}, but this model only combined a single matrix tri-factorisation with a single matrix factorisation.
	Secondly, our model is a hybrid between nonnegative and real-valued factors. If multiple datasets are jointly decomposed, one can be a nonnegative matrix factorisation, where another can be semi-nonnegative, and another can be real-valued.
	Finally, through formulating the method as a Bayesian probabilistic model, it can deal with missing values, perform automatic model selection, and is much less prone to overfitting (especially for sparse datasets).
	
	In this paper we are demonstrating our method on two specific biological datasets. However, it can be widely applied to other biological applications such as predicting drug-target interactions (\cite{Gonen2012}) or gene functions (\cite{MatthiasSchubert2008}), as well as other fields like collaborative filtering (\cite{Mnih2008}) and image analysis (\cite{Zhang2005}).

\section{DATASETS}

	\begin{table*}[t]
		\caption{Overview of the four drug sensitivity dataset after preprocessing and filtering.} 
		\label{Drug sensitivity datasets}
		\centering
		\begin{tabular}{lccccccc}
			\toprule
			& Number & Number & Fraction & \multicolumn{4}{l}{Overlap with other datasets} \\
			\cmidrule(lr){5-8}
			Dataset & cell lines & drugs & observed & GDSC $IC_{50}$ & CTRP $EC_{50}$ & CCLE $IC_{50}$ & CCLE $EC_{50}$ \\
			\midrule
			GDSC $IC_{50}$ 		& 399 		& 48 		& 73.57\%	& - 	& 52.25\%	& 9.34\%	& 6.00\% \\
			CTRP $EC_{50}$		& 379 		& 46 		& 86.03\%	& 57.39\%	& - 	& 11.96\% 	& 7.37\% \\
			CCLE $IC_{50}$	 	& 253 		& 16  		& 96.42\% 	& 44.19\% 	& 51.51\% 	& - 	& 55.06\% \\
			CCLE $EC_{50}$		& 252 		& 16 		& 58.88\% 	& 28.52\% 	& 31.87\% 	& 55.28\% 	& - \\
			\bottomrule
		\end{tabular}
	\end{table*}

	To demonstrate the advantages of our approach for missing values prediction, we consider two different applications. Firstly, integrating four drug sensitivity datasets, where the datasets are similar and hence predictivity of the datasets is high.
	Here we perform in-matrix predictions of missing values. 
	Secondly, integrating gene expression and methylation level datasets for breast cancer patients and cancer driver genes, where the datasets are much more dissimilar. We perform out-of-matrix predictions, using the methylation levels of patients to predict gene expression values, and vice versa. 
	We briefly introduce the datasets below; a more thorough description of the datasets can be found in the Supplementary Materials (Section 3).
	
	\subsection{Drug Sensitivity Data}
		We consider four different drug sensitivity datasets, containing 650 unique drugs and 1209 cell lines. Each of these datasets shows the response (sensitivity) of a given cell line (cancer type in a tissue) to a given drug, either measuring the drug concentration needed to inhibit undesired cell line activity by half ($IC_{50}$), or the drug concentration that achieves half the maximal desired effect on the cell line ($EC_{50}$). 
		\begin{itemize}
			\item Genomics of Drug Sensitivity in Cancer (GDSC v5.0, \citet{Yang2013}). Natural log of $IC_{50}$ values for 139 drugs across 707 cell lines, with 80\% observed entries.
			\item Cancer Therapeutics Response Portal (CTRP v2, \cite{Seashore-Ludlow2015}). $EC_{50}$ values for 545 drugs across 887 cell lines, with 80\% observed entries.
			\item Cancer Cell Line Encyclopedia (CCLE, \cite{Barretina2012}). Both $IC_{50}$ and $EC_{50}$ values for 24 drugs across 504 cell lines, with 96\% and 63\% observed entries, respectively.
		\end{itemize}
		
		
		We selected the drugs and cell lines that are present in at least two of the four datasets, and for which we had side information like gene expression profiles available. This resulted in a lot of drugs and cell lines being filtered. For the GDSC dataset we undid the log transform. We rescaled the values per cell line to the range [0,1] in each dataset. We used the cell line features provided by the GDSC dataset (gene expression levels, copy number variations, and mutation information), and for the drugs we extracted 1D and 2D descriptors and structural fingerprints. We obtained primary protein targets from GDSC for 48 of the 52 drugs. 
		
		After preprocessing and filtering, the four datasets span 52 unique drugs and 399 cell lines, with 95.1\% of the entries having at least one observed value, and 62.9\% of the entries having at least two observed values. The information on the four datasets is summarised in Table \ref{Drug sensitivity datasets}, along with the fraction of overlapping observed entries.

	\subsection{Methylation and Gene Expression Data}	
		Our second application is that of integrating promoter-region methylation (PM) and gene body methylation (GM) datasets with a gene expression (GE) profile for breast cancer patients, coming from the The Cancer Genome Atlas (TCGA, \cite{Koboldt2012}).
		There are 254 different samples (both healthy and tumor tissues), across 13966 genes. We focus on 160 breast cancer driver genes, from the IntOGen database (\cite{Gonzalez-Perez2013}). We standardise the datasets to have zero mean and unit standard deviation per gene. Note that this dataset is not nonnegative.
		In our experiments we predict values in one of the three datasets, given the values of the other two.

\section{IN-MATRIX PREDICTIONS} \label{Cross-validation drug sensitivity}

	
	
		\begin{table*}[t]
			\caption{Mean squared error (MSE) of 10-fold in-matrix cross-validation results on the drug sensitivity datasets. We also give the relative improvement (\% impr.) compared to NMF. The best performances are highlighted in bold.} \label{Results drug sensitivity}
			\centering
			\begin{tabular}{lcccccccc}
				\toprule
				& \multicolumn{2}{c}{GDSC $IC_{50}$} & \multicolumn{2}{c}{CTRP $EC_{50}$} & \multicolumn{2}{c}{CCLE $IC_{50}$} & \multicolumn{2}{c}{CCLE $EC_{50}$} \\
				\cmidrule(lr){2-3} \cmidrule(lr){4-5} \cmidrule(lr){6-7} \cmidrule(lr){8-9}
				Method & MSE & \% impr. & MSE & \% impr. & MSE & \% impr. & MSE & \% impr. \\
				\midrule
				NMF \hspace{40pt} 				& 0.0896 	& - 		&  0.0959 & -	 & 0.0746 & -		 & 0.1535 & - \\
				NMTF 							& 0.0879 	& 1.91\% 	&  0.0954 & 0.44\% & 0.0747 & -0.18\% & 0.1506 & 1.91\% \\
				\vspace{5pt} Multiple NMF	 	& 0.0859 	& 4.10\% 	&  0.0928 & 3.18\% & 0.0666 & 10.64\% & 0.1157 & 24.66\% \\
			 	BNMF 				& 0.0805 	& 10.20\% 	& 0.0919 	& 4.05\% & 0.0594 & 20.29\% & 0.1318 & 14.19\% \\
				BNMTF 							& 0.0799 	& 10.81\% 	&  0.0920 & 4.03\% & 0.0593 & 20.52\% & 0.1292 & 15.84\% \\
				\vspace{5pt} KBMF 				& 0.0819 	& 8.60\% 	&  0.0919 & 4.13\% & 0.0618 & 17.13\% & 0.1303 & 15.13\% \\
				LR 								& 0.0886 	& 1.10\% 	&  0.0949 & 1.00\% & 0.0719 & 3.62\% & 0.1342 & 12.60\% \\
				RF 								& 0.0876 	& 2.21\% 	&  0.0989 & -3.15\% & 0.0668 & 10.47\% & 0.1219 & 20.62\% \\
				\vspace{5pt} SVR 				& 0.1091 	& -21.72\% 	&  0.1091 & -13.80\% & 0.0916 & -22.76\% & 0.1230 & 19.92\% \\
				HMF D-MF 						& 0.0775 	& 13.54\% 	&  0.0919 & 4.11\% & 0.0592 & 20.65\% & \textbf{0.1062} & \textbf{30.81\%} \\
				HMF D-MTF 						& \textbf{0.0768} 	& \textbf{14.25\%} 	&  \textbf{0.0908} & \textbf{5.28\%} & \textbf{0.0558} & \textbf{25.17\%} & 0.1073 & 30.12\% \\
				\bottomrule
			\end{tabular}
		\end{table*}
		
		We performed 10-fold cross-validation on each of the four drug sensitivity datasets to predict missing values. 
		We tested two variants of our HMF model: multiple matrix tri-factorisation using all four drug sensitivity datasets (HMF D-MTF, $\R_n$), and multiple matrix factorisation on all four drug sensitivity datasets, sharing the cell line factors (HMF D-MF, $\D_l$). 
				
		We compared our model to several state of the art methods.
		Since the four datasets are all nonnegative, we can use nonnegative matrix factorisation (NMF) and tri-factorisation (NMTF) models. We compare with non-probabilistic NMF by \citet{Lee2001} (NP-NMF), Bayesian NMF by \citet{Schmidt2009a} (BNMF), non-probabilistic NMTF by \citet{Yoo2009} (NP-NMTF), Bayesian NMTF (BNMTF), and Multiple NMF (sharing the cell line factors). 
		We also applied several state-of-the-art machine learning models using the skikit-learn Python package, particularly: Linear Regression (LR), Random Forests (RF, 100 trees), and  Support Vector Regression (SVR, \textit{rbf} kernel). These methods were given the drug and cell line features for training.
		Finally, we used a method called Kernelised Bayesian Matrix Factorisation (KBMF, \citet{Gonen2014}), which was used by \citet{Ammad-ud-din2014} to predict drug sensitivity values for the GDSC dataset. This method leverages similarity kernels of the drugs and cell lines, which we reconstructed for the feature datasets (Jaccard kernel for binary features, Gaussian for real-valued features after standardising each feature).
		
		We performed nested cross-validation to select the dimensionality $K$ for the matrix factorisation models and KBMF.
		In contrast, our model simply used $ K_t = 10 $ for each entity type $E_t$, and let the ARD choose the correct number of factors. 
		We used nonnegative factors for the entity type factor matrices ($\F_t$), and real-valued for all other factors. We used $K$-means and least squares initialisation, and set all importance values to one. 
		
		The results for cross-validation are given in Table \ref{Results drug sensitivity}. We see that our HMF models outperform all other methods, giving predictive gains of up to 30\%. The multiple matrix tri-factorisation approach (HMF D-MTF) achieves the best performance on three of the datasets, and is a close second on the fourth.
		We also see that the Bayesian matrix factorisation models outperform both the non-probabilistic approaches, and the state-of-the-art machine learning methods, demonstrating that Bayesian matrix factorisation is a powerful paradigm for in-matrix predictions, with our proposed HMF model giving significant gains in predictive performance. 

\section{SPARSE DATA PREDICTIONS}
	\begin{figure*}[t]
		\centering
		\begin{subfigure}[t]{1\columnwidth}
			\centering
			\includegraphics[width=1\columnwidth]{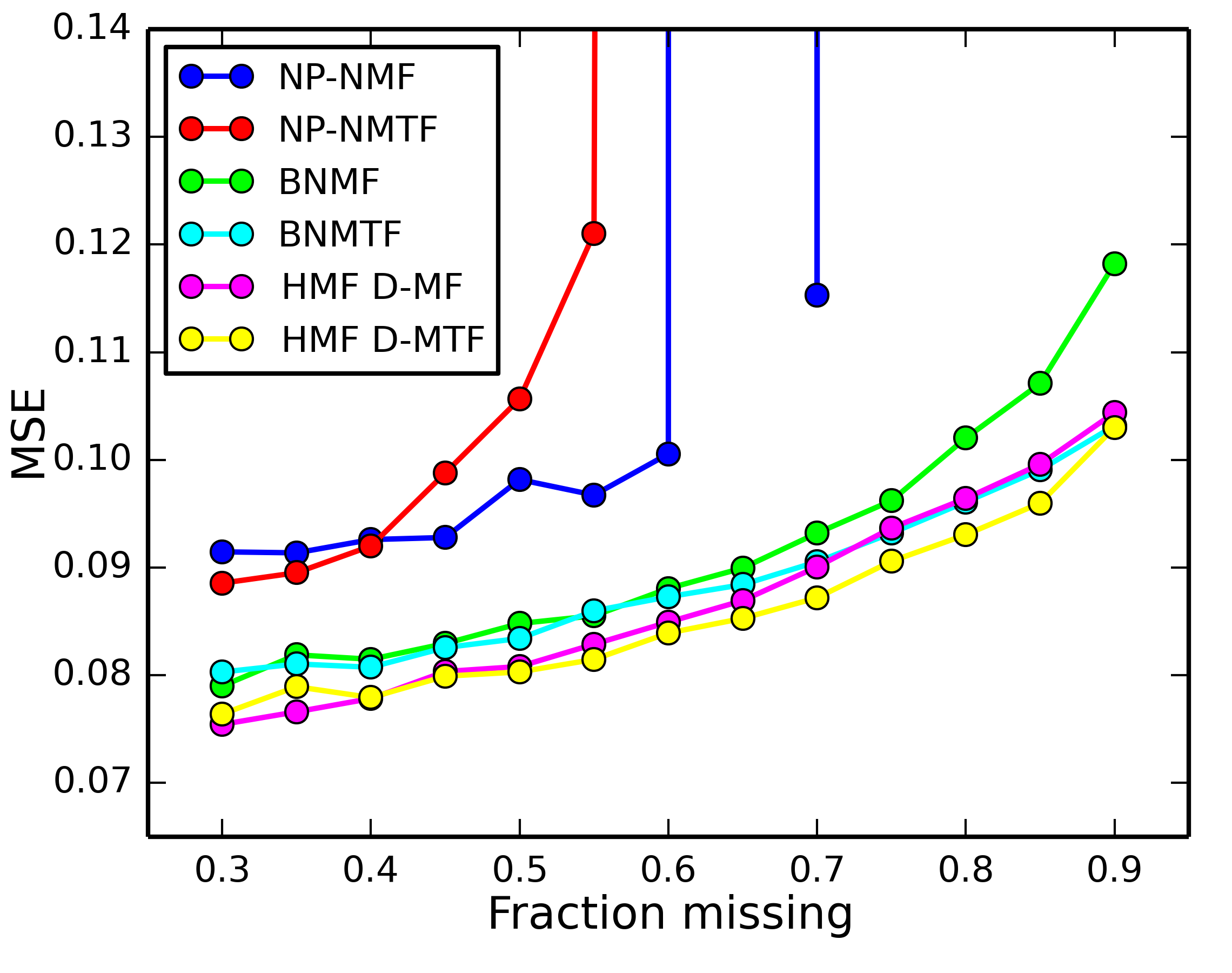}
			\captionsetup{width=0.95\columnwidth}
			\caption{GDSC} 
			\label{mse_missing_values_average_gdsc}
		\end{subfigure}
		\begin{subfigure}[t]{1\columnwidth}
			\centering
			\includegraphics[width=1\columnwidth]{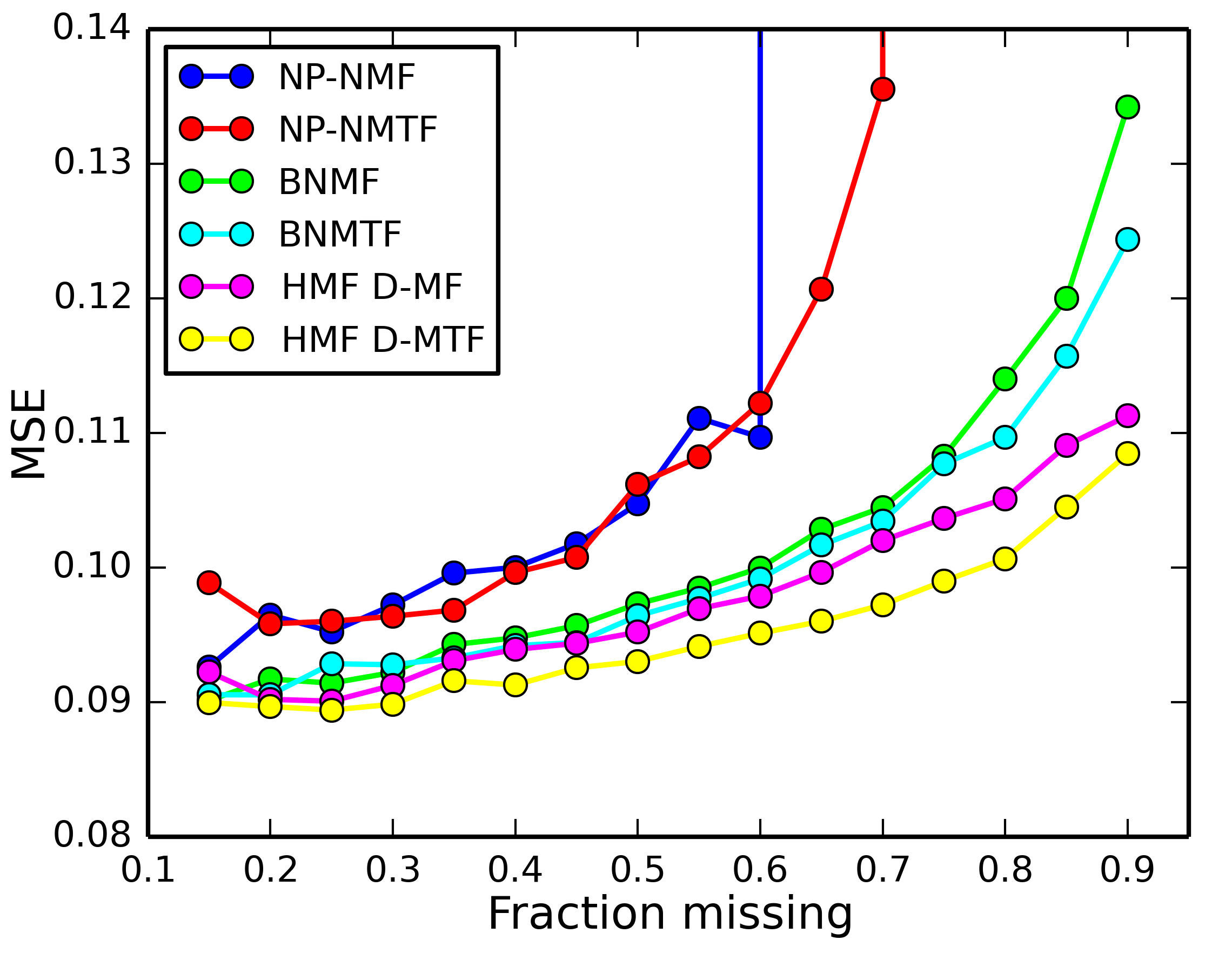}
			\captionsetup{width=0.95\columnwidth}
			\caption{CTRP} 
			\label{mse_missing_values_average_ctrp}
		\end{subfigure}
		\caption{Graphs showing average mean squared error (MSE) and standard deviation of in-matrix predictions on the GDSC (left) and CTRP (right) drug sensitivity datasets. We vary the fraction of missing entries, averaging performance across 20 random splits between train and test data, and compare our HMF models (HMF D-MF, HMF D-MTF) with several matrix factorisation models (NMF, NMTF, BNMF, BNMTF).}
		\label{mse_missing_values_average}
	\end{figure*}
	
	A very important use case is when there are few observed entries, leading to a sparse matrix. We measured the performances of in-matrix predictions on sparse matrices, focusing on the GDSC and CTRP drug sensitivity datasets as these are the largest. We vary the fraction of missing values and predict those entries, taking the average of twenty random training-test data splits per fraction. We compared our model's multiple matrix factorisation and tri-factorisation models (HMF D-MF and HMF D-MTF) with the other matrix factorisation models (NMF, NMTF, BNMF, BNMTF). For the dimensionality of HMF we use $K_t = 10$ as before, and for the matrix factorisation models we use the most common dimensionality used in the cross-validation from Section \ref{Cross-validation drug sensitivity}.\footnote{GDSC: $K=2$, $(K,L) = (4,4)$, $K=4$, $(K,L) = (7,7)$. CTRP: $K=2$, $(K,L) = (2,4)$, $K=3$, $(K,L) = (3,3)$.}
	
	Figure \ref{mse_missing_values_average} shows that the non-probabilistic models start overfitting very quickly as the sparsity levels of two datasets increase, on both the GDSC (\ref{mse_missing_values_average_gdsc}) and CTRP (\ref{mse_missing_values_average_ctrp}) datasets. The Bayesian versions perform lot better, but our HMF models consistently outperform all other models, even when only 10\% of the values are observed. The multiple matrix tri-factorisation model (HMF D-MTF) performs particularly well.

\section{OUT-OF-MATRIX PREDICTIONS}
	We did three out-of-matrix prediction experiments on the methylation and gene expression data. 
	We performed ten-fold cross-validation, splitting the 254 samples into ten folds. We predicted the gene expression values for new samples, given the gene expression values of the other samples and both of the methylation datasets (PM, GM to GE).
	We also did this for the other two combinations (GE, GM to PM; GE, PM to GM).
	Methylation data is known to be correlated with gene expression values (\cite{Kundaje2015}), although this correlation is generally weak. 
	We therefore expected a weak predictive performance, but it is interesting to see which methods perform best.
	
	We used the HMF D-MF and HMF D-MTF models described earlier. 
	We also considered the similarity dataset part of our model ($\C_m$) by constructing a similarity kernel for the samples using each of the datasets (see Supplementary Materials, Section 3.4). We give the model the dataset we are trying to predict (e.g. GE), decomposing it using matrix factorisation, and also give it the similarity kernels for the other two (e.g. GM and PM).
	We call this approach HMF S-MF. We could have also used matrix tri-factorisation, but since the third matrix is not shared this is effectively the same model.

	For the HMF D-MF models we used $K_t = 40$, 0.5 as the importance value for the dataset we are trying to predict, and 1.5 for the other two.
	For HMF D-MTF we used $K_t = 40$, and 0.5 as importance for all three datasets.
	Finally, for HMF S-MF we used $K_t = 30$, and 1.0 as importance for all three datasets.
	For all three, we used nonnegative factors for shared matrices ($K$-means initialisation), and real-valued ones for private matrices (least squares initialisation).

	We compared with the LR, RF, and SVR algorithms, giving two datasets as features, and the third as regression values. We used the gene average as a baseline.
	Since the datasets are real-valued, we cannot compare with any nonnegative matrix factorisation models.
	
	\begin{table}[t]
		\caption{Mean squared error (MSE) of 10-fold out-of-matrix cross-validation results on the promoter-region methylation (PM), gene body methylation (GM), and gene expression (GE) datasets. We use two datasets as features, and predict values for new samples in the third dataset. The best results are highlighted in bold.} \label{Methylation results}
		\centering
		\begin{tabular}{lccc}
			\toprule
			& GM, PM & GE, GM & GE, PM \\
			Method & to GE & to PM & to GM \\
			\midrule
			Gene average 	&  	1.009			& 	1.008			&	1.009			\\
			LR		 		&  	2.847			& 	2.036			&	1.478			\\
			RF		 		& 	0.811 			&  	0.799			&	0.714			\\
			SVR				&  	\textbf{0.767}	& 	0.749			&	0.657			\\
			HMF D-MF		&	0.788			&	\textbf{0.735}	&	\textbf{0.602}	\\
			HMF D-MTF		&	0.850			&	0.798			&	0.640			\\
			HMF S-MF		&	0.820			&	0.794			&	0.672			\\
			\bottomrule
		\end{tabular}
	\end{table}
	
	The results for this out-of-matrix cross-validation are given in Table \ref{Methylation results}. 
	The HMF D-MF model outperforms all state-of-the-art machine learning methods on two of the three datasets, and is only beaten by SVR on the first one. Our model performs especially well on the third case (GE, PM to GM), implying our method works best when the predictivity of values is high (lower MSE).
	The HMF D-MTF and HMF S-MF methods perform slightly worse, but are still competitive with the other machine learning methods.
	
	Many of the model choices in the experiments (such as model selection, initialisation, factorisation and negativity choices, and importance values) are explored extensively in Section 4 of the Supplementary Materials.

\section{CONCLUSION} \label{Conclusion}
	We have presented a fully Bayesian model for data integration, based on a hybrid of nonnegative, semi-nonnegative, and real-valued matrix factorisation and tri-factorisation models.
	The general nature of this model allows it to easily integrate many datasets across different entity types, including repeated experiments, similarity matrices, and very sparse datasets.
	
	We demonstrated the model on two different biological applications.
	On four drug sensitivity datasets we obtained significant in-matrix prediction improvements compared to state-of-the-art matrix factorisation and machine learning methods. Our data fusion approach based on multiple matrix tri-factorisation (HMF D-MTF) is particularly powerful, achieving the best performance on three of the four datasets.
	We also show that our proposed model can provide consistently better predictions on very sparse datasets, outperforming all other matrix factorisation models.
	Finally, we integrated methylation and gene expression data in an out-of-matrix prediction setting, and here the approach based on multiple matrix factorisation (HMF D-MF) proved to be very powerful, beating all state-of-the-art machine learning methods on two of the three datasets. The approaches using multiple matrix tri-factorisation and similarity datasets are also promising.
	
	We showcased our model on different biological datasets, but we believe that this is a powerful and general framework that can also be applied to many other fields.

\subsubsection*{Acknowledgements}
This work was supported by the UK Engineering and Physical Sciences Research Council (EPSRC), grant reference EP/M506485/1; and Methods for Integrated analysis of Multiple Omics datasets (MIMOmics, 305280).

\subsubsection*{References}
\bibliographystyle{abbrvnat}
\bibliography{bibliography}

\end{document}


\title{Bayesian Hybrid Matrix Factorisation for Data Integration \\ \vspace{10pt} Supplementary Materials \\ \vspace{10pt} Thomas Brouwer and Pietro Li\'{o}}
	\author{}
	\date{}
	\maketitle{}
	
	\noindent \large{20th International Conference on Artificial Intelligence and Statistics (AISTATS 2017).} \\
	
	\tableofcontents
	
	\clearpage
		

\section{Models}
	In the model for Hybrid Matrix Factorisation (HMF), we combine models for Bayesian matrix factorisation, tri-factorisation, and tri-factorisation of similarity kernels. For each model, there are three versions: nonnegative, semi-nonnegative, and real-valued. In this section, we give the details for each of these nine models, as well as the Gibbs sampling algorithms. \\
	
	\noindent In addition, for the semi-nonnegative and real-valued versions we can either use a univariate Gaussian (resulting in individual draws, but those can be drawn in parallel per column), or a multivariate Gaussian as the posterior (resulting in row-wise draws of new values, but each row can be drawn in parallel at the same time). We give Gibbs sampling algorithms for both options. For the nonnegative models, where the posterior is a truncated normal, this is technically also possible. However, we did not succeed in finding an efficiently implemented library for multivariate truncated normal draws, and therefore do not support it.  \\
	
	\noindent We first introduce each of the models separately in Section \ref{Matrix factorisation models}, extending them with Automatic Relevance Determination in Section \ref{Matrix factorisation with ARD}, and finally explain how we combine them into one to form the HMF model in Section \ref{Hybrid matrix factorisation model}.
	
	\subsection{Matrix factorisation models} \label{Matrix factorisation models}
		\noindent The matrix factorisation models are:
		\begin{itemize}
			\item Bayesian matrix factorisation (BMF), nonnegative matrix factorisation (BNMF), semi-nonnegative matrix factorisation (BSNMF). We decompose $\D \approx \F \cdot \G^T$. The Gibbs samplers are given in Table \ref{bmf}. 
			\item Bayesian matrix tri-factorisation (BMTF), nonnegative matrix tri-factorisation (BNMTF), and semi-nonnegative matrix tri-factorisation (BSNMTF). We decompose $\R \approx \F \cdot \S \cdot \G^T$. The Gibbs samplers are given Table \ref{bmtf}. 
			\item Bayesian similarity matrix tri-factorisation (BSMTF), nonnegative similarity matrix tri-factorisation (BNSMTF), and semi-nonnegative similarity matrix tri-factorisation (BSNSMTF). We decompose $\C \approx \F \cdot \S \cdot \F^T$. The Gibbs sampler are given in Table \ref{bsmtf}. 
		\end{itemize}
	
		\begin{table*}[h!]
			\captionsetup{width=\columnwidth}
			\caption{Model definitions, and Gibbs sampling posteriors.}
			\vspace{5pt}
			\label{models}
			\centering
			\begin{tabular}{llllc}
				\toprule
				Model name & Likelihood & Priors & Posteriors & Table \\
				\cmidrule(lr){1-1} \cmidrule(lr){2-2} \cmidrule(lr){3-3} \cmidrule(lr){4-4} \cmidrule(lr){5-5}
				BMF & $ D_{ij} \sim \mathcal{N} (D_{ij} | \boldsymbol F_i \cdot \boldsymbol G_j, \tau^{-1} ) $ & $F_{ik} \sim \mathcal{N} ( F_{ik} | 0, \lambda_F^{-1})$ & $\mathcal{N} ( F_{ik} | \muFik, (\tauFik)^{-1} )$ & \ref{bmf} \\
				& & $G_{jk} \sim \mathcal{N} ( G_{jk} | 0, \lambda_G^{-1})$ & $\mathcal{N} ( G_{jk} | \muGjk, (\tauGjk)^{-1} )$ & \\
				\vspace{5pt} & & $\tau \sim \mathcal{G} (\tau | \alpha, \beta )$ & $\mathcal{G} (\tau | \alpha^*, \beta^* )$ & \\
				BMF (multivariate) & & & $\mathcal{N} ( \F_{i} | \muFi, \SigmaFi ) $ & \ref{bmf} \\
				\vspace{5pt} & & & $\mathcal{N} ( \G_{j} | \muGj, \SigmaGj ) $ & \\
				BNMF & $ D_{ij} \sim \mathcal{N} (D_{ij} | \boldsymbol F_i \cdot \boldsymbol G_j, \tau^{-1} ) $ & $F_{ik} \sim \mathcal{E} ( F_{ik} | \lambda_F)$ & $\mathcal{TN} ( F_{ik} | \muFik, \tauFik )$ & \ref{bmf} \\
				& & $G_{jk} \sim \mathcal{E} ( G_{jk} | \lambda_G)$ & $\mathcal{TN} ( G_{jk} | \muGjk, \tauGjk )$ & \\
				\vspace{5pt} & & $\tau \sim \mathcal{G} (\tau | \alpha, \beta )$ & $\mathcal{G} (\tau | \alpha^*, \beta^* )$ & \\
				BSNMF & $ D_{ij} \sim \mathcal{N} (D_{ij} | \boldsymbol F_i \cdot \boldsymbol G_j, \tau^{-1} ) $ & $F_{ik} \sim \mathcal{E} ( F_{ik} | \lambda_F)$ & $\mathcal{TN} ( F_{ik} | \muFik, \tauFik )$ & \ref{bmf} \\
				& & $G_{jk} \sim \mathcal{N} ( G_{jk} | 0, \lambda_G^{-1})$ & $\mathcal{N} ( G_{jk} | \muGjk, (\tauGjk)^{-1} )$ & \\
				\vspace{5pt} & & $\tau \sim \mathcal{G} (\tau | \alpha, \beta )$ & $\mathcal{G} (\tau | \alpha^*, \beta^* )$ & \\
				\vspace{5pt}BSNMF (multivariate) & & & $\mathcal{N} ( \G_{j} | \muGj, \SigmaGj ) $ & \ref{bmf} \\
				BMTF & $ R_{ij} \sim \mathcal{N} (R_{ij} | \boldsymbol F_i \cdot \S \cdot \boldsymbol G_j, \tau^{-1} ) $ & $F_{ik} \sim \mathcal{N} ( F_{ik} | 0, \lambda_F^{-1})$ & $\mathcal{N} ( F_{ik} | \muFik, (\tauFik)^{-1} )$ & \ref{bmtf} \\
				& & $S_{kl} \sim \mathcal{N} ( S_{kl} | 0, \lambda_S^{-1})$ & $\mathcal{N} ( S_{kl} | \muSkl, (\tauSkl)^{-1} )$ & \\
				& & $G_{jl} \sim \mathcal{N} ( G_{jl} | 0, \lambda_G^{-1})$ & $\mathcal{N} ( G_{jl} | \muGjl, (\tauGjl)^{-1} )$ & \\
				\vspace{5pt} & & $\tau \sim \mathcal{G} (\tau | \alpha, \beta )$ & $\mathcal{G} (\tau | \alpha^*, \beta^* )$ & \\
				BMTF (multivariate) & & & $\mathcal{N} ( \F_{i} | \muFi, \SigmaFi ) $ & \ref{bmtf} \\
				& & & $\mathcal{N} ( \S_{k} | \muSk, \SigmaSk ) $ & \\
				\vspace{5pt} & & & $\mathcal{N} ( \G_{j} | \muGj, \SigmaGj ) $ & \\
				BNMTF & $ R_{ij} \sim \mathcal{N} (R_{ij} | \boldsymbol F_i \cdot \S \cdot \boldsymbol G_j, \tau^{-1} ) $ & $F_{ik} \sim \mathcal{E} ( F_{ik} | \lambda_F)$ & $\mathcal{TN} ( F_{ik} | \muFik, \tauFik )$ & \ref{bmtf} \\
				& & $S_{kl} \sim \mathcal{E} ( S_{kl} | \lambda_S)$ & $\mathcal{TN} ( S_{kl} | \muSkl, \tauSkl )$ & \\
				& & $G_{jl} \sim \mathcal{E} ( G_{jl} | \lambda_G)$ & $\mathcal{TN} ( G_{jl} | \muGjl, \tauGjl )$ & \\
				\vspace{5pt} & & $\tau \sim \mathcal{G} (\tau | \alpha, \beta )$ & $\mathcal{G} (\tau | \alpha^*, \beta^* )$ & \\
				BSNMTF & $ R_{ij} \sim \mathcal{N} (R_{ij} | \boldsymbol F_i \cdot \S \cdot \boldsymbol G_j, \tau^{-1} ) $ & $F_{ik} \sim \mathcal{E} ( F_{ik} | \lambda_F)$ & $\mathcal{TN} ( F_{ik} | \muFik, \tauFik )$ & \ref{bmtf} \\
				& & $S_{kl} \sim \mathcal{N} ( S_{kl} | 0, \lambda_S^{-1})$ & $\mathcal{N} ( S_{kl} | \muSkl, (\tauSkl)^{-1} )$ & \\
				& & $G_{jl} \sim \mathcal{E} ( G_{jl} | \lambda_G)$ & $\mathcal{TN} ( G_{jl} | \muGjl, \tauGjl )$ & \\
				\vspace{5pt} & & $\tau \sim \mathcal{G} (\tau | \alpha, \beta )$ & $\mathcal{G} (\tau | \alpha^*, \beta^* )$ & \\
				\vspace{5pt}BSNMTF (multivariate) & & & $\mathcal{N} ( \S_{k} | \muSk, \SigmaSk ) $ & \ref{bmtf} \\
				BSMTF & $ C_{ij} \sim \mathcal{N} (C_{ij} | \boldsymbol F_i \cdot \S \cdot \boldsymbol F_j, \tau^{-1} ) $ & $F_{ik} \sim \mathcal{N} ( F_{ik} | 0, \lambda_F^{-1})$ & $\mathcal{N} ( F_{ik} | \muFik, (\tauFik)^{-1} )$ & \ref{bsmtf} \\
				& & $S_{kl} \sim \mathcal{N} ( S_{kl} | 0, \lambda_S^{-1})$ & $\mathcal{N} ( S_{kl} | \muSkl, (\tauSkl)^{-1} )$ & \\
				\vspace{5pt} & & $\tau \sim \mathcal{G} (\tau | \alpha, \beta )$ & $\mathcal{G} (\tau | \alpha^*, \beta^* )$ & \\
				BSMTF (multivariate) & & & $\mathcal{N} ( \F_{i} | \muFi, \SigmaFi ) $ & \ref{bsmtf} \\
				\vspace{5pt} & & & $\mathcal{N} ( \S_{k} | \muSk, \SigmaSk ) $ & \\
				BNSMTF & $ C_{ij} \sim \mathcal{N} (C_{ij} | \boldsymbol F_i \cdot \S \cdot \boldsymbol F_j, \tau^{-1} ) $ & $F_{ik} \sim \mathcal{E} ( F_{ik} | \lambda_F)$ & $\mathcal{TN} ( F_{ik} | \muFik, \tauFik )$ & \ref{bsmtf} \\
				& & $S_{kl} \sim \mathcal{E} ( S_{kl} | \lambda_S)$ & $\mathcal{TN} ( S_{kl} | \muSkl, \tauSkl )$ & \\
				\vspace{5pt} & & $\tau \sim \mathcal{G} (\tau | \alpha, \beta )$ & $\mathcal{G} (\tau | \alpha^*, \beta^* )$ & \\
				BSNSMTF & $ C_{ij} \sim \mathcal{N} (C_{ij} | \boldsymbol F_i \cdot \S \cdot \boldsymbol F_j, \tau^{-1} ) $ & $F_{ik} \sim \mathcal{E} ( F_{ik} | \lambda_F)$ & $\mathcal{TN} ( F_{ik} | \muFik, \tauFik )$ & \ref{bsmtf} \\
				& & $S_{kl} \sim \mathcal{N} ( S_{kl} | 0, \lambda_S^{-1})$ & $\mathcal{N} ( S_{kl} | \muSkl, (\tauSkl)^{-1} )$ & \\
				\vspace{5pt} & & $\tau \sim \mathcal{G} (\tau | \alpha, \beta )$ & $\mathcal{G} (\tau | \alpha^*, \beta^* )$ & \\
				\vspace{5pt}BSNSMTF (multivariate) & & & $\mathcal{N} ( \S_{k} | \muSk, \SigmaSk ) $ & \ref{bsmtf} \\
				\bottomrule
			\end{tabular}
		\end{table*}
		
		\noindent The model priors and Gibbs sampling posteriors are given Table \ref{models}. Here,
		\begin{itemize}
			\item $ \R, \D, \C \in \mathbb{R}^{I \times J} $, with $ i = 1..I$, $ j = 1..J $.
			\item $\F \in \mathbb{R}^{I \times K} $, with $ i = 1..I$, $k = 1..K $.
			\item $\S \in \mathbb{R}^{K \times L} $ (matrix tri-factorisation), or $\S \in \mathbb{R}^{K \times K} $ (similarity matrix tri-factorisation), with $ k = 1..K$, $l = 1..L $.
			\item $\G \in \mathbb{R}^{J \times K} $ (matrix factorisation), or $\G \in \mathbb{R}^{J \times L} $ (matrix tri-factorisation), with $ j=1..J$, $k = 1..K$, $l = 1..L $.
			\item $\mathcal{N} ( \boldsymbol{x} | \boldsymbol{\mu}, \boldsymbol{\Sigma} ) $ is a multivariate Gaussian distribution with mean vector $\boldsymbol{\mu}$ and covariance matrix $ \boldsymbol{\Sigma} $.
		\end{itemize}
		
		\noindent The Gibbs sampling posteriors can be obtained using Bayes' theorem, for example for BNMTF: 
		\begin{alignat*}{1}
			p(F_{ik}|&\tau,\F_{-ik},\S,\G,\R,h) \\
			&\propto p(\R|\tau,\F,\S,\G) \times p(F_{ik}|\lambda_F) \\
			&\propto \prod_{j \in \Omega^1_i} \mathcal{N} (R_{ij} | \F_i \cdot \S \cdot \G_j, \tau^{-1} ) \times \mathcal{E} ( F_{ik} | \lambda_F) \\
			&\propto \exp \left\{ - \frac{\tau}{2} \sum_{j \in \Omega^1_i} (R_{ij} - \F_i \cdot \S \cdot \G_j)^2 \right\} \times \exp \left\{ - \lambda_F F_{ik} \right\} \times u(x) \\
			&\propto \exp \left\{ - \frac{F_{ik}^2}{2} \left[ \displaystyle \tau \sum_{j \in \Omega^1_i} \left( \S_k \cdot \G_j \right)^2 \right] \right. \\
			& \left. \hspace{30.5pt} + F_{ik} \left[ - \lambda_F + \tau \sum_{j \in \Omega^1_i} \diffTRIexclK \left( \S_k \cdot \G_j \right) \right] \right\} \times u(x) \\
			&\propto \exp \left\{ - \frac{\tauFik}{2} ( F_{ik} - \muFik )^2 \right\} \times u(x) \\
			&\propto \mathcal{TN} ( F_{ik} | \muFik, \tauFik )
		\end{alignat*}
		\noindent where $ h = \lbrace \lambda_F, \lambda_G, \lambdaS, \alpha, \beta \rbrace $ are the hyperparameters to the model, $u(x)$ is the unit step function, and $ \Omega^1_i = \left\{ j \text{ } \vert (i,j) \in \Omega \right\} $, $ \Omega^2_j = \left\{ i \text{ } \vert (i,j) \in \Omega \right\} $ indicate the observed entries per row and column, respectively. \\
		
		\noindent All the parameter values can be found in Tables \ref{bmf}-\ref{bsmtf}. We use $\otimes$ to denote the outer product, $\boldsymbol I$ for the identity matrix, and $\S_{\cdot,l}$ for the $l$th column of matrix $\S$. \\
		
		\noindent For the similarity matrix factorisation we could have also decided to decompose $ \C = \F \F^T + \E $, without the intermediate matrix $\S$. \cite{Ding2005} includes a good discussion of the benefits to our approach.
		For this decomposition we do not consider diagonal entries of $\C$ (in other words, $ (i,i) \notin \Omega$, $i=1..I $) as this leads to third and fourth order terms in the posteriors and makes Gibbs sampling impossible. See \cite{Zhang2012} for a non-probabilistic approach that does consider these elements, leading to a very complicated optimisation problem. 
		
		
		\begin{table*}[h]
			\caption{Gibbs Samplers for Bayesian Matrix Factorisation (BMF, BNMF, BSNMF).} \label{bmf}
			\begin{center}
				\begin{tabular}{c|l|l}
					{\bf PARAM}	&{\bf UPDATE (GAUSSIAN PRIOR)}	&{\bf UPDATE (EXPONENTIAL PRIOR)} \\
					\hline & & \\
					$ \alpha^* $ 		& $ \displaystyle \alpha + \frac{|\Omega|}{2} $		&	$ \displaystyle \alpha + \frac{|\Omega|}{2} $	\\
					$ \beta^* $			& $ \displaystyle \beta + \frac{1}{2} \sumOmega \diff^2 $	 &	$ \displaystyle \beta + \frac{1}{2} \sumOmega \diff^2 $	\\
					$ \tauFik $			& $ \displaystyle \lambda_F + \tau \sumOmegai G_{jk}^2 $		&	$ \displaystyle \tau \sumOmegai G_{jk}^2 $	\\
					$ \muFik $			& $ \displaystyle \frac{1}{\tauFik} \left( \tau \sumOmegai \diffexclK G_{jk} \right) $ 	&	$ \displaystyle \frac{1}{\tauFik} \left(  - \lambda_F + \tau \sumOmegai \diffexclK G_{jk} \right) $	\\	
					$ \tauGjk $			& $ \displaystyle \lambda_G + \tau \sumOmegaj F_{ik}^2 $ 		&	$ \displaystyle \tau \sumOmegaj F_{ik}^2 $	\\
					$ \muGjl $			& $ \displaystyle \frac{1}{\tauGjl} \left( \tau \sumOmegaj \diffexclK F_{ik} \right) $ 	&	$ \displaystyle \frac{1}{\tauGjl} \left( - \lambda_G + \tau \sumOmegaj \diffexclK F_{ik} \right) $	\\ 
					$ \SigmaFi $		& $ \displaystyle \left( \lambda^F \boldsymbol I + \tau \sumOmegai \G_j \otimes \G_j \right)^{-1} $ & - \\
					$ \muFi $			& $ \displaystyle \SigmaFi \cdot \left( \tau \sumOmegai D_{ij} \G_j \right) $ & -	\\
					$ \SigmaGj $		& $ \displaystyle \left( \lambda^G \boldsymbol I + \tau \sumOmegaj \F_i \otimes \F_i \right)^{-1} $ & -		\\
					$ \muGj $			& $ \displaystyle \SigmaGj \cdot \left( \tau \sumOmegaj D_{ij} \F_i \right) $ & -	\\
					& & \\
					\hline
				\end{tabular}
			\end{center}
		\end{table*}
		
		\begin{landscape}
		\begin{table*}[h]
			\caption{Gibbs Samplers for Bayesian Matrix Tri-Factorisation (BMTF).} \label{bmtf}
			\begin{center}
				\begin{tabular}{c|l|l}
					{\bf PARAM}	&{\bf UPDATE (GAUSSIAN PRIOR)} &{\bf UPDATE (EXPONENTIAL PRIOR)} \\
					\hline & & \\
					$ \alpha^* $ 		& $ \displaystyle \alpha + \frac{|\Omega|}{2} $ & $ \displaystyle \alpha + \frac{|\Omega|}{2} $	\\
					$ \beta^* $			& $ \displaystyle \beta + \frac{1}{2} \sumOmega\diffTRI^2 $ & $ \displaystyle \beta + \frac{1}{2} \sumOmega\diffTRI^2 $	 \\
					$ \tauFik $			& $ \displaystyle \lambda_F + \tau \sumOmegai \left( \S_k \cdot \G_j \right)^2 $	& $ \displaystyle \tau \sumOmegai \left( \S_k \cdot \G_j \right)^2 $	\\
					$ \muFik $			& $ \displaystyle \frac{1}{\tauFik} \left( \tau \sumOmegai \diffTRIexclK \left( \S_k \cdot \G_j \right) \right) $ & $ \displaystyle \frac{1}{\tauFik} \left(  - \lambda_F + \tau \sumOmegai \diffTRIexclK \left( \S_k \cdot \G_j \right) \right) $ \\ 	
					$ \tauSkl $			& $ \displaystyle \lambdaS + \tau \sumOmega F_{ik}^2 G_{jl}^2 $	 & $ \displaystyle \tau \sumOmega F_{ik}^2 G_{jl}^2 $	\\
					$ \muSkl $			& $ \displaystyle \frac{1}{\tauSkl} \left( \tau \sumOmega \diffTRIexclKL F_{ik} G_{jl} \right) $ & $ \displaystyle \frac{1}{\tauSkl} \left( - \lambdaS + \tau \sumOmega \diffTRIexclKL F_{ik} G_{jl} \right) $ \\
					$ \tauGjl $			& $ \displaystyle \lambda_G + \tau \sumOmegaj \left( \F_i \cdot \S_{\cdot,l} \right)^2 $ & $ \displaystyle \tau \sumOmegaj \left( \F_i \cdot \S_{\cdot,l} \right)^2 $ \\
					$ \muGjl $			& $ \displaystyle \frac{1}{\tauGjl} \left( \tau \sumOmegaj \diffTRIexclL \left( \F_i \cdot \S_{\cdot,l} \right) \right) $ & $ \displaystyle \frac{1}{\tauGjl} \left( - \lambda_G + \tau \sumOmegaj \diffTRIexclL \left( \F_i \cdot \S_{\cdot,l} \right) \right) $ \\ 
					$ \SigmaFi $		& $ \displaystyle \left( \lambda^F \boldsymbol I + \tau \sumOmegai (\S \cdot \G_j) \otimes (\S \cdot \G_j) \right)^{-1} $ & - \\
					$ \muFi $			& $ \displaystyle \SigmaFi \cdot \left( \tau \sumOmegai R_{ij} (\S \cdot \G_j) \right) $ & - \\
					$ \SigmaSk $		& $ \displaystyle \left( \lambda^S \boldsymbol I + \tau \sumOmega F_{ik} ( \G_j \otimes \G_j ) \right)^{-1} $ & - \\
					$ \muSk $			& $ \displaystyle \SigmaSk \cdot \left( \tau \sumOmega ( R_{ij} - \sumexclk F_{ik'} ( \S_{k'} \cdot \G_j ) ) F_{ik} \G_j \right) $ & - \\
					$ \SigmaGj $		& $ \displaystyle \left( \lambda^G \boldsymbol I + \tau \sumOmegaj (\F_i \cdot \S) \otimes (\F_i \cdot \S) \right)^{-1} $ & - \\
					$ \muGj $			& $ \displaystyle \SigmaGj \cdot \left( \tau \sumOmegaj R_{ij} (\F_i \cdot \S) \right) $ & - \\	
					& & \\
					\hline
				\end{tabular}
			\end{center}
		\end{table*}
		\end{landscape}
		
		\begin{landscape}
		\begin{table*}[h]
			\caption{Gibbs Samplers for Bayesian Similarity Matrix Tri-Factorisation (BSMTF).}
			\label{bsmtf}
			\begin{center}
				\begin{tabular}{c|l|l}
					{\bf PARAM}	&{\bf UPDATE (GAUSSIAN PRIOR)} &{\bf UPDATE (EXPONENTIAL PRIOR)} \\
					\hline & & \\
					$ \alpha^* $ 		& $ \displaystyle \alpha + \frac{|\Omega|}{2} $	& $ \displaystyle \alpha + \frac{|\Omega|}{2} $	\\
					$ \beta^* $			& $ \displaystyle \beta + \frac{1}{2} \sumOmega\diffsymTRI^2 $	& $ \displaystyle \beta + \frac{1}{2} \sumOmega\diffsymTRI^2 $	 \\
					$ \tauFik $			& $ \displaystyle \lambda_F + \tau \left[ \sumOmegai \left( \S_k \cdot \F_j \right)^2 + \sumOmegajp \left( \F_{i'} \cdot \S_{\cdot,k} \right)^2 \right] $	& $ \displaystyle \tau \left[ \sumOmegai \left( \S_k \cdot \F_j \right)^2 + \sumOmegajp \left( \F_{i'} \cdot \S_{\cdot,k} \right)^2 \right] $		\\
					$ \muFik $			& $ \displaystyle \frac{1}{\tauFik} \left( \tau \sumOmegai \diffsymTRIexclK \left( \S_k \cdot \F_j \right) \right. $	& $ \displaystyle \frac{1}{\tauFik} \left( - \lambda_F + \tau \sumOmegai \diffsymTRIexclK \left( \S_k \cdot \F_j \right) \right. $ \\
					& \hspace{25pt} $ \displaystyle \left. + \tau \sumOmegajp \diffsymTRIexclL \left( \F_{i'} \cdot \S_{\cdot,k} \right) \right) $ 	&	\hspace{43pt} $ \displaystyle \left. + \tau \sumOmegajp \diffsymTRIexclL \left( \F_{i'} \cdot \S_{\cdot,k} \right) \right) $	\\	
					$ \tauSkl $			& $ \displaystyle \lambdaS + \tau \sumOmega F_{ik}^2 F_{jl}^2 $	&	$ \displaystyle \tau \sumOmega F_{ik}^2 F_{jl}^2 $		\\
					$ \muSkl $			& $ \displaystyle \frac{1}{\tauSkl} \left( \tau \sumOmega \diffsymTRIexclKL F_{ik} F_{jl} \right) $	& $ \displaystyle \frac{1}{\tauSkl} \left( - \lambdaS + \tau \sumOmega \diffsymTRIexclKL F_{ik} F_{jl} \right) $ \\ 
					$ \SigmaFi $		& $ \displaystyle \left( \lambda^F \boldsymbol I + \tau \left[ \sumOmegai (\S \cdot \F_j) \otimes (\S \cdot \F_j) + \sumOmegajp (\F_{i'} \cdot \S) \otimes (\F_{i'} \cdot \S) \right] \right)^{-1} $ & - \\
					$ \muFi $			& $ \displaystyle \SigmaFi \cdot \left( \tau \sumOmegai C_{ij} (\S \cdot \F_j) + \tau \sumOmegajp C_{i'i} (\F_{i'} \cdot \S) \right) $ & - \\
					$ \SigmaSk $		& $ \displaystyle \left( \lambda^S \boldsymbol I + \tau \sumOmega F_{ik} ( \F_j \otimes \F_j ) \right)^{-1} $ & - \\
					$ \muSk $			& $ \displaystyle \SigmaSk \cdot \left( \tau \sumOmega ( C_{ij} - \sumexclk F_{ik'} ( \S_{k'} \cdot \F_j ) ) F_{ik} \F_j \right) $ & - \\
					& & \\
					\hline
				\end{tabular}
			\end{center}
		\end{table*}
		\end{landscape}


	\subsection{Matrix factorisation with ARD and importance values} \label{Matrix factorisation with ARD}
		We will now explain how to extend matrix factorisation with Automatic Relevance Determination (ARD) and importance values. 
		
		\paragraph{ARD} 
		We change the model definition to the following. 
		\begin{alignat*}{2}
			D_{ij} &\sim \mathcal{N} (D_{ij} | \boldsymbol F_i \cdot \boldsymbol G_j, \tau^{-1} ) \\
			F_{ik} &\sim \mathcal{N} ( F_{ik} | 0, (\lambda_k)^{-1} )  \quad\quad &\text{ or }  \quad\quad \mathcal{E} ( F_{ik} | \lambda_k ) \\
			G_{jk} &\sim \mathcal{N} ( G_{jk} | 0, (\lambda_k)^{-1} )  \quad\quad &\text{ or }  \quad\quad \mathcal{E} ( G_{jk} | \lambda_k ) \\
			\tau &\sim \mathcal{G} (\tau | \alpha, \beta )
		\end{alignat*}
		Notice that the main difference is replacing $\lambda_F$ and $\lambda_G$ by the parameter $\lambda_k$, for $k=1..K$. We place a Gamma prior over these variables,
		\begin{equation*}
			\lambda_k \sim \mathcal{G} (\lambda_k | \alpha_0, \beta_0 ).
		\end{equation*}
		This can similarly be done for matrix tri-factorisation, by placing one ARD over $\F$ (using $\lambda^F_k$) and another over $\G$ (using $\lambda^G_l$). \\
		
		\noindent The Gibbs sampling algorithms remain largely the same, simply replacing $\lambda_F, \lambda_G$ in the updates by $\lambda_k$. For the multivariate posteriors, we replace $\lambda^F \boldsymbol I$ by $\text{diag}(\boldsymbol \lambda)$, a diagonal matrix where the $k$th diagonal element is given by $\lambda_k$. The posterior Gibbs sampling distribution for $\lambda_k$ itself can be derived to be another Gamma distribution,
		\begin{equation*}
			p(\lambda_k | \D, \F, \G, \tau ) = \mathcal{G} (\lambda_k | \alpha^*_0, \beta^*_0 )
		\end{equation*}
		where
		\begin{align*}
			\alpha^*_0 = \alpha_0 + \frac{I}{2} + \frac{J}{2} 	\quad\quad\quad\quad 	\beta^*_0 = \beta_0 + \frac{1}{2} \sum_{i=1}^I F_{ik}^2 + \frac{1}{2} \sum_{j=1}^J G_{jk}^2
		\end{align*}
		for the real-valued (Gaussian prior) model, and
		\begin{align*}
			\alpha^*_0 = \alpha_0 + I + J 	\quad\quad\quad\quad 	\beta^*_0 = \beta_0 + \sum_{i=1}^I F_{ik} + \sum_{j=1}^J G_{jk}
		\end{align*}
		for the nonnegative (exponential prior) model. Note that each $\lambda_k$ therefore depends only on the values in the $k$th column of $\F$ and $\G$, and those values in turn depend on the value of $\lambda_k$: if most values in that column are high, $\lambda_k$ gets a small value (indicating that the factor is active); and if $\lambda_k$ has a high value, the $k$th column will get a Gibbs sampling posterior around 0, resulting in low values (pushing the other values for this factor down, since it is inactive).
		
		\paragraph{Importance value}
		As discussed in the paper, we modify the likelihood with an importances values $\alpha^n, \alpha^l, \alpha^m$ as follows.
		\begin{alignat*}{1}
			p( \boldsymbol \theta | \R, \D, \C ) \propto & \prod_{t=1}^T p( \F^t | \boldsymbol \lambda^t ) \prod_{n=1}^N p(\tau^n) \prod_{l=1}^L p(\tau^l) \prod_{m=1}^M p(\tau^m) \\
			\times & \prod_{n=1}^N p(\R^n | \F^{t_n}, \S^n, \F^{u_n}, \tau^n )^{\alpha^n} \\
			\times & \prod_{l=1}^L p(\D^l | \F^{t_l}, \G^l, \tau^l )^{\alpha^l} \\
			\times & \prod_{m=1}^M p(\C^m | \F^{t_m}, \S^m, \tau^m )^{\alpha^m} 
		\end{alignat*}
		where $\boldsymbol \theta$ is the set of model parameters ($\F^t, \S^n, \S^m, \G^l, \boldsymbol \lambda^t, \tau^n, \tau^l, \tau^m$). We effectively repeat the dataset $\R^n$ $\alpha^n$ times, and similarly for $\D^l, \C^m$, requiring a better fit to the data. The Gibbs samplers remain largely the same, with the addition of several $\alpha$ values in the updates. This is illustrated below for a single dataset $\D \approx \F \cdot \G^T$ with nonnegative factors, importance value $\alpha$, and without ARD.
		\begin{alignat*}{2}
			&\tau \sim \mathcal{G} (\tau | \alpha_*, \beta_* ) 	\quad\quad	 &&\alpha_* = \alpha_{\tau} + \alpha \frac{|\Omega^|}{2} 	\\
			& && \beta_* = \beta_{\tau} +  \alpha \frac{1}{2} \sum_{(i,j) \in \Omega^l} (D_{ij} - \F_i \cdot \G_j)^2 \\
			& F_{ik} \sim \mathcal{TN} (F_{ik} | \mu_{ik}, \tau_{ik} ) \quad\quad	 && \mu_{ik} = \frac{1}{\tau_{ik}} \left( - \lambda_F + \alpha \tau \sum_{j \in \Omega^1_i} ( D_{ij} - \sum_{k' \neq k} F_{ik'} G_{jk'} ) G_{jk} \right)	\\
			& && \tau_{ik} = \alpha \tau \sum_{j \in \Omega^1_i} G_{jk}^2 \\
			& G_{jk} \sim \mathcal{TN} (G_{jk} | \mu_{jk}, \tau_{jk} ) \quad\quad	 && \mu_{jk} = \frac{1}{\tau_{jk}} \left( - \lambda_G + \alpha \tau \sum_{i \in \Omega^2_j} ( D_{ij} - \sum_{k' \neq k} F_{ik'} G_{jk'} ) F_{ik} \right)	\\
			& && \tau_{jk} = \alpha \tau \sum_{i \in \Omega^2_j} (F_{ik})^2
		\end{alignat*}
		Similar derivations can be done for matrix tri-factorisation, and real-valued versions.

	\clearpage
	\subsection{Hybrid matrix factorisation model} \label{Hybrid matrix factorisation model}
		The hybrid matrix factorisation (HMF) model combines all the ideas presented in the previous two sections. Recall we are given three types of datasets:
		\begin{enumerate}
			\item Main datasets $ \R = \lbrace \R^1, .., \R^N \rbrace $, relating two different entity types. 
			Each dataset $\R^n \in \mathbb{R}^{I_{t_n} \times I_{u_n}} $ relates entity types $E_{t_n}$, $E_{u_n}$. We use matrix tri-factorisation to decompose it into two entity type factor matrices $\F^{t_n}, \F^{u_n}$, and a dataset-specific matrix $ \S^n \in \mathbb{R}^{K_{t_n} \times K_{u_n}} $.
			\begin{equation*}
				\R^n = \F^{t_n} \S^n (\F^{u_n})^T + \E^n.
			\end{equation*}
			\item Feature datasets $ \D = \lbrace \D^1, .., \D^L \rbrace $, representing features for an entity type. 
			Each dataset $\D^l \in \mathbb{R}^{I_{t_l} \times J_l} $ relates an entity type $E_{t_l}$ to $J_l$ features. We use matrix factorisation to decompose it into one entity type factor matrix $\F^{t_l}$, and a dataset-specific matrix $ \G^l \in \mathbb{R}^{J_l \times K_{t_l}} $.
			\begin{equation*}
				\D^l = \F^{t_l} (\G^l)^T + \E^l.
			\end{equation*}
			\item Similarity datasets $ \C = \lbrace \C^1, .., \C^M \rbrace $, giving similarities between entities of the same entity type. 
			Each dataset $ \C^m \in \mathbb{R}^{I_{t_m} \times I_{t_m}} $ relates an entity type $ E_{t_m} $ to itself. We use matrix tri-factorisation to decompose it into a entity type factor matrix $\F^{t_m}$, a dataset-specific matrix $ \S^m \in \mathbb{R}^{K_{t_m} \times K_{t_m}} $, and $\F^{t_m}$ again.
			\begin{equation*}
				\C^m = \F^{t_m} \S^m (\F^{t_m})^T + \E^m.
			\end{equation*}
		\end{enumerate}
		Observed entries are given by the sets $ \Omega^n = \lbrace (i,j) \vert \text{ } R^n_{ij} \text{ observed} \rbrace $, $ \Omega^l = \lbrace (i,j) \vert \text{ } D^l_{ij} \text{ observed} \rbrace $, $ \Omega^m = \lbrace (i,j) \vert \text{ } C^m_{ij} \text{ observed} \rbrace $, respectively.
			
		\subsubsection{Model definition}
			\noindent The model likelihood functions are
			\begin{align*}
				& R^n_{ij} \sim \mathcal{N} (R^n_{ij} | \F^{t_n}_i \cdot \S^n \cdot \F^{s_n}_j, (\tau^n)^{-1} )	\\
				& D^m_{ij} \sim \mathcal{N} (D^l_{ij} | \F^{t_l}_i \cdot \G^{l}_j, (\tau^l)^{-1} ) 		\\
				& C^m_{ij} \sim \mathcal{N} (C^m_{ij} | \F^{t_m}_i \cdot \S^m \cdot \F^{t_m}_j, (\tau^m)^{-1} ),
			\end{align*}
			with Bayesian priors
			\begin{align*}
			 	& \tau^n, \tau^l, \tau^m \sim \mathcal{G} (\tau^* | \alpha_{\tau}, \beta_{\tau} ) \\
				& F^t_{ik} \sim \mathcal{E} ( F^t_{ik} | \lambda_k^t) && \text{ or } \quad 		F^t_{ik} \sim \mathcal{N} ( F^t_{ik} | 0, (\lambda_k^t)^{-1}) \\
				& G^l_{jk} \sim \mathcal{E}( G^l_{jl} | \lambda_k^{t_l}) && \text{ or } \quad 		G^l_{jk} \sim \mathcal{N}( G^l_{jl} | 0, (\lambda_k^{t_l})^{-1} ) \\
				& S^n_{kl} \sim \mathcal{E}( S^n_{kl} | \lambdaS^n) && \text{ or } \quad		 S^n_{kl} \sim \mathcal{N}( S^n_{kl} | 0, (\lambdaS^n)^{-1}) \\
				& S^m_{kl} \sim \mathcal{E}( S^m_{kl} | \lambdaS^m) && \text{ or } \quad		 S^m_{kl} \sim \mathcal{N}( S^m_{kl} | 0, (\lambdaS^m)^{-1}). \\
				& \lambda_k^t \sim \mathcal{G} (\lambda_k^t | \alpha_0, \beta_0 ).
			\end{align*}
			Finally, we add an importance value for each of the $\R^n, \D^l, \C^m$ datasets, respectively $\alpha_n, \alpha_l, \alpha_m$.
			
		\subsubsection{Gibbs sampler}
			The Gibbs sampling algorithm has updates that combine the parameter values given in Tables \ref{bmf}, \ref{bmtf}, and \ref{bsmtf}, for the single-dataset matrix factorisations, matrix tri-factorisations, and similarity matrix tri-factorisations. Because there are so many different parts of the models involved, careful notational definition is essential. \\
			
			\noindent The datasets relating a given entity type $E_t$ are indicated by the following sets,
			\begin{alignat*}{1}
				U_1^t &= \left\{ n \text{ } \vert \text{ } \R^n \in \R \text{ } \land t_n = t \text{ } \right\} \\
				U_2^t &= \left\{ n \text{ } \vert \text{ } \R^n \in \R \text{ } \land u_n = t \text{ } \right\} \\
				V^t &= \left\{ l \text{ } \vert \text{ } \D^l \in \D \text{ } \land t_l = t \text{ } \right\} \\
				W^t &= \left\{ m \text{ } \vert \text{ } \C^m \in \C \text{ } \land t_m = t \text{ } \right\}.
			\end{alignat*}
			Since the updates for the ARD can be different if a feature dataset is decomposed using negative factors or real-valued factors for $\G^l$, we also introduce the sets
			\begin{align*}
				V^t_+ = \left\{ l \in V^t \text{ } \vert \text{ $\G^l$ is nonnegative} \right\}		\quad\quad		V^t_- = \left\{ l \in V^t \text{ } \vert \text{ $\G^l$ is real-valued} \right\}.
			\end{align*}
			Observed entries per row $i$ and column $j$ are given by 
			\begin{alignat*}{2}
				& \Omega^{n1}_i = \left\{ j \text{ } \vert (i,j) \in \Omega^n \right\}  \quad\quad
				&& \Omega^{n2}_j = \left\{ i \text{ } \vert (i,j) \in \Omega^n \right\} \\
				& \Omega^{l1}_i = \left\{ j \text{ } \vert (i,j) \in \Omega^l \right\}  \quad\quad
				&& \Omega^{l2}_j = \left\{ i \text{ } \vert (i,j) \in \Omega^l \right\} \\
				& \Omega^{m1}_i = \left\{ j \text{ } \vert (i,j) \in \Omega^m \right\}  \quad\quad
				&& \Omega^{m2}_j = \left\{ i \text{ } \vert (i,j) \in \Omega^m \right\}.
			\end{alignat*}
			We obtain the following posterior distributions and parameter values:
			
			\paragraph{Noise parameters}
			\begin{alignat*}{3}
				&\tau^n \sim \mathcal{G} (\tau^n | \alpha^n_*, \beta^n_* ) 	\quad\quad	 &&\alpha^n_* = \alpha_{\tau} + \alpha^n \frac{|\Omega^n|}{2} 	\quad\quad	&&\beta^n_* = \beta_{\tau} + \alpha^n \frac{1}{2} \sum_{(i,j) \in \Omega^n} (R^n_{ij} - \F^{t_n}_i \cdot \S^n \cdot \F^{u_n}_j)^2 \\
				&\tau^l \sim \mathcal{G} (\tau^l | \alpha^l_*, \beta^l_* ) 	\quad\quad	 &&\alpha^l_* = \alpha_{\tau} + \alpha^l \frac{|\Omega^l|}{2} 	\quad\quad	&&\beta^l_* = \beta_{\tau} + \alpha^l \frac{1}{2} \sum_{(i,j) \in \Omega^l} (D^l_{ij} - \F^{t_l}_i \cdot \G^l_j)^2 \\
				&\tau^m \sim \mathcal{G} (\tau^m | \alpha^m_*, \beta^m_* ) 	\quad\quad	 &&\alpha^m_* = \alpha_{\tau} + \alpha^m \frac{|\Omega^m|}{2} 	\quad\quad	&&\beta^m_* = \beta_{\tau} + \alpha^m \frac{1}{2} \sum_{(i,j) \in \Omega^m} (C^m_{ij} - \F^{t_m}_i \cdot \S^m \cdot \F^{t_m}_j)^2
			\end{alignat*}
			
			\paragraph{ARD} If $\F^t$ contains nonnegative factors:
			\begin{alignat*}{2}
				&\lambda^t_k \sim \mathcal{G} (\lambda^t_k | \alpha^t_k, \beta^t_k ) \quad\quad	 &&\alpha^t_k = \alpha_0 + I_t + \sum_{l \in V^t_+} I_{t_l} + \sum_{l \in V^t_-} \frac{I_{t_l}}{2}	\\
				& &&\beta^t_k = \beta_0 + \sum_{i=1}^{I_t} F_{ik} + \sum_{l \in V^t_+} \sum_{j=1}^{J_l} G_{jk} + \sum_{l \in V^t_-} \frac{1}{2} \sum_{j=1}^{J_l} G_{jk}^2.
			\end{alignat*}
			If $\F^t$ contains real-valued factors:
			\begin{alignat*}{2}
				&\lambda^t_k \sim \mathcal{G} (\lambda^t_k | \alpha^t_k, \beta^t_k ) \quad\quad	 &&\alpha^t_k = \alpha_0 + \frac{I_t}{2} + \sum_{l \in V^t_+} I_{t_l} + \sum_{l \in V^t_-} \frac{I_{t_l}}{2}	\\
				& &&\beta^t_k = \beta_0 + \frac{1}{2} \sum_{i=1}^{I_t} F_{ik}^2 + \sum_{l \in V^t_+} \sum_{j=1}^{J_l} G_{jk} + \sum_{l \in V^t_-} \frac{1}{2} \sum_{j=1}^{J_l} G_{jk}^2.
			\end{alignat*}
			
			\paragraph{Dataset-specific factor matrices} If $\G^l, \S^n, \S^m$ contain nonnegative factors:
			\begin{alignat*}{2}
				& G^l_{jk} \sim \mathcal{TN} (G^l_{jk} | \mu^l_{jk}, \tau^l_{jk} ) \quad\quad	 && \mu^l_{jk} = \frac{1}{\tau^l_{jk}} \left( - \lambda^{t_l}_k + \tau^l \alpha^l \sum_{i \in \Omega^{l2}_j} ( D^l_{ij} - \sum_{k' \neq k} F^{t_l}_{ik'} G^l_{jk'} ) F^{t_l}_{ik} \right)	\\
				& && \tau^l_{jk} = \tau^l \alpha^l \sum_{i \in \Omega^{l2}_j} (F^{t_l}_{ik})^2 \\
				& S^n_{kl} \sim \mathcal{TN} (S^n_{kl} | \mu^n_{kl}, \tau^n_{kl} ) \quad\quad	 && \mu^n_{kl} = \frac{1}{\tau^n_{kl}} \left( - \lambda^n_S + \tau^n \alpha^n \sum_{(i,j) \in \Omega^n} ( R^n_{ij} - \sumexclkl F^{t_n}_{ik'} S^n_{k'l'} F^{u_n}_{jl'} ) F^{t_n}_{ik} F^{u_n}_{jl} \right)	\\
				& && \tau^n_{kl} = \tau^n \alpha^n \sum_{(i,j) \in \Omega^n} (F^{t_n}_{ik})^2 (F^{u_n}_{jl})^2 \\
				& S^m_{kl} \sim \mathcal{TN} (S^m_{kl} | \mu^m_{kl}, \tau^m_{kl} ) \quad\quad	 && \mu^m_{kl} = \frac{1}{\tau^m_{kl}} \left( - \lambda^m_S + \tau^m \alpha^m \sum_{(i,j) \in \Omega^m} ( C^m_{ij} - \sumexclkl F^{t_m}_{ik'} S^m_{k'l'} F^{t_m}_{jl'} ) F^{t_m}_{ik} F^{t_m}_{jl} \right)	\\
				& && \tau^m_{kl} = \tau^m \alpha^m \sum_{(i,j) \in \Omega^m} (F^{t_m}_{ik})^2 (F^{t_m}_{jl})^2
			\end{alignat*}
			If $\G^l, \S^n, \S^m$ contain real-valued factors:
			\begin{alignat*}{2}
				& G^l_{jk} \sim \mathcal{N} (G^l_{jk} | \mu^l_{jk}, (\tau^l_{jk})^{-1} ) \quad\quad	 && \mu^l_{jk} = \frac{1}{\tau^l_{jk}} \left( \tau^l \alpha^l \sum_{i \in \Omega^{l2}_j} ( D^l_{ij} - \sum_{k' \neq k} F^{t_l}_{ik'} G^l_{jk'} ) F^{t_l}_{ik} \right)	\\
				& && \tau^l_{jk} = \lambda^{t_l}_k + \tau^l \alpha^l \sum_{i \in \Omega^{l2}_j} (F^{t_l}_{ik})^2 \\
				& \G^l_{j} \sim \mathcal{N} (\G^l_{j} | \boldsymbol \mu^l_{j}, \boldsymbol \Sigma^l_j ) \quad\quad	 && \boldsymbol \mu^l_{j} = \boldsymbol \Sigma^l_j \cdot \left( \tau^l \alpha^l \sum_{i \in \Omega^{l2}_j} D^l_{ij} \F^{t_l}_i \right)	\\
				& && \boldsymbol \Sigma^l_j = \left( \text{diag}(\boldsymbol \lambda^t) + \tau^l \alpha^l \sum_{i \in \Omega^{l2}_j} \F^{t_l}_i \otimes \F^{t_l}_i \right)^{-1} \\
				& S^n_{kl} \sim \mathcal{N} (S^n_{kl} | \mu^n_{kl}, (\tau^n_{kl})^{-1} ) \quad\quad	 && \mu^n_{kl} = \frac{1}{\tau^n_{kl}} \left( \tau^n \alpha^n \sum_{(i,j) \in \Omega^n} ( R^n_{ij} - \sumexclkl F^{t_n}_{ik'} S^n_{k'l'} F^{u_n}_{jl'} ) F^{t_n}_{ik} F^{u_n}_{jl} \right)	\\
				& && \tau^n_{jk} = \lambda^n_S + \tau^n \alpha^n \sum_{(i,j) \in \Omega^n} (F^{t_n}_{ik})^2 (F^{u_n}_{jl})^2 \\
				& \S^n_k \sim \mathcal{N} (\S^n_k | \boldsymbol \mu^n_k, \boldsymbol \Sigma^n_k ) \quad\quad	 && \boldsymbol \mu^n_k = \boldsymbol \Sigma^n_k \cdot \left( \tau^n \alpha^n \sum_{(i,j) \in \Omega^n} ( R^n_{ij} - \sumexclk F^{t_n}_{ik'} ( \S^n_{k'} \cdot \F^{u_n}_j ) ) F^{t_n}_{ik} \F^{u_n}_j \right)	\\
				& && \boldsymbol \Sigma^n_k = \left( \lambda_S^n \boldsymbol I + \tau^n \alpha^n \sum_{(i,j) \in \Omega^n} F^{t_n}_{ik} ( \F^{u_n}_j \otimes \F^{u_n}_j ) \right)^{-1} \\
				& S^m_{kl} \sim \mathcal{N} (S^m_{kl} | \mu^m_{kl}, (\tau^m_{kl})^{-1} ) \quad\quad	 && \mu^m_{kl} = \frac{1}{\tau^m_{kl}} \left( \tau^m \alpha^m \sum_{(i,j) \in \Omega^m} ( C^m_{ij} - \sumexclkl F^{t_m}_{ik'} S^m_{k'l'} F^{t_m}_{jl'} ) F^{t_m}_{ik} F^{t_m}_{jl} \right)	\\
				& && \tau^m_{kl} = \lambda^m_S + \tau^m \alpha^m \sum_{(i,j) \in \Omega^m} (F^{t_m}_{ik})^2 (F^{t_m}_{jl})^2 \\
				& \S^m_k \sim \mathcal{N} (\S^m_k | \boldsymbol \mu^m_k, \boldsymbol \Sigma^m_k ) \quad\quad	 && \boldsymbol \mu^m_k = \boldsymbol \Sigma^m_k \cdot \left( \tau^m \alpha^m \sum_{(i,j) \in \Omega^m} ( C^m_{ij} - \sumexclk F^{t_m}_{ik'} ( \S^m_{k'} \cdot \F^{t_m}_j ) ) F^{t_m}_{ik} \F^{t_m}_j \right)	\\
				& && \boldsymbol \Sigma^m_k = \left( \lambda^m_S \boldsymbol I + \tau^m \alpha^m \sum_{(i,j) \in \Omega^m} F^{t_m}_{ik} ( \F^{t_m}_j \otimes \F^{t_m}_j ) \right)^{-1} \\
			\end{alignat*}
	
			\paragraph{Shared factor matrices} If $\F^t$ contains nonnegative factors:
			\begin{alignat*}{3}
				& F^t_{ik} \sim \mathcal{TN} (F^t_{ik} | \mu^t_{ik}, \tau^t_{ik} ) \quad\quad	 && \mu^t_{ik} = && \frac{1}{\tau^t_{ik}} \left( - \lambda^t_k + \sum_{n \in U^t_1} \tau^n \alpha^n \sum_{j \in \Omega^{n1}_i} ( R^n_{ij} - \sumexclk F^t_{ik'} (\S^n_{k'} \cdot \F^{u_n}_j) ) (\S^n_{k} \cdot \F^{u_n}_j) \right. \\
				& && && \hspace{49pt} \left. + \sum_{n \in U^t_2} \tau^n \alpha^n \sum_{i' \in \Omega^{n2}_i} ( R^n_{i'i} - \sum_{l \neq k} F^t_{il} ( \F^{t_n}_{i'} \cdot \S^n_{.,l} ) ) ( \F^{t_n}_{i'} \cdot \S^n_{.,k} ) \right. \\
				& && && \hspace{49pt} \left. + \sum_{l \in V^t} \tau^l \alpha^l \sum_{j \in \Omega^{l1}_i} ( D^l_{ij} - \sumexclk F^t_{ik'} G^l_{jk'} ) G^l_{jk} \right) \\
				& && && \hspace{49pt} \left. + \sum_{m \in W^t} \tau^m \alpha^m \left[ \sum_{j \in \Omega^{m1}_i} ( C^m_{ij} - \sumexclk F^t_{ik'} (\S^m_{k'} \cdot \F^t_j) ) (\S^m_{k} \cdot \F^t_j) \right. \right. \\
				& && && \hspace{129pt} \left. \left. + \sum_{i' \in \Omega^{m2}_i} ( C^m_{i'i} - \sum_{l \neq k} F^t_{il} ( \F^t_{i'} \cdot \S^m_{.,l} ) ) ( \F^t_{i'} \cdot \S^m_{.,k} ) \right] \right) \\
				& && \tau^t_{ik} = && \sum_{n \in U^t_1} \tau^n \alpha^n \sum_{j \in \Omega^{n1}_i} (\S^n_k \cdot \F^{u_n}_j)^2 + \sum_{n \in U^t_2} \tau^n \alpha^n \sum_{i' \in \Omega^{n2}_i} (\F^{t_n}_{i'} \cdot \S^n_{.,k})^2	\\
				& && && + \sum_{l \in V^t} \tau^l \alpha^l \sum_{j \in \Omega^{l1}_i} (G^l_{jk})^2 \\
				& && && + \sum_{m \in W^t} \tau^m \alpha^m \left[ \sum_{j \in \Omega^{m1}_i} (\S^m_k \cdot \F^{t}_j)^2 + \sum_{i' \in \Omega^{m2}_i} (\F^{t}_{i'} \cdot \S^m_{.,k})^2 \right]
			\end{alignat*}
			\noindent If $\F^t$ contains real-valued factors:
			\begin{alignat*}{3}
				& F^t_{ik} \sim \mathcal{N} (F^t_{ik} | \mu^t_{ik}, (\tau^t_{ik})^{-1} ) \quad\quad	 && \mu^t_{ik} = && \frac{1}{\tau^t_{ik}} \left( \sum_{n \in U^t_1} \tau^n \alpha^n \sum_{j \in \Omega^{n1}_i} ( R^n_{ij} - \sumexclk F^t_{ik'} (\S^n_{k'} \cdot \F^{u_n}_j) ) (\S^n_{k} \cdot \F^{u_n}_j) \right. \\
				& && && \hspace{29pt} \left. + \sum_{n \in U^t_2} \tau^n \alpha^n \sum_{i' \in \Omega^{n2}_i} ( R^n_{i'i} - \sum_{l \neq k} F^t_{il} ( \F^{t_n}_{i'} \cdot \S^n_{.,l} ) ) ( \F^{t_n}_{i'} \cdot \S^n_{.,k} ) \right. \\
				& && && \hspace{29pt} \left. + \sum_{l \in V^t} \tau^l \alpha^l \sum_{j \in \Omega^{l1}_i} ( D^l_{ij} - \sumexclk F^t_{ik'} G^l_{jk'} ) G^l_{jk} \right. \\
				& && && \hspace{29pt} \left. + \sum_{m \in W^t} \tau^m \alpha^m \left[ \sum_{j \in \Omega^{m1}_i} ( C^m_{ij} - \sumexclk F^t_{ik'} (\S^m_{k'} \cdot \F^t_j) ) (\S^m_{k} \cdot \F^t_j) \right. \right. \\
				& && && \hspace{109pt} \left. \left. + \sum_{i' \in \Omega^{m2}_i} ( C^m_{i'i} - \sum_{l \neq k} F^t_{il} ( \F^t_{i'} \cdot \S^m_{.,l} ) ) ( \F^t_{i'} \cdot \S^m_{.,k} ) \right] \right) \\ \\
				& && \tau^t_{ik} = && \lambda_k^t + \sum_{n \in U^t_1} \tau^n \alpha^n \sum_{j \in \Omega^{n1}_i} (\S^n_k \cdot \F^{u_n}_j)^2 + \sum_{n \in U^t_2} \tau^n \alpha^n \sum_{i' \in \Omega^{n2}_i} (\F^{t_n}_{i'} \cdot \S^n_{.,k})^2	\\
				& && && + \sum_{l \in V^t} \tau^l \alpha^l \sum_{j \in \Omega^{l1}_i} (G^l_{jk})^2 \\
				& && && + \sum_{m \in W^t} \tau^m \alpha^m \left[ \sum_{j \in \Omega^{m1}_i} (\S^m_k \cdot \F^{t}_j)^2 + \sum_{i' \in \Omega^{m2}_i} (\F^{t}_{i'} \cdot \S^m_{.,k})^2 \right]
			\end{alignat*}
			\begin{alignat*}{3}
				& \F^t_i \sim \mathcal{N} (\F^t_i | \boldsymbol \mu^t_i, \boldsymbol \Sigma^t_{i} ) \quad\quad	 && \boldsymbol \mu^t_i = && \boldsymbol \Sigma^t_{i} \cdot \left( \sum_{n \in U^t_1} \tau^n \alpha^n \sum_{j \in \Omega^{n1}_i} R^n_{ij} (\S^n \cdot \F^{u_n}_j) + \sum_{n \in U^t_2} \tau^n \alpha^n \sum_{i' \in \Omega^{n2}_i} R^n_{i'i} (\F^{t_n}_{i'} \cdot \S^n) \right. \\
				& && && \hspace{30pt} \left. + \sum_{l \in V^t} \tau^l \alpha^l \sum_{j \in \Omega^{l1}_i} D^l_{ij} \G^l_j \right. \\
				& && && \hspace{30pt} \left. + \sum_{m \in W^t} \tau^m \alpha^m \left[ \sum_{j \in \Omega^{m1}_i} C^m_{ij} (\S^m \cdot \F^t_j) + \sum_{i' \in \Omega^{l2}_i} C^m_{i'i} (\F^t_{i'} \cdot \S^m) \right] \right) \\
				& && \boldsymbol \Sigma^t_{i} = && \left( \text{diag}(\boldsymbol \lambda^t_k) + \sum_{n \in U^t_1} \tau^n \alpha^n \sum_{j \in \Omega^{n1}_i} (\S^n \cdot \F^{u_n}_j) \otimes (\S^n \cdot \F^{u_n}_j) \right. \\
				& && && \hspace{56pt} \left. + \sum_{n \in U^t_2} \tau^n \alpha^n \sum_{i' \in \Omega^{n2}_i} (\F^{t_n}_{i'} \cdot \S^n) \otimes (\F^{t_n}_{i'} \cdot \S^n) \right. \\
				& && && \hspace{56pt} \left. + \sum_{l \in V^t} \tau^l \alpha^l \sum_{j \in \Omega^{l1}_i} (\G^l_j \otimes \G^l_j) \right. \\
				& && && \hspace{56pt} \left. + \sum_{m \in W^t} \tau^m \alpha^m \left[ \sum_{j \in \Omega^{m1}_i} (\S^m \cdot \F^t_j) \otimes (\S^m \cdot \F^t_j) \right. \right. \\
				& && && \hspace{136pt} \left. \left. + \sum_{i' \in \Omega^{m2}_i} (\F^t_{i'} \cdot \S^m) \otimes (\F^t_{i'} \cdot \S^m) \right] \right)^{-1}
			\end{alignat*}


\clearpage
\section{Model discussion}
	\subsection{Software}
	We have provided an open-source Python implementation of all models discussed in the paper, available at \url{https://github.com/ThomasBrouwer/HMF}. We furthermore provide all datasets, preprocessing scripts, and Python code for the experiments. Please refer to the README in the Github project.
	
	\subsection{Complexity}
	The updates for the Gibbs sampler for Bayesian matrix tri-factorisation have time complexity $ \mathcal{O}( I J K^2 L ) $, compared to $ \mathcal{O}( I J K^2 ) $ for Bayesian matrix factorisation. For HMF the complexity becomes $ \mathcal{O}( (N + M + L) I^2 K^3 ) $ where $ I = \max_t I_t$ and $ K = \max_t K_t $. Notice that \textbf{our model scales linearly in the number of observed datasets}. Furthermore, the random draws for columns of the factor matrices are independent of each other, and therefore the parameter updates can be formulated as efficient joint matrix operations and new values drawn in parallel. Alternatively, the draws can be done per row by using a multivariate posterior, and then all these row-wise draws can be done in parallel as well.

	\subsection{Missing values and predictions}
	Missing values can be indicated to the model through the mask sets $\Omega^n, \Omega^l, \Omega^m$. Note that this also means that if specific feature values are missing for one of the entities, these features can still be included for the other entities, simply by marking them as unobserved when we do not know their value. This is much better than imputing those values, for example using the row or column average, as the model will still fit to those imputed values. \\
	
	\noindent The missing values can then be predicted, by using the posterior draws of the Gibbs sampler (after burn-in and thinning) to estimate the posteriors of the factor matrices. For example, if we wish to predict missing values in the matrix $\D^l$, we estimate $\F^{t_l}$ and $\G^l$, and predictions for the missing entries are given by $\F^{t_l} \cdot (\G^l)^T$.
	
	\subsection{Initialisation} \label{Initialisation}
	Gibbs sampling can easily get stuck in a local minimum of posterior likelihood, and therefore initialisation of the random variables is essential to obtain a good solution. 
	There are two obvious ways to do this. Since the user specifies the values of the hyperparameters, $\alpha_0$, $\beta_0$, $\alpha_{\tau}$, $\beta_{\tau}$, $\lambdaS^n$, $\lambdaS^m$, we can use the model definition to initialise the variables $\F^t$, $\G^l$, $\S^n$, $\S^m$, $\tau^n$, $\tau^l$, $\tau^m$, $\lambda^t_k$ either using the expectation of the prior model distribution, or by randomly drawing their value according to that distribution. \\
	
	\noindent Alternatively, we can initialise the entity type factor matrices $\F^t$ using $K$-means clustering, as suggested by \cite{Ding2006a}, and initialise the dataset-specific matrices $\S^n, \S^m, \F^l$ using least squares. This can be done using the Moore-Penrose pseudo-inverse ($^+$), as long as the dataset-specific matrices ($\S^n, \S^m, \G^l$) are real-valued. For example,
	\begin{equation*}
		\S^n = ( \F^{t_n} )^+ \cdot \R^n \cdot ( ( \F^{u_n} )^T )^+.
	\end{equation*}
	If the datasets are not real-valued, we can still initialise $\S^n$ or the other factor matrices in this way, but then set all values below zero to zero.
	We measure the effectiveness of the different initialisation methods in Section \ref{Initialisation experiment}, which shows that this combination of $K$-means and least squares initialisation generally gives the fastest convergence.
	
	\subsection{Relation to tensor decomposition}
\begin{figure}[b!]
		\captionsetup{width=1\columnwidth}
		\includegraphics[width=\columnwidth]{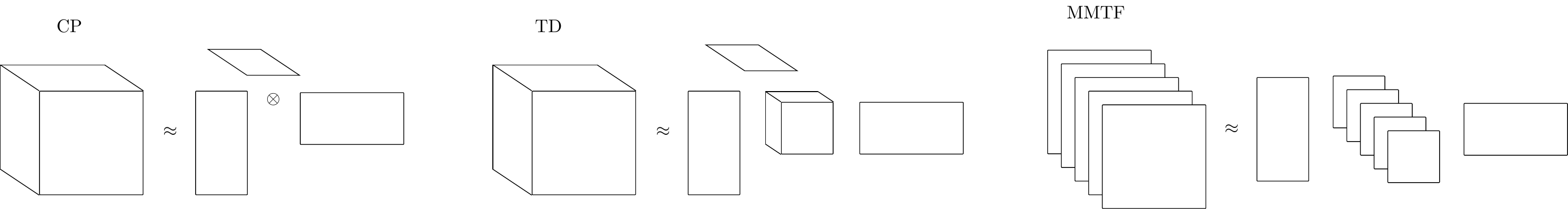}
		\caption{Overview of the CANDECOMP/PARAFAC (CP, left), Tucker Decomposition (TD, middle), and multiple matrix tri-factorisation (MMTF, right) methods. CP uses the outer product ($\otimes$), whereas TD and MMTF use the matrix product.}
		\label{cp_td_mmtf}
	\end{figure}

	Multiple matrix factorisation and tri-factorisation methods are closely linked with tensor decomposition. Here, we explore some of these connections. In particular, we show that the CANDECOMP/PARAFAC (CP, \cite{Harshman1970}) method is a less general version of the multiple matrix tri-factorisation (MMTF) part of our HMF model; and furthermore that the Tucker Decomposition (TD, \cite{Tucker1966}), without its orthogonality constraints, is equivalent to MMTF. All three decompositions are illustrated in Figure \ref{cp_td_mmtf}, and we define them mathematically below. \\
	
	\noindent The CP method decomposes a given tensor $ \R \in \mathbb{R}^{I \times J \times N} $ into the sum of $K$ rank-1 tensors. This is effectively a generalisation of matrix factorisation to three (rather than two) dimensions, with each dimension getting its own factor matrix: $\boldsymbol F^1 \in \mathbb{R}^{I \times K}, \boldsymbol F^2 \in \mathbb{R}^{J \times K}, \boldsymbol S \in \mathbb{R}^{N \times K}$. Overall, we perform the factorisation $ \R = \boldsymbol F^1 \otimes \boldsymbol F^2 \otimes \boldsymbol S$, where $\otimes$ denotes the matrix outer product. Each individual entry in $\R$ is decomposed as follows:
	\begin{equation} \label{CP}
		R_{ijn} = \sum_{k=1}^K F^1_{ik} \cdot F^2_{jk} \cdot S_{nk}.
	\end{equation}
	The Tucker decomposition is defined similarly, but in addition to the three factor matrices, we also get a core tensor $ \G \in \mathbb{R}^{K \times L \times Q} $, and the factor matrices have its own number of latent factors $K,L,Q$; $\boldsymbol F^1 \in \mathbb{R}^{I \times K}, \boldsymbol S \in \mathbb{R}^{J \times L}, \boldsymbol F^2 \in \mathbb{R}^{N \times Q}$. 
	We now factorise $ \R = \boldsymbol G \bigcdot_1 \boldsymbol F^1 \bigcdot_2 \boldsymbol F^2 \bigcdot_3 \boldsymbol S $, where $\bigcdot_i$ denotes the matrix dot product using the $i$th dimension of tensor $\boldsymbol G$. Individual entries in $\R$ are decomposed as:
	\begin{equation} \label{TD}
	R_{ijn} = \sum_{k=1}^K \sum_{l=1}^L \sum_{q=1}^Q F^1_{ik} \cdot F^2_{jl} \cdot S_{nq} \cdot G_{klq}.
	\end{equation}
	Now consider the multiple matrix tri-factorisation part of our HMF model. Say we are given $N$ datasets $\R^n$, all spanning the same two entity types $E_1, E_2$, with $I$ rows, $J$ columns, $K$ row factors (for entity type $E_1$), and $L$ row factors (for entity type $E_2$). Performing MMTF on these datasets can be seen as concatenating the $N$ matrices into one big tensor. Each entry is decomposed as:
	\begin{equation} \label{MMTF}
	R_{ij}^n = \sum_{k=1}^K \sum_{l=1}^L F^1_{ik} \cdot F^2_{jl} \cdot S^n_{kl}.
	\end{equation}
	Firstly, compare this with the TD formulation. If we define a new matrix $H_{kl}^n = \sum_{q=1}^Q S_{nq} \cdot G_{klq} = \S_n \cdot \boldsymbol G_{kl} $, we can rewrite the TD expression as: 
	\begin{equation} \label{TD_new}
		R_{ijn} = \sum_{k=1}^K \sum_{l=1}^L F^1_{ik} \cdot F^2_{jl} \cdot H_{kl}^n.
	\end{equation}
	Note that this is now equivalent to our MMTF expression in Equation \ref{MMTF}, with $F_{ik}^1 \leftrightarrow F^1_{ik}, F_{jl}^2 \leftrightarrow F^2_{jl}, S_{kl}^n \leftrightarrow H_{kl}^n$. 
	Since both the $\S$ and $\boldsymbol G$ matrices have to be inferred by our model, we can in fact merge them into one -- as long as no further constraints are placed on them individually. Often in the TD method the three factor matrices ($\boldsymbol F^1, \boldsymbol F^2, \boldsymbol S$) have orthogonality constraints placed on them. If these constraints are dropped, TD is equivalent to MMTF. \\
	
	\noindent Moving on to CP, consider constraining our MMTF model to have only diagonal entries in the $\S^n$ matrices (so that $S_{kl}^n = 0$ for $k \neq l$). Equation \ref{MMTF} then becomes: 
	\begin{equation} \label{MMTF_constrained}
	R_{ij}^n = \sum_{k=1}^K F^1_{ik} \cdot F^2_{jk} \cdot S^n_{kk}.
	\end{equation}
	This is equivalent to the CP formulation in Equation \ref{CP}, with $F_{ik}^1 \leftrightarrow F^1_{ik}, F_{jk}^2 \leftrightarrow F^2_{jk}, S_{kk}^n \leftrightarrow S_{nk} $, showing that CP is a constrained version of MMTF, where the middle factor matrices $S_{kl}^n$ are constrained to be diagonal. \\
	
	\noindent Finally, we wanted to validate that the more general formulation offered by our HMF model is necessary to obtain good predictive performances, by comparing it with the CP method. We constrained our HMF model to have diagonal $\S^n$ matrices (we call this method HMF CP), but otherwise the exact same Bayesian priors and settings. The results are given in Tables \ref{CP_drug_sensitivity} and \ref{CP_methylation}. We can see that in the in-matrix prediction setting CP still does fairly well, although our unconstrained MMTF model (HMF D-MTF) outperforms it on all four datasets. However, in the out-of-matrix prediction setting CP does not manage to give sensible predictions, barely doing better than the gene average baseline. 
	
	\begin{table*}[h]
		\caption{Mean squared error (MSE) of 10-fold in-matrix cross-validation results on the drug sensitivity datasets. The best performances are highlighted in bold.} \label{CP_drug_sensitivity}
		\centering
		\begin{tabular}{lcccc}
			\toprule
			Method & GDSC $IC_{50}$ & CTRP $EC_{50}$ & CCLE $IC_{50}$ & CCLE $EC_{50}$ \\
			\midrule
			HMF D-MF    & 0.0775 	           & 0.0919              & 0.0592              & \textbf{0.1062} \\
			HMF D-MTF  & \textbf{0.0768}  & \textbf{0.0908} & \textbf{0.0558} & 0.1073 \\
			HMF CP        & 0.0796 & 0.0913 & 0.0560 & 0.1104 \\
			\bottomrule
		\end{tabular}
	\end{table*}
	
	\begin{table}[h]
		\caption{Mean squared error (MSE) of 10-fold out-of-matrix cross-validation results on the methylation datasets. The best performances are highlighted in bold. } \label{CP_methylation}
		\centering
		\begin{tabular}{lccc}
			\toprule
			Method & GM, PM to GE & GE, GM to PM & GE, PM to GM \\
			\midrule
			Gene average 	&  	1.009			& 	1.008			&	1.009			\\
			HMF D-MF		&	\textbf{0.788}			&	\textbf{0.735}	&	\textbf{0.602}	\\
			HMF D-MTF		&	0.850			&	0.798			&	0.640			\\
			HMF S-MF		&	0.820			&	0.794			&	0.672			\\
			HMF CP			&	1.006			&	0.972			&	0.968			\\
			\bottomrule
		\end{tabular}
	\end{table}
	
	
\clearpage
\section{Datasets and preprocessing}
	\subsection{Drug sensitivity datasets}
		We will now describe the preprocessing steps undertaken for the drug sensitivity datasets used in the paper. We used four different datasets:
		\begin{itemize}
			\item Genomics of Drug Sensitivity in Cancer (GDSC v5.0, \citet{Yang2013}) -- giving the natural log of $IC_{50}$ values for 139 drugs across 707 cell lines, with 80\% observed entries.
			\item Cancer Therapeutics Response Portal (CTRP v2, \cite{Seashore-Ludlow2015}) -- giving $EC_{50}$ values for 545 drugs across 887 cell lines, with 80\% observed entries.
			\item Cancer Cell Line Encyclopedia (CCLE, \cite{Barretina2012}) -- giving both $IC_{50}$ and $EC_{50}$ values for 24 drugs across 504 cell lines, with 96\% and 63\% observed entries respectively.
		\end{itemize}
		$IC_{50}$ values indicate the required drug concentration needed to reduce the activity of a given cell line (cancer type in a tissue) by half. We thus measure when an undesired effect has been inhibited by half. With $EC_{50}$ values we measure the maximal (desired) effect a drug can have on a cell line, and then measure the concentration of the drug where we achieve half of this value. In both cases, a lower value is better. \\
		
\begin{figure}[b!]
			\captionsetup{width=0.95\columnwidth}
			\begin{subfigure}{0.5 \columnwidth}
				\includegraphics[width=\columnwidth]{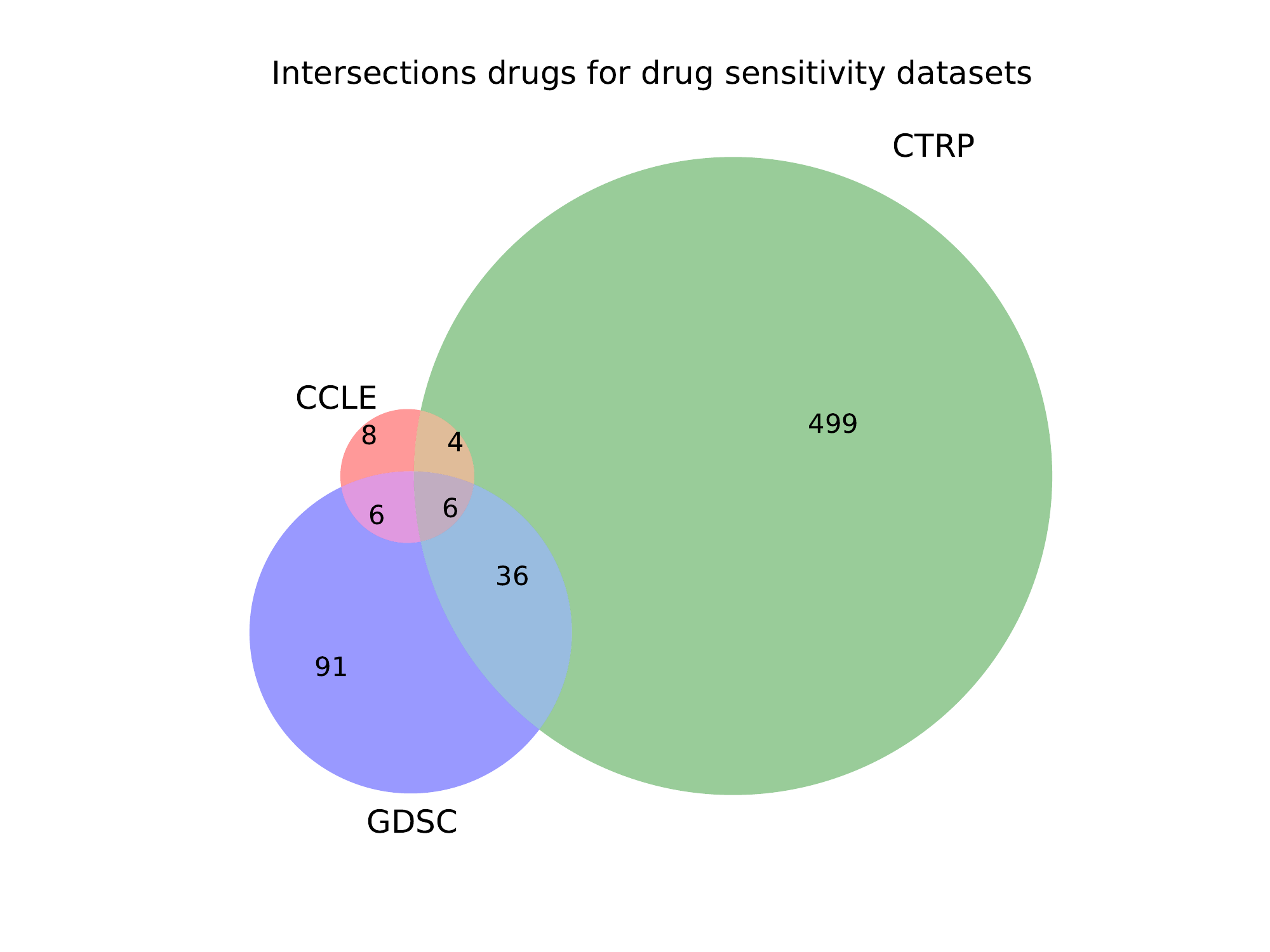}
				\caption{Drugs}
				\label{venn_drugs}
			\end{subfigure} %
			\begin{subfigure}{0.5 \columnwidth}
				\includegraphics[width=\columnwidth]{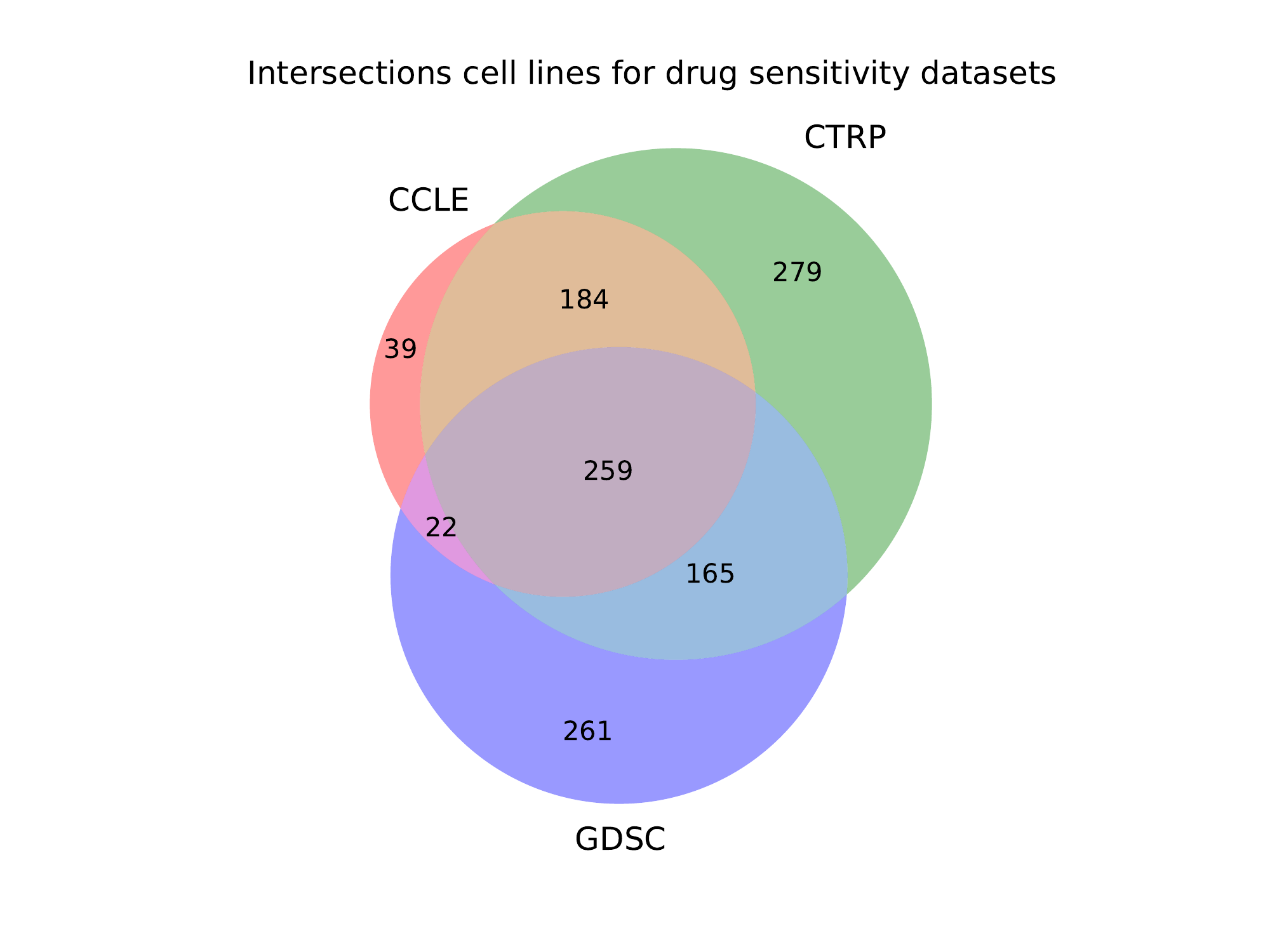}
				\caption{Cell lines}
				\label{venn_cell_lines}
			\end{subfigure}
			\caption{Venn diagrams of the drugs and cell lines in the four datasets.}
		\end{figure}

		\noindent In this paper we are most interested in enhancing predictive power by integrating different datasets. Therefore we focus on drugs and cell lines for which at least two of the four datasets have values available, giving 52 drugs and 630 cell lines. Venn diagrams displaying the overlaps between drugs and cell lines are given in Figures \ref{venn_drugs} and \ref{venn_cell_lines}, respectively. The CTRP dataset contains a large number of small molecule probes (311) causing very little intersection with the other datasets. We also filtered out cell lines with no features available, as discussed in Subsection \ref{features}. \\
		
	\subsection{Preprocessing drug sensitivity values}
\begin{figure}
			\centering
			\captionsetup{width=0.95\columnwidth}
			\begin{subfigure}{0.24 \columnwidth}
				\includegraphics[width=\columnwidth]{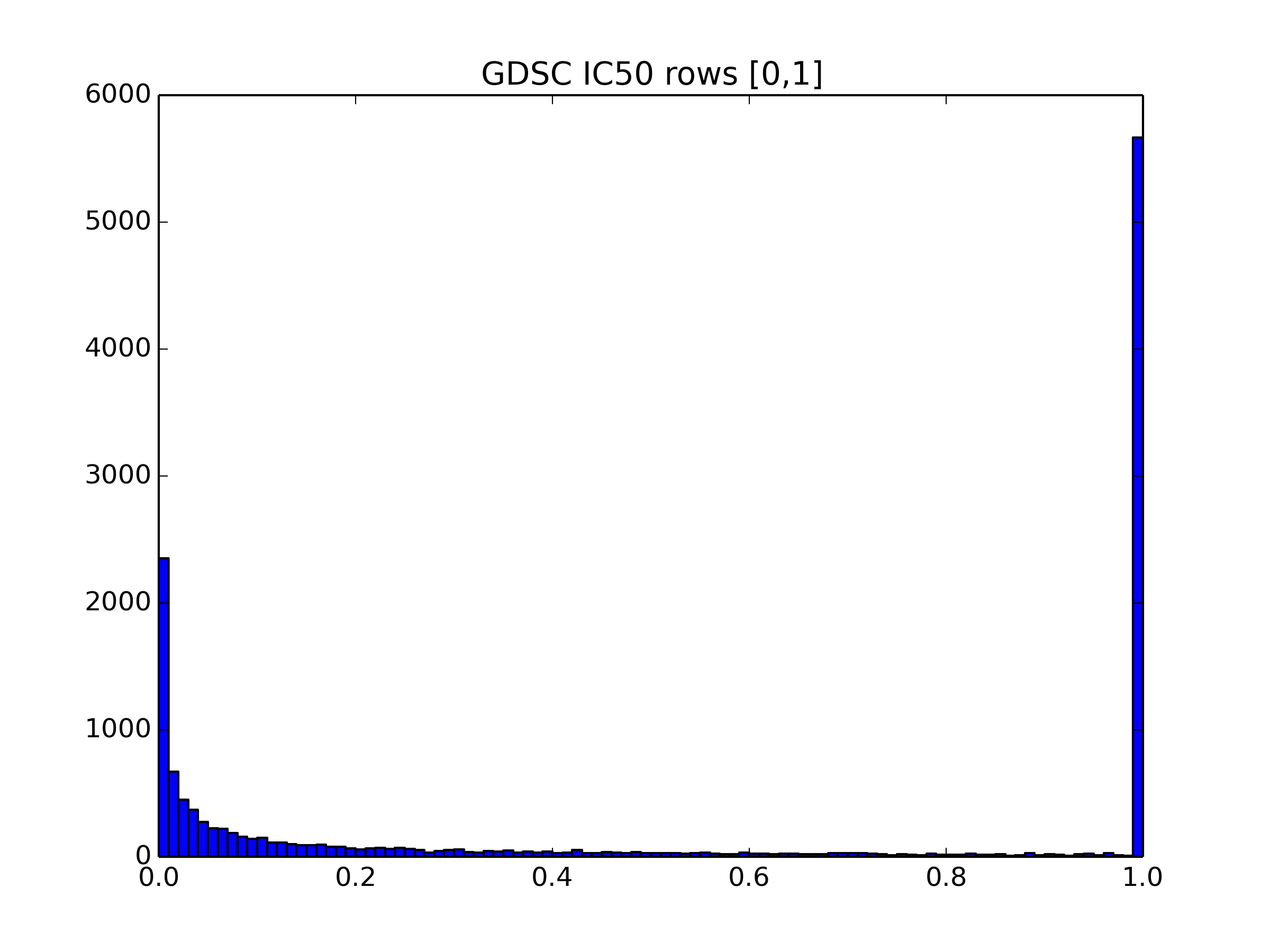}
				\captionsetup{width=0.9\columnwidth}
				\caption{GDSC $IC_{50}$} 
			\end{subfigure} %
			\begin{subfigure}{0.24 \columnwidth}
				\includegraphics[width=\columnwidth]{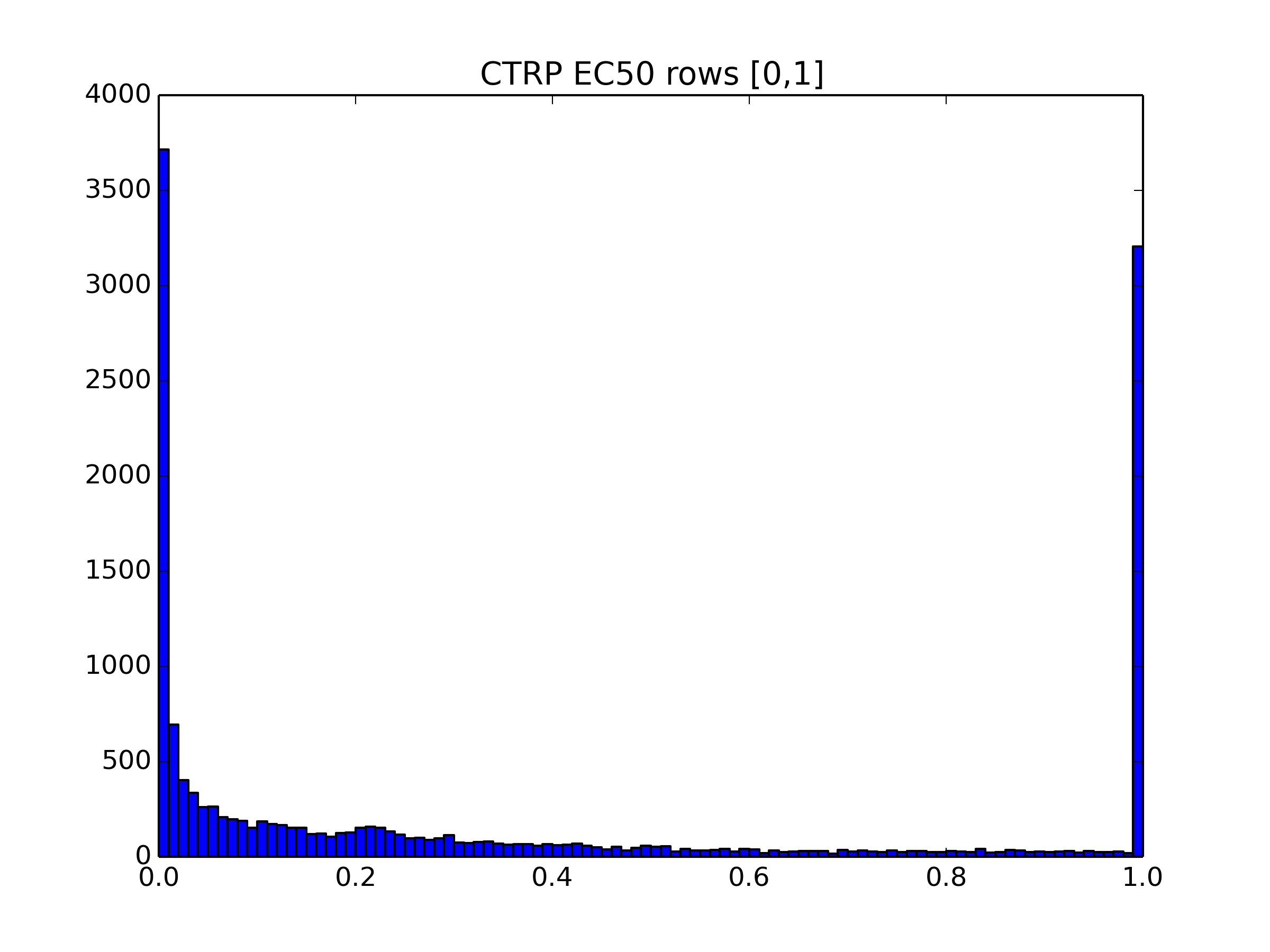}
				\captionsetup{width=0.9\columnwidth}
				\caption{CTRP $EC_{50}$} 
			\end{subfigure} %
			\begin{subfigure}{0.24 \columnwidth}
				\includegraphics[width=\columnwidth]{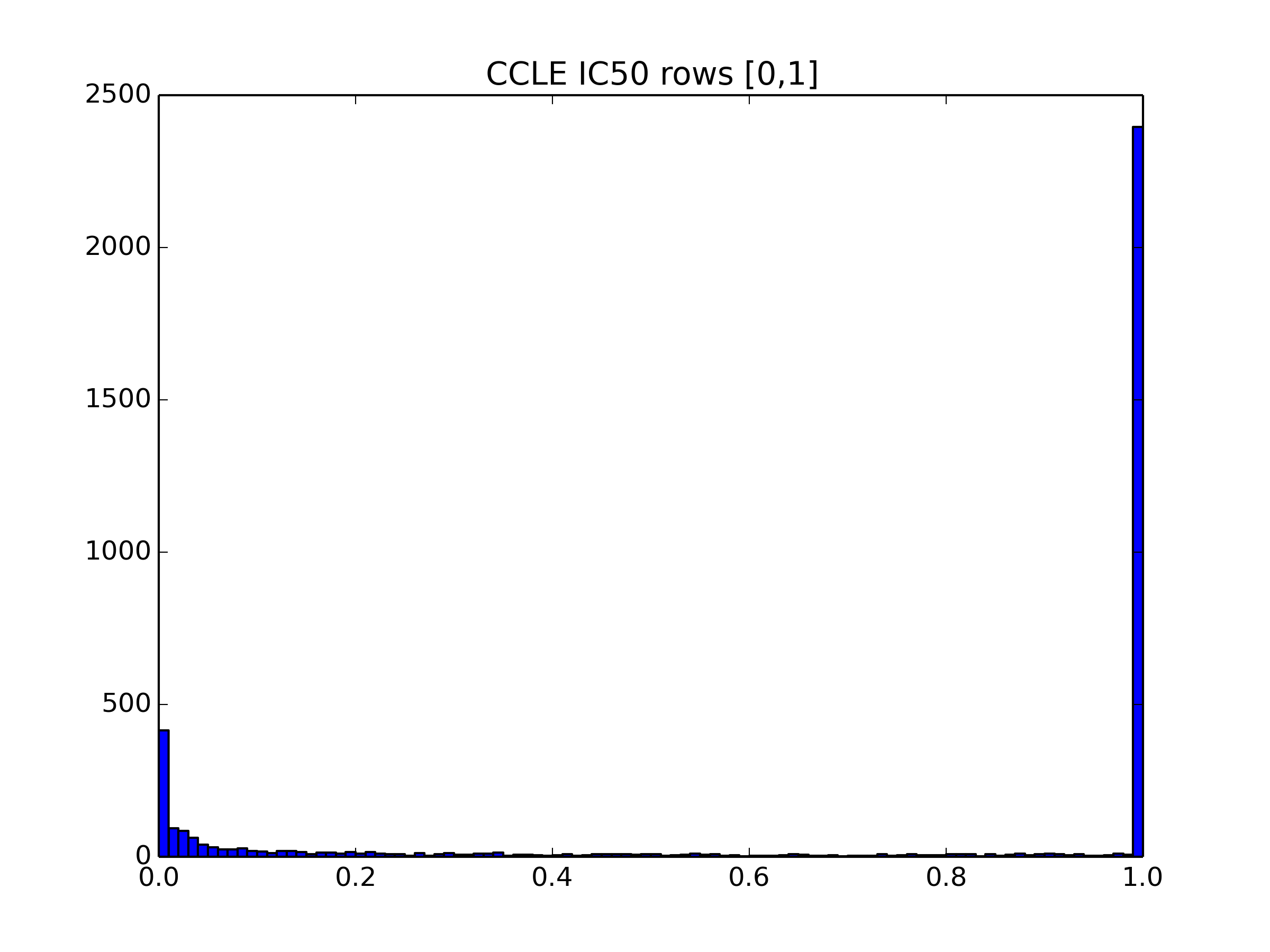}
				\captionsetup{width=0.9\columnwidth}
				\caption{CCLE $IC_{50}$} 
			\end{subfigure} %
			\begin{subfigure}{0.24 \columnwidth}
				\includegraphics[width=\columnwidth]{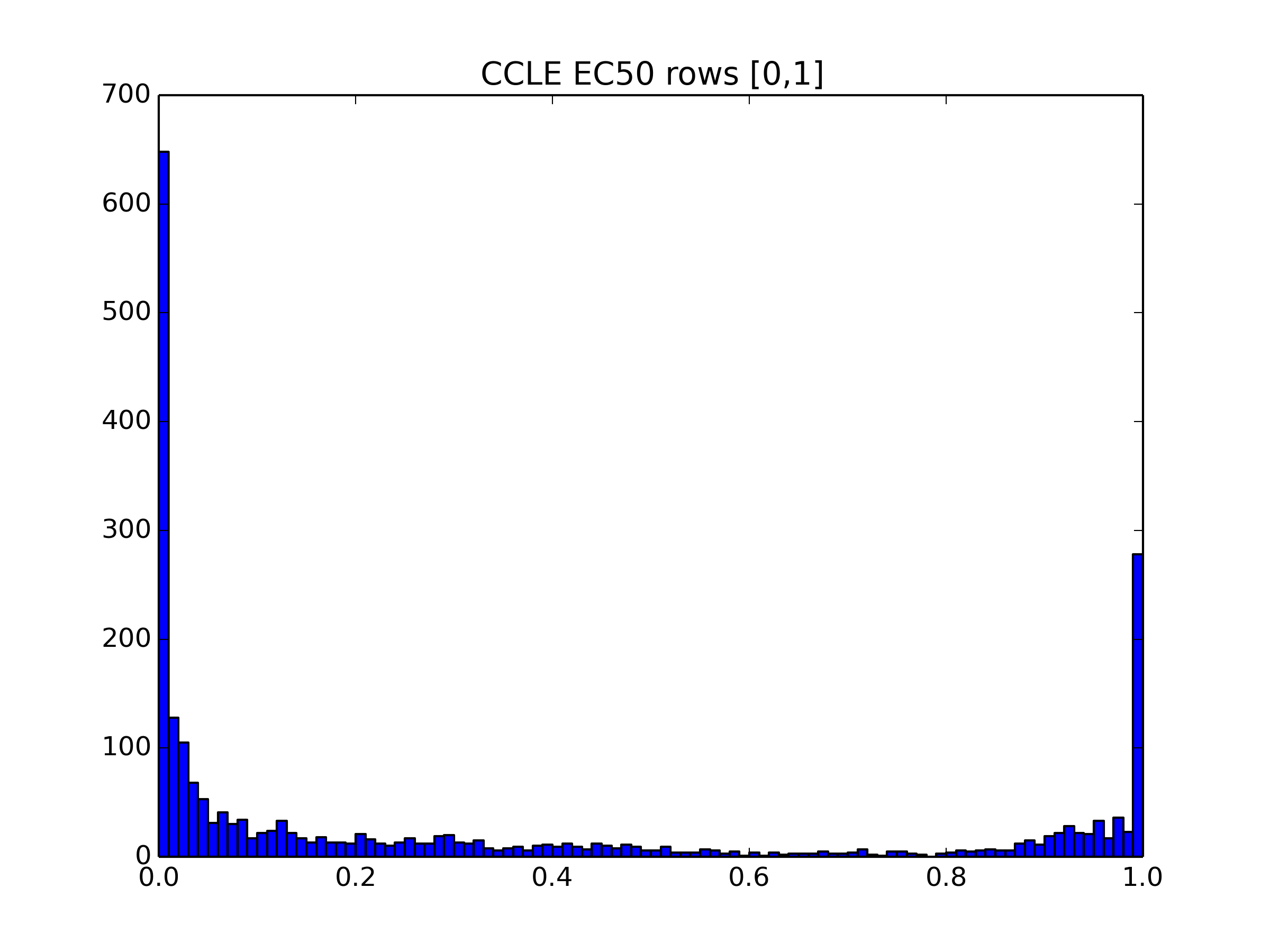}
				\captionsetup{width=0.9\columnwidth}
				\caption{CCLE $EC_{50}$} 
			\end{subfigure}
			\caption{Plots of the distribution of values in the drug sensitivity datasets, after capping the extremely high values in the CTRP $EC_{50}$ and GDSC $IC_{50}$ datasets to 20.}
			\label{distribution_drug_sensitivity}
		\end{figure}

		\noindent The CCLE and CTRP datasets all give the drug concentration levels, but the GDSC dataset gives the natural log transform of these values. We undo this transform by taking the exponent of each value. The drug sensitivity values for the CCLE $IC_{50}$ and $EC_{50}$ datasets lie in the range [0,8] and [0,10], but the other two datasets sporadically have extremely large values. This is a result of the curve fitting procedure used to approximate $IC_{50}$ and $EC_{50}$, and in those cases it indicates an inefficient drug for the cell line. We cap all values above 20 to 20 to resolve this issue, and as a result obtain a similar shape of distribution of values to the CCLE datasets.
		Finally, we map the values in each row (per cell line) to the range [0,1]. 
		This is shown in Figure \ref{distribution_drug_sensitivity}, where we see that the data tends to be bimodal. \\ 
		
	\subsection{Features} \label{features}
		\noindent We also want to incorporate feature information, which is readily provided by the GDSC dataset for 399 of the 630 cell lines. This gave us gene expression information (13321 genes, positive values), copy number variations (CNV; 426 features, count data), and mutations (82, binary data). We filtered out cell lines without this information. \\
		
		\noindent For the drugs we used the PubChem Identifier Exchange Service (\url{https://pubchem.ncbi.nlm.nih.gov/idexchange/idexchange.cgi}) to obtain the PubChem identifiers for all the drugs. Where there were multiple, we used the one in the GDSC database, or otherwise the first one. We then used the PaDeL-Descriptor software (\url{http://www.yapcwsoft.com/dd/padeldescriptor/}) to extract 1D and 2D descriptors, as well as Pubchem fingerprints of functional groups in the drugs. The GDSC dataset has drug target information available for 48 out of the 52 drugs, which we extracted from their website and encoded as a binary dataset. For the four remaining targets we mark the entries as unknown using the mask matrix in the feature dataset and kernel. We removed features with the same value across all drugs. \\ 
		
\begin{figure*}
			\centering
			\captionsetup{width=0.95\columnwidth}
			\begin{subfigure}{0.32 \columnwidth}
				\includegraphics[width=\columnwidth]{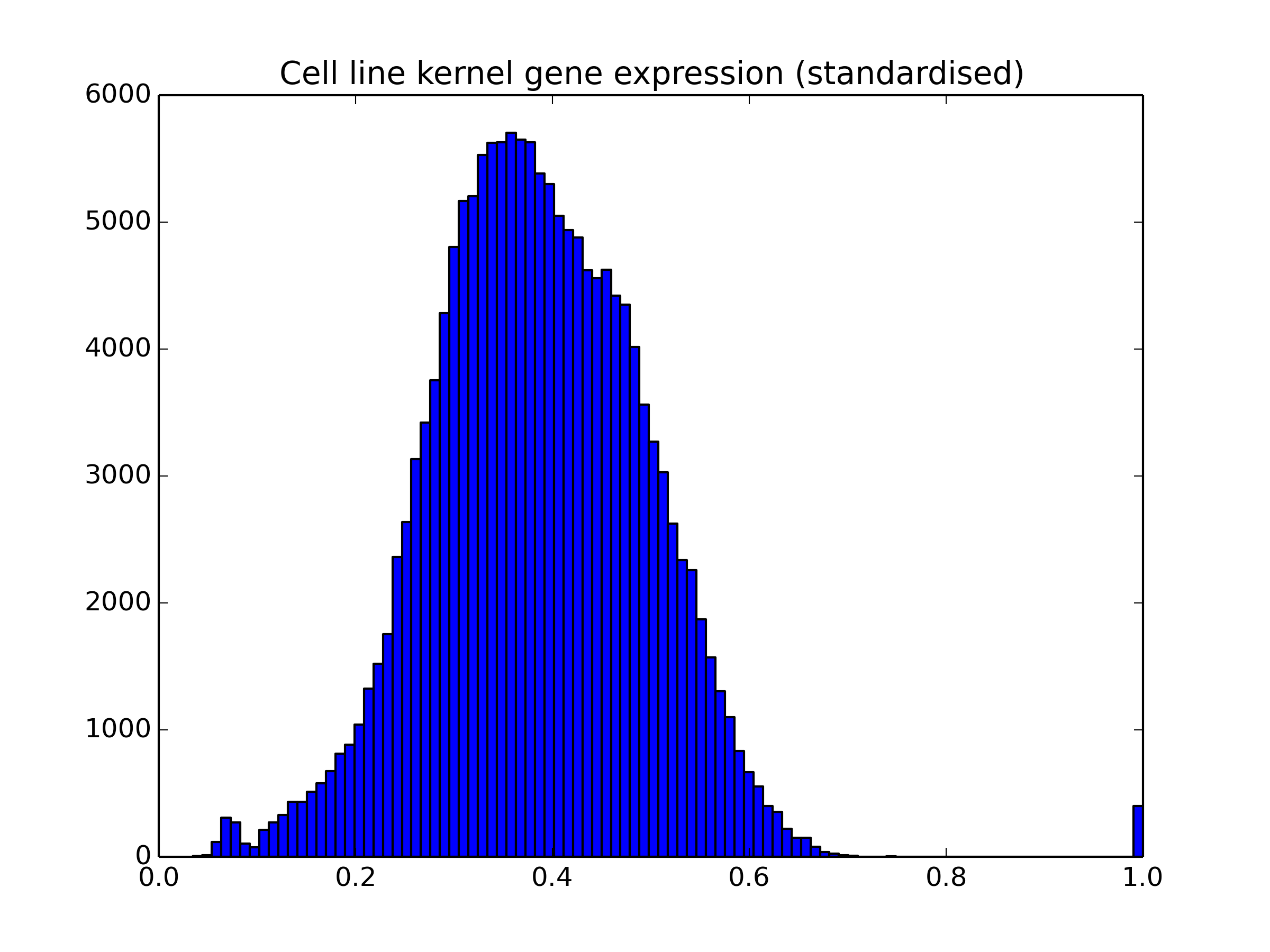}
				\captionsetup{width=0.9\columnwidth}
				\caption{Gene expression} 
			\end{subfigure} %
			\begin{subfigure}{0.32 \columnwidth}
				\includegraphics[width=\columnwidth]{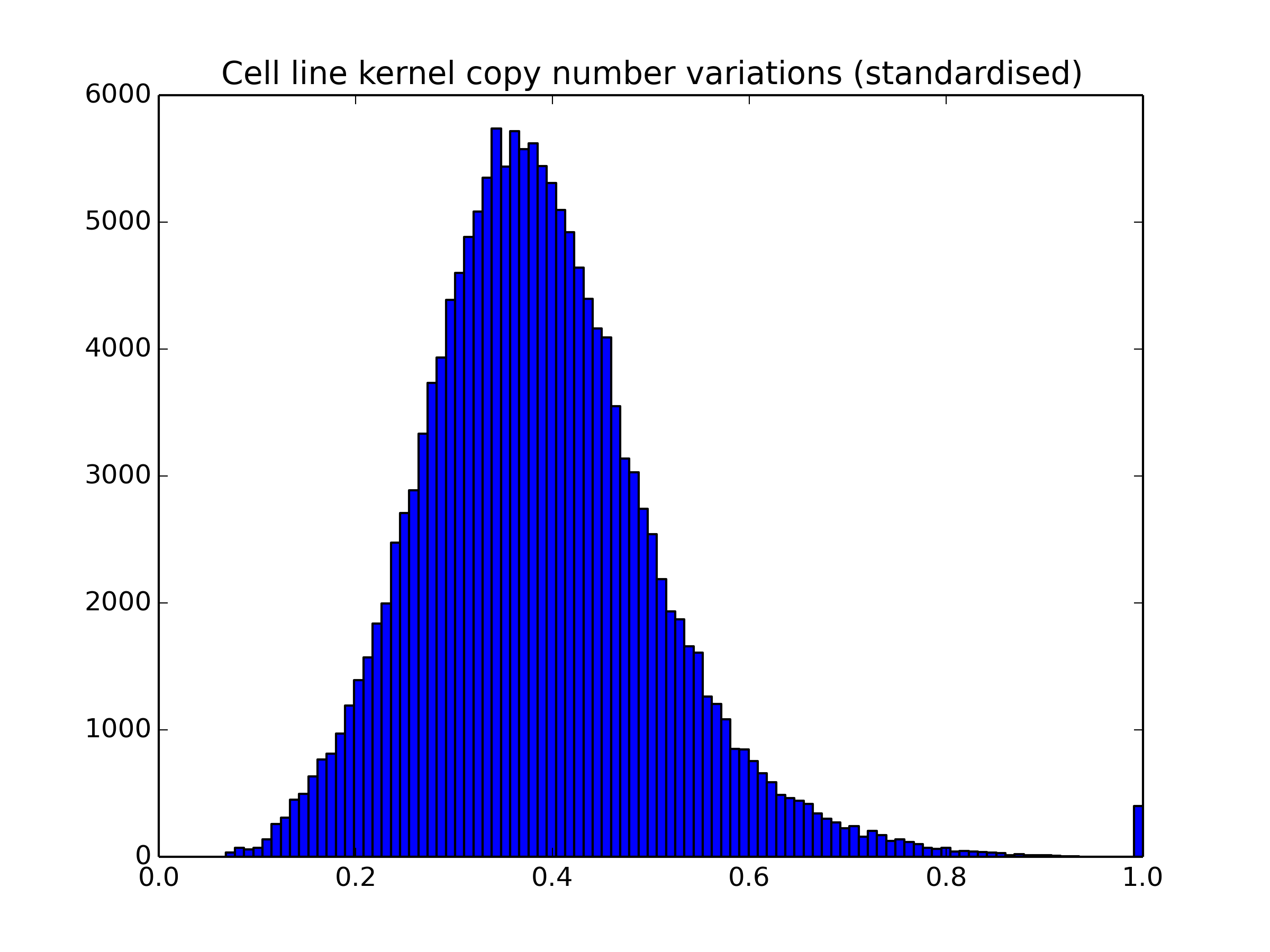}
				\captionsetup{width=0.9\columnwidth}
				\caption{CNV} 
			\end{subfigure} %
			\begin{subfigure}{0.32 \columnwidth}
				\includegraphics[width=\columnwidth]{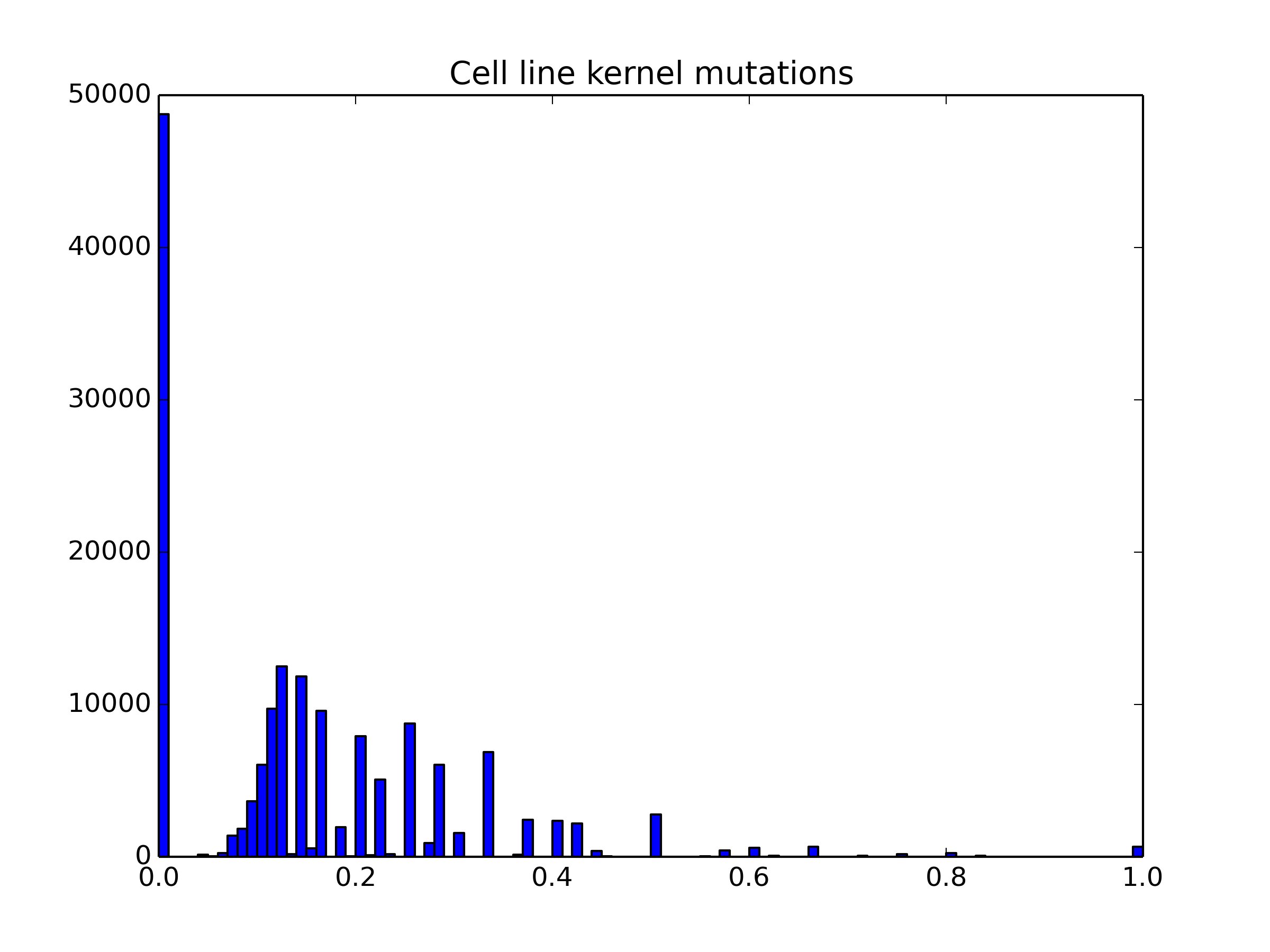}
				\captionsetup{width=0.9\columnwidth}
				\caption{Mutations} 
			\end{subfigure}
			\begin{subfigure}{0.32 \columnwidth}
				\includegraphics[width=\columnwidth]{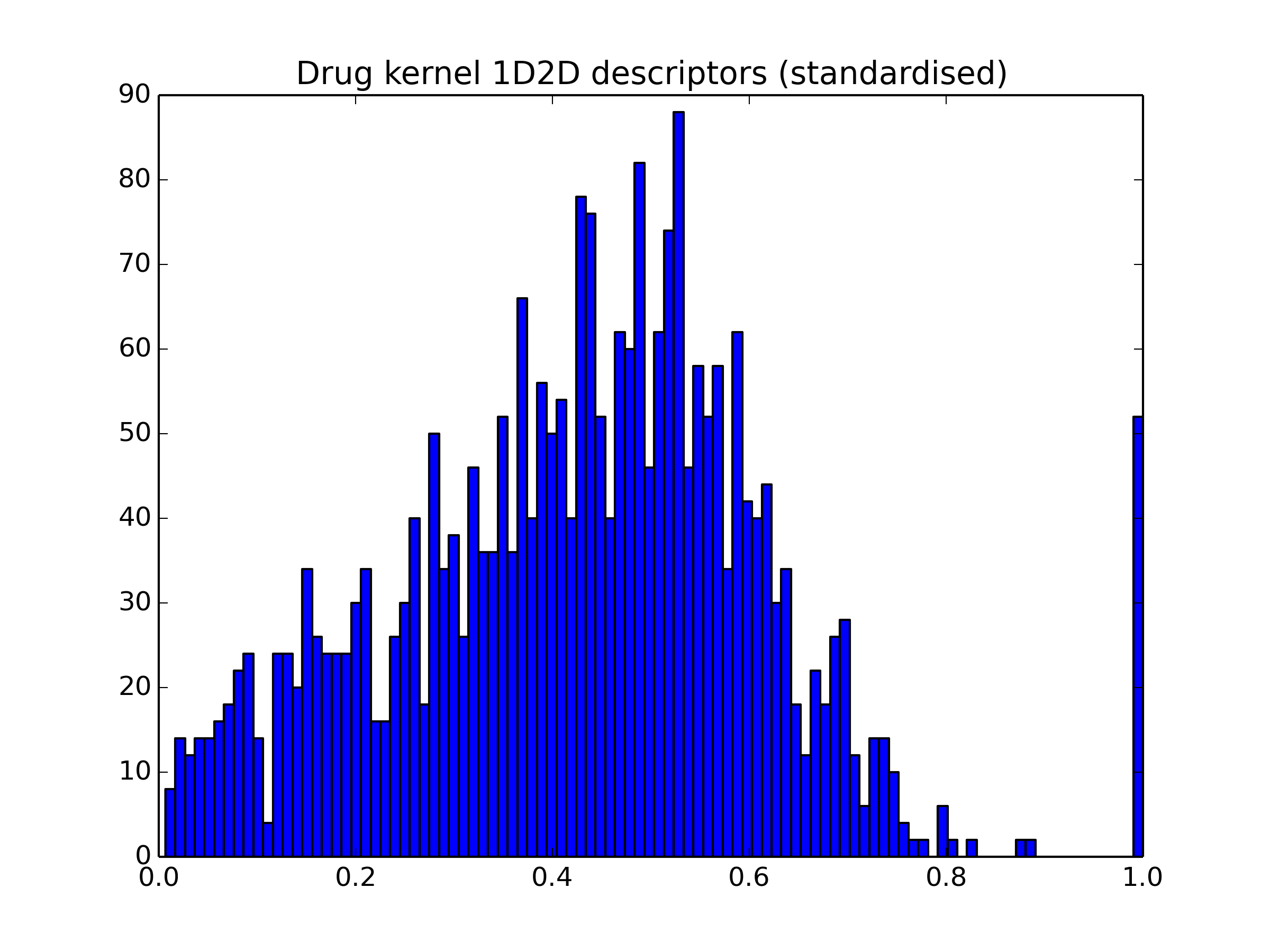}
				\captionsetup{width=0.9\columnwidth}
				\caption{1D 2D descriptors} 
			\end{subfigure} %
			\begin{subfigure}{0.32 \columnwidth}
				\includegraphics[width=\columnwidth]{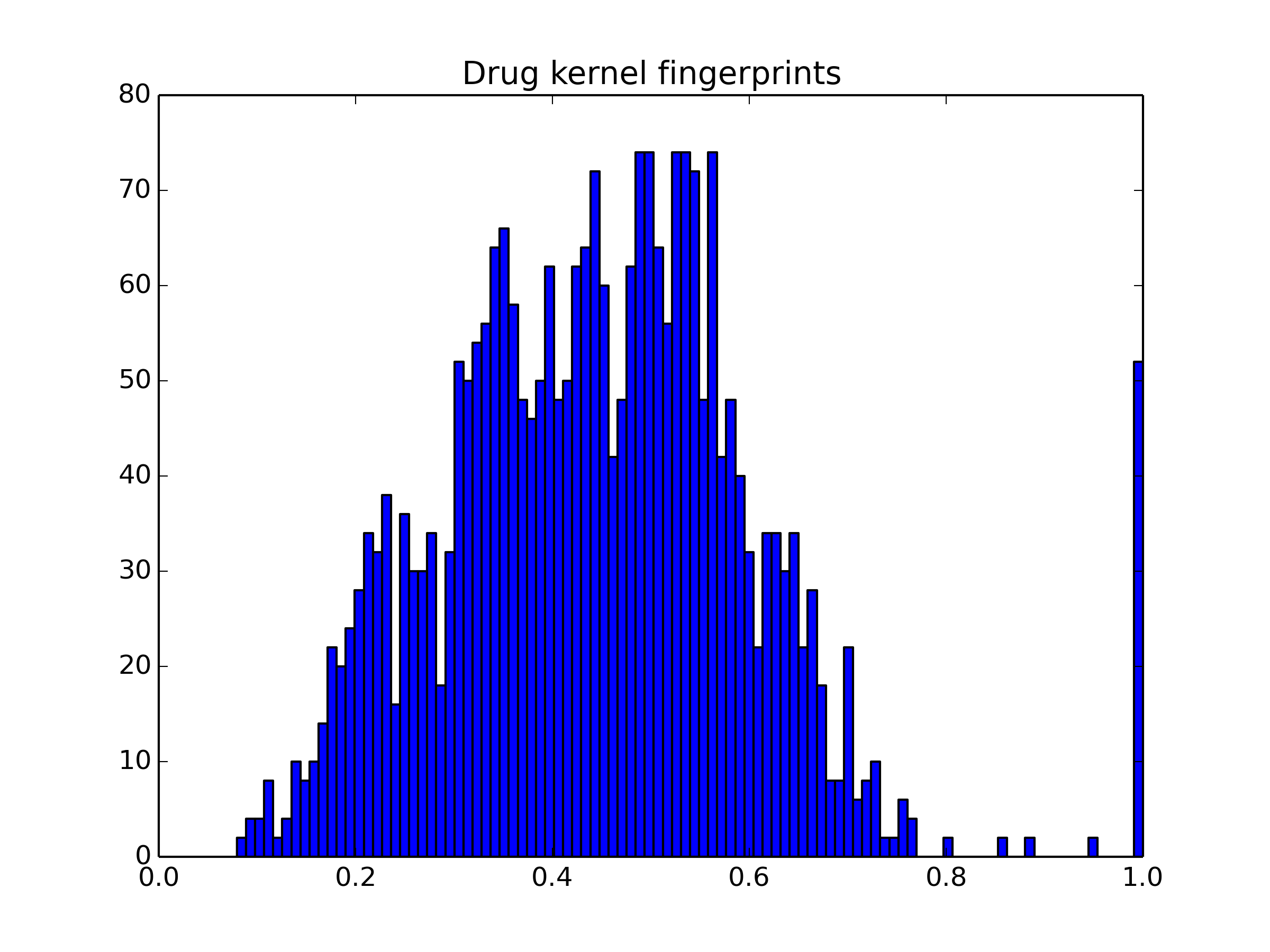}
				\captionsetup{width=0.9\columnwidth}
				\caption{Fingerprints} 
			\end{subfigure} %
			\begin{subfigure}{0.32 \columnwidth}
				\includegraphics[width=\columnwidth]{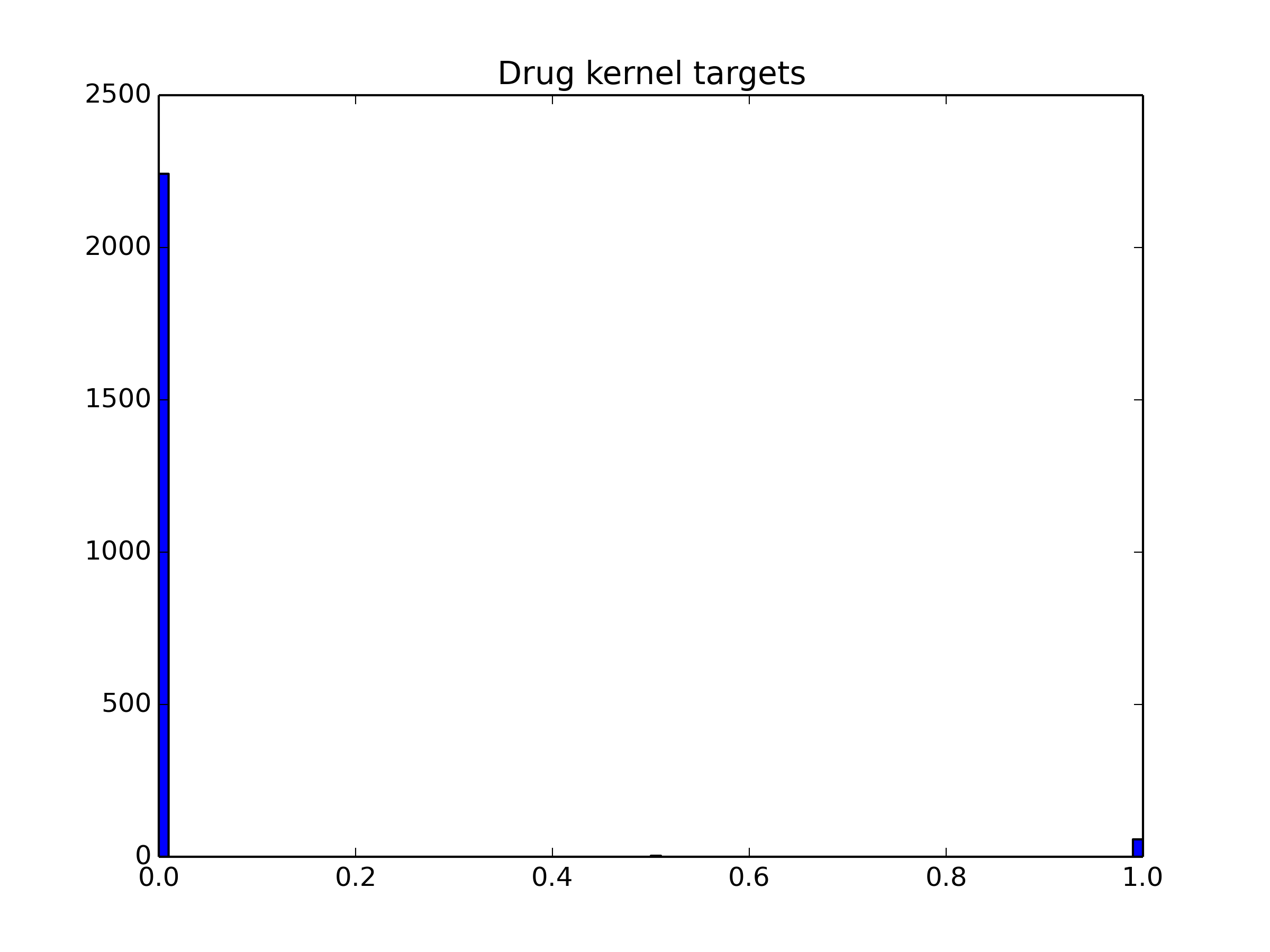}
				\captionsetup{width=0.9\columnwidth}
				\caption{Targets} 
			\end{subfigure}
			\caption{Plots of the distribution of similarity values in the kernel matrices based on the feature datasets. The similarity scores based on drug fingerprints, drug targets, and cell line mutations were constructed using a Jaccard kernel. The scores for gene expression data, copy number variations, and 1D and 2D descriptors were computed using a Gaussian kernel after standardising the values per feature (zero mean, unit variance) and using the number of features as the variance parameter.}
			\label{distribution_kernels}
		\end{figure*}

		\noindent For each of the feature datasets we constructed a similarity kernel. For binary data we used a Jaccard kernel, and for real-valued data we first standardised each feature to have zero mean and unit variance, and then used a Gaussian kernel to compute similarities, with as variance parameter the number of features. The resulting distributions of kernel similarity values are given in Figure \ref{distribution_kernels}. We found that adding the similarity kernels in our HMF model did not improve the predictions. This is because adding dissimilar datasets makes it very hard (if not impossible) to find a good solution, and instead converges to a bad one. \\ 
		
\begin{figure}[t]
			\centering
			\includegraphics[width=0.4\columnwidth]{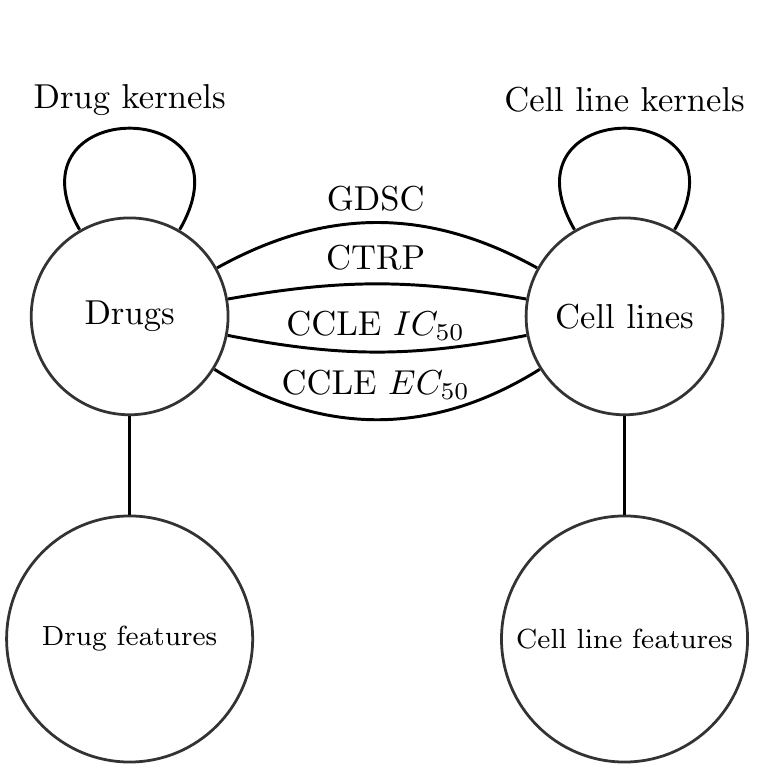}
			\quad\quad\quad\quad\quad\quad
			\includegraphics[width=0.4\columnwidth]{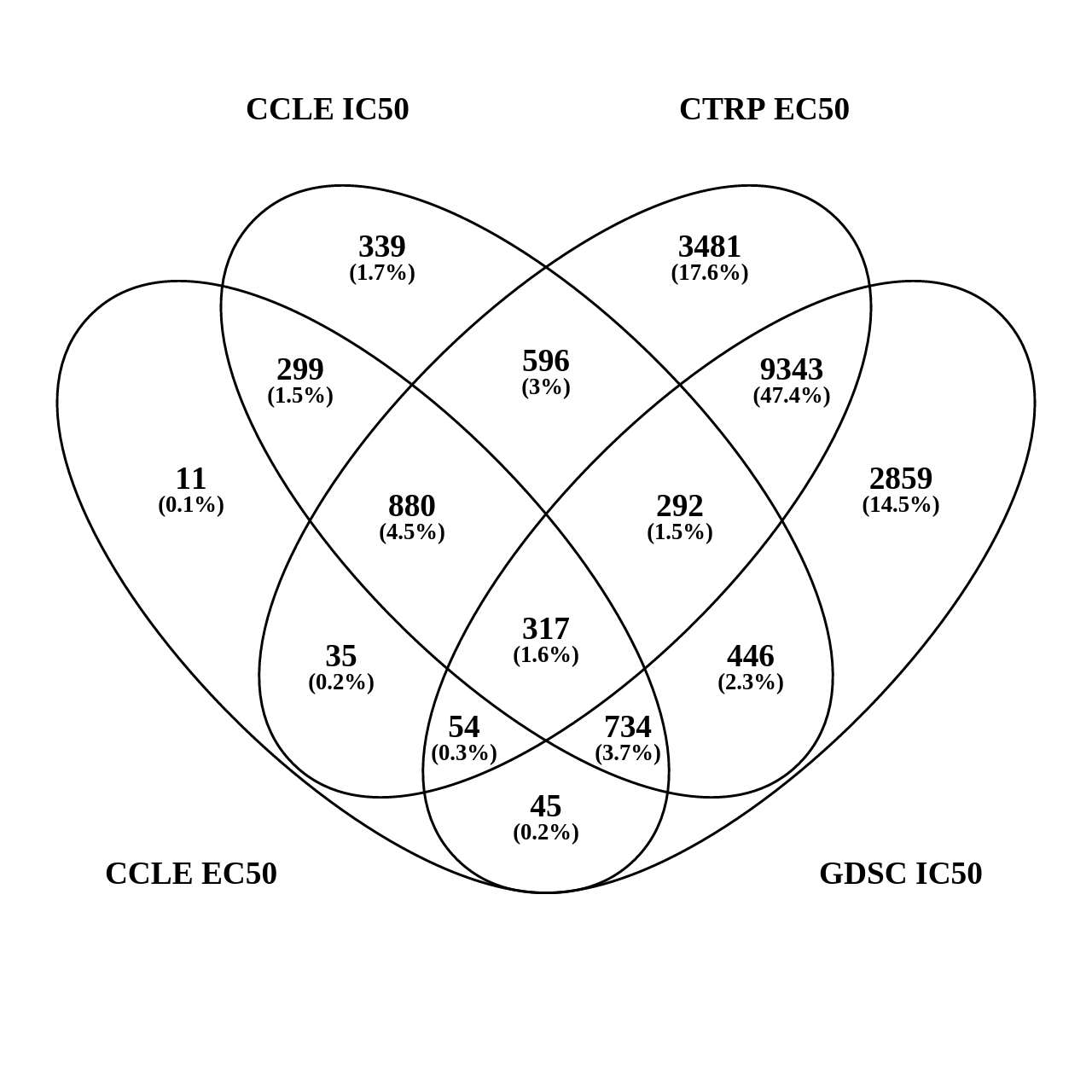}
			\captionsetup{width=1\columnwidth}
			\caption{The datasets used for drug sensitivity prediction on the left; and the overlap of values between the four datasets on the right.}
			\label{diagram_datasets_and_venn_overlap_observed}
		\end{figure}

		\noindent The datasets are summarised in Table \ref{drug_sensitivity_info_datasets}, and represented graphically in Figure \ref{diagram_datasets_and_venn_overlap_observed}. Of the remaining datasets (52 drugs by 399 cell lines), 95.1\% of the entries have a value in at least one of the four datasets, and 62.9\% have an entry in two or more. GDSC $IC_{50}$ has 67.9\% observed entries, CTRP $EC_{50}$ has 72.3\%, CCLE $IC_{50}$ has 18.8\%, and CCLE $EC_{50}$ has 11.4\%. This is also shown in Figure \ref{diagram_datasets_and_venn_overlap_observed}. Individually, the GDSC dataset contains entries for 48 drugs and 399 cell lines (73.6\% observed), CTRP spans 46 drugs and 379 cell lines (86.0\% observed), and finally CCLE $IC_{50}$ and $EC_{50}$ have 16 drugs and 253 cell lines (96.4\% and 58.6\% observed). \\
		
		\begin{table}[t]
			\caption{Drug sensitivity datasets and feature datasets, summarising the entities they relate, the sizes of the datasets, and the fraction of observed entries.} \label{drug_sensitivity_info_datasets}
			\begin{center}
			\begin{tabular}{cccc}
			\toprule
			{\bf Dataset}  &{\bf Entities}	&{\bf Size}	&{\bf Observed}  \\ 
			\midrule
			GDSC $IC_{50}$						&	Cell lines, drugs			&	399, 52		&	67.9\%	 \\
			CTRP $EC_{50}$						&	Cell lines, drugs			&	399, 52		&	72.3\%	 \\
			CCLE $IC_{50}$						&	Cell lines, drugs			&	399, 52		&	18.8\%	 \\
			\vspace{5pt} CCLE $EC_{50}$			&	Cell lines, drugs			&	399, 52		&	11.4\%	 \\
			Targets								&	Drugs, targets				&	52, 53		&	92.3\%	 \\
			1D2D								&	Drugs, 1D 2D descriptors	&	52, 1160	&	100\%	 \\
			\vspace{5pt} FP 					&	Drugs, fingerprints			&	52, 495		&	100\%	 \\
			Targets kernel						&	Drugs, drugs				&	52, 52		&	85.2\%	 \\
			1D2D kernel							&	Drugs, drugs				&	52, 52		&	100\%	 \\
			\vspace{5pt} FP kernel 				&	Drugs, drugs				&	52, 52		&	100\%	 \\
			CNV									&	Cell lines, CNVs			&	399, 426	&	100\%	 \\
			Mutations 							&	Cell lines, mutations		&	52, 82		&	100\%	 \\
			\vspace{5pt} Gene expression kernel	&	Cell lines, cell lines		&	399, 399	&	100\%	 \\
			CNV kernel							&	Cell lines, cell lines		&	399, 399	&	100\%	 \\
			Mutations kernel 					&	Cell lines, cell lines		&	399, 399	&	100\%	 \\
			\bottomrule
			\end{tabular}
			\end{center}
		\end{table}
		
	\clearpage
	\subsection{Methylation and gene expression datasets}
\begin{figure*}[h!]
			\centering
			\captionsetup{width=0.95\columnwidth}
			\begin{subfigure}{0.32 \columnwidth}
				\includegraphics[width=\columnwidth]{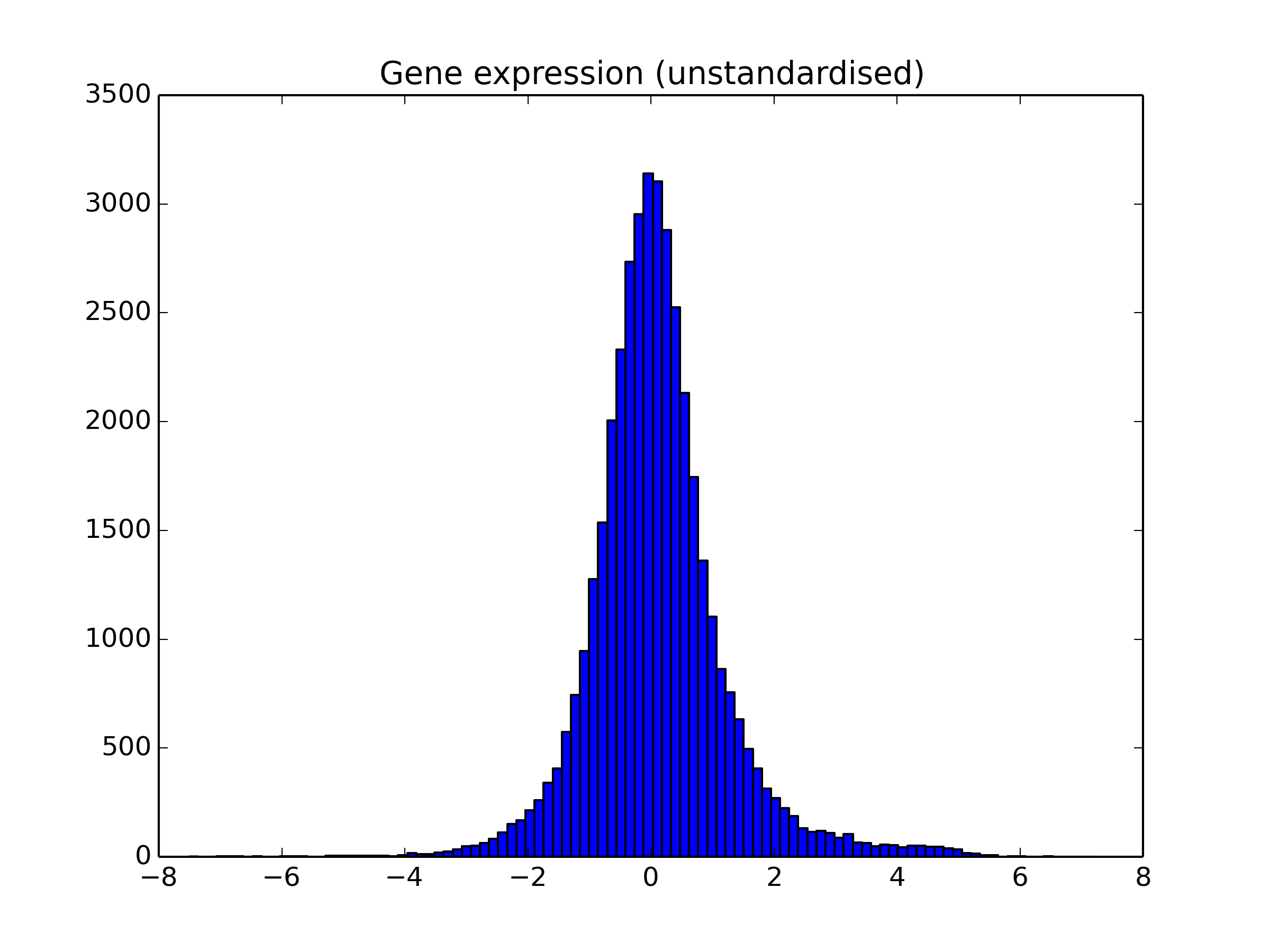}
				\captionsetup{width=0.55\columnwidth}
				\caption{Gene expression, unstandardised} 
			\end{subfigure} %
			\begin{subfigure}{0.32 \columnwidth}
				\includegraphics[width=\columnwidth]{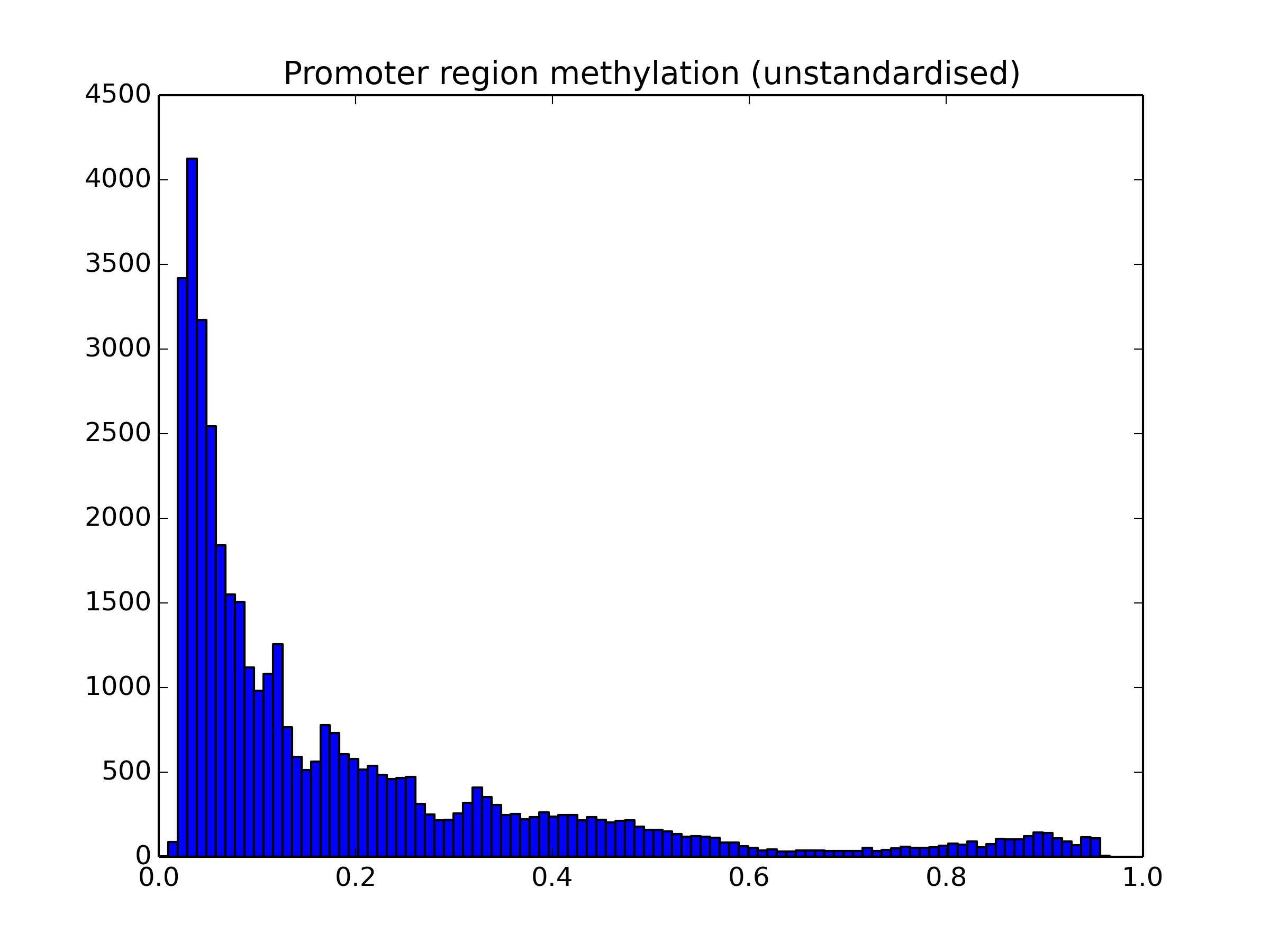}
				\captionsetup{width=0.9\columnwidth}
				\caption{Promoter-region methylation, unstandardised} 
			\end{subfigure} %
			\begin{subfigure}{0.32 \columnwidth}
				\includegraphics[width=\columnwidth]{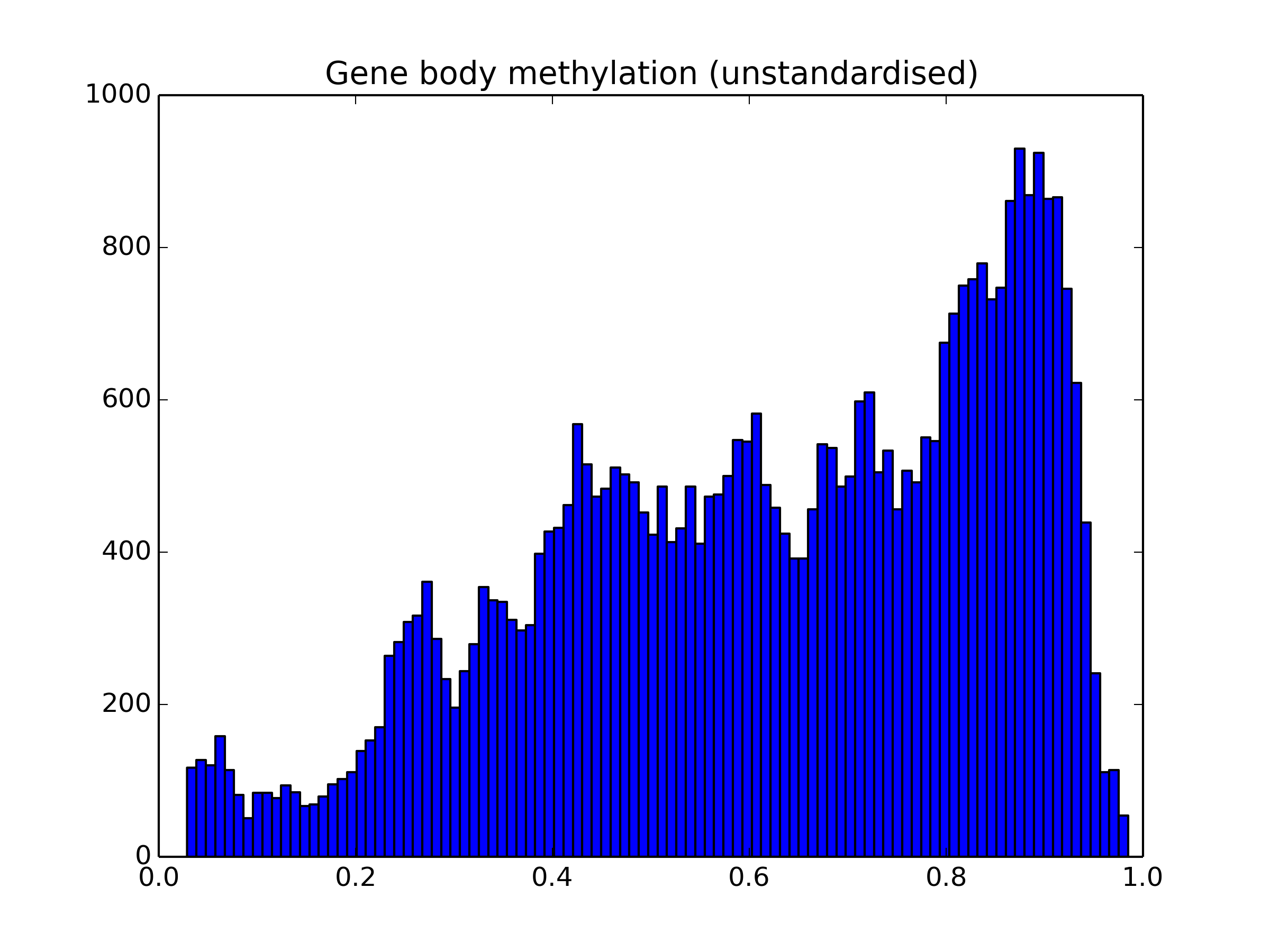}
				\captionsetup{width=0.75\columnwidth}
				\caption{Gene body methylation, unstandardised} 
			\end{subfigure} %
			\begin{subfigure}{0.32 \columnwidth}
				\includegraphics[width=\columnwidth]{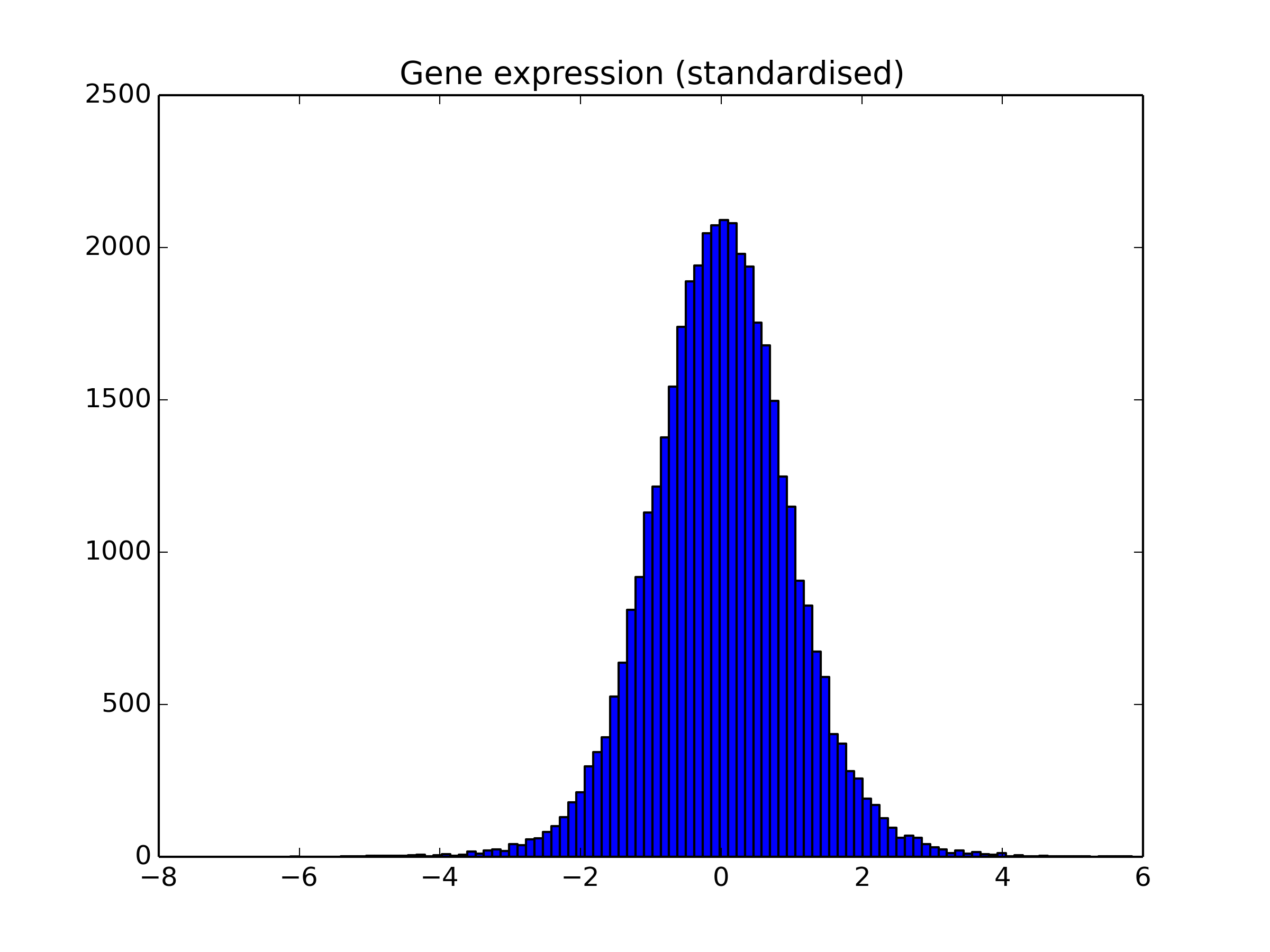}
				\captionsetup{width=0.55\columnwidth}
				\caption{Gene expression, standardised} 
			\end{subfigure} %
			\begin{subfigure}{0.32 \columnwidth}
				\includegraphics[width=\columnwidth]{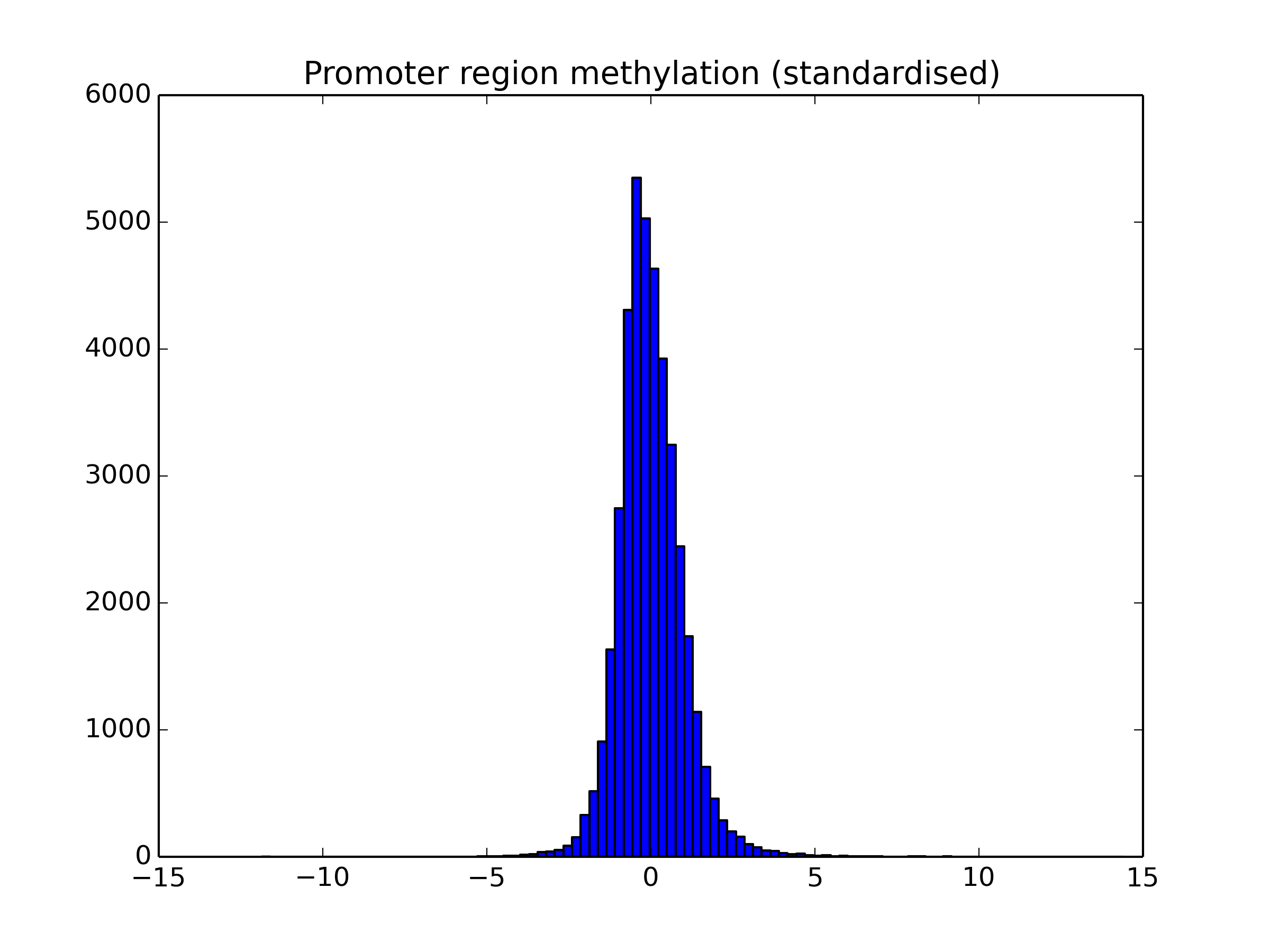}
				\captionsetup{width=0.9\columnwidth}
				\caption{Promoter-region methylation, standardised} 
			\end{subfigure} %
			\begin{subfigure}{0.32 \columnwidth}
				\includegraphics[width=\columnwidth]{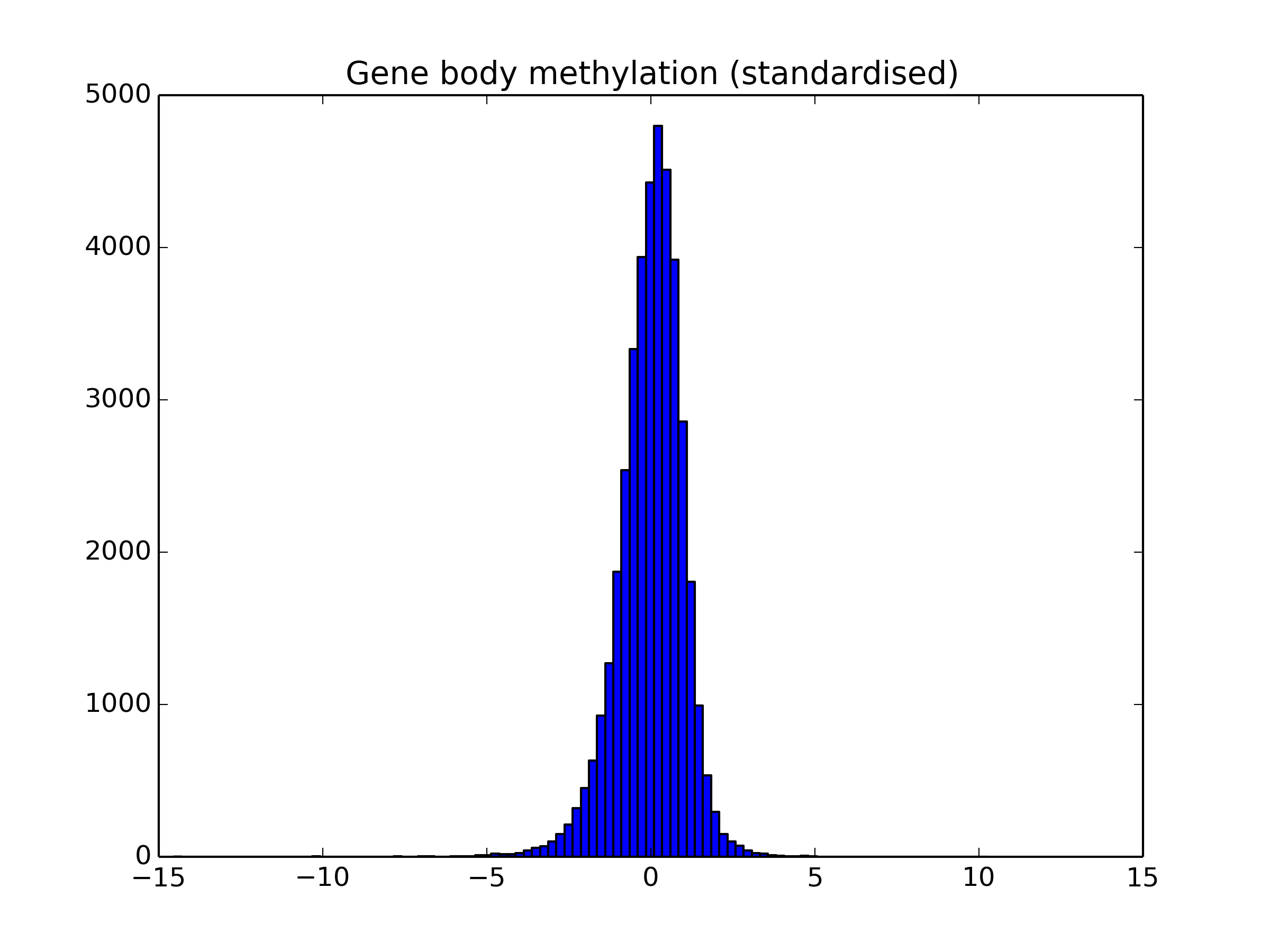}
				\captionsetup{width=0.75\columnwidth}
				\caption{Gene body methylation, standardised} 
			\end{subfigure} %
			\caption{Plots of the distribution of the methylation datasets, before standardising (top row) and after (bottom row).}
			\label{distribution_methylation}
		\end{figure*}

		The preprocessing for the methylation and gene expression datasets is much simpler. We obtained three datasets from the The Cancer Genome Atlas (TCGA, \cite{Koboldt2012}): promoter-region methylation (PM), gene body methylation (GM), and gene expression (GE) profiles for 254 breast cancer patients. This dataset originally spanned 13966 genes, but we focused on 160 breast cancer driver genes, given by the IntOGen database (\cite{Gonzalez-Perez2013}). We standardised each of the datasets to have zero mean and unit standard deviation per gene. Plots of the datasets containing the 160 genes, before and after standardising, can be found in Figure \ref{distribution_methylation}. \\
		
\begin{figure*}[h!]
			\centering
			\captionsetup{width=0.95\columnwidth}
			\begin{subfigure}{0.32 \columnwidth}
				\includegraphics[width=\columnwidth]{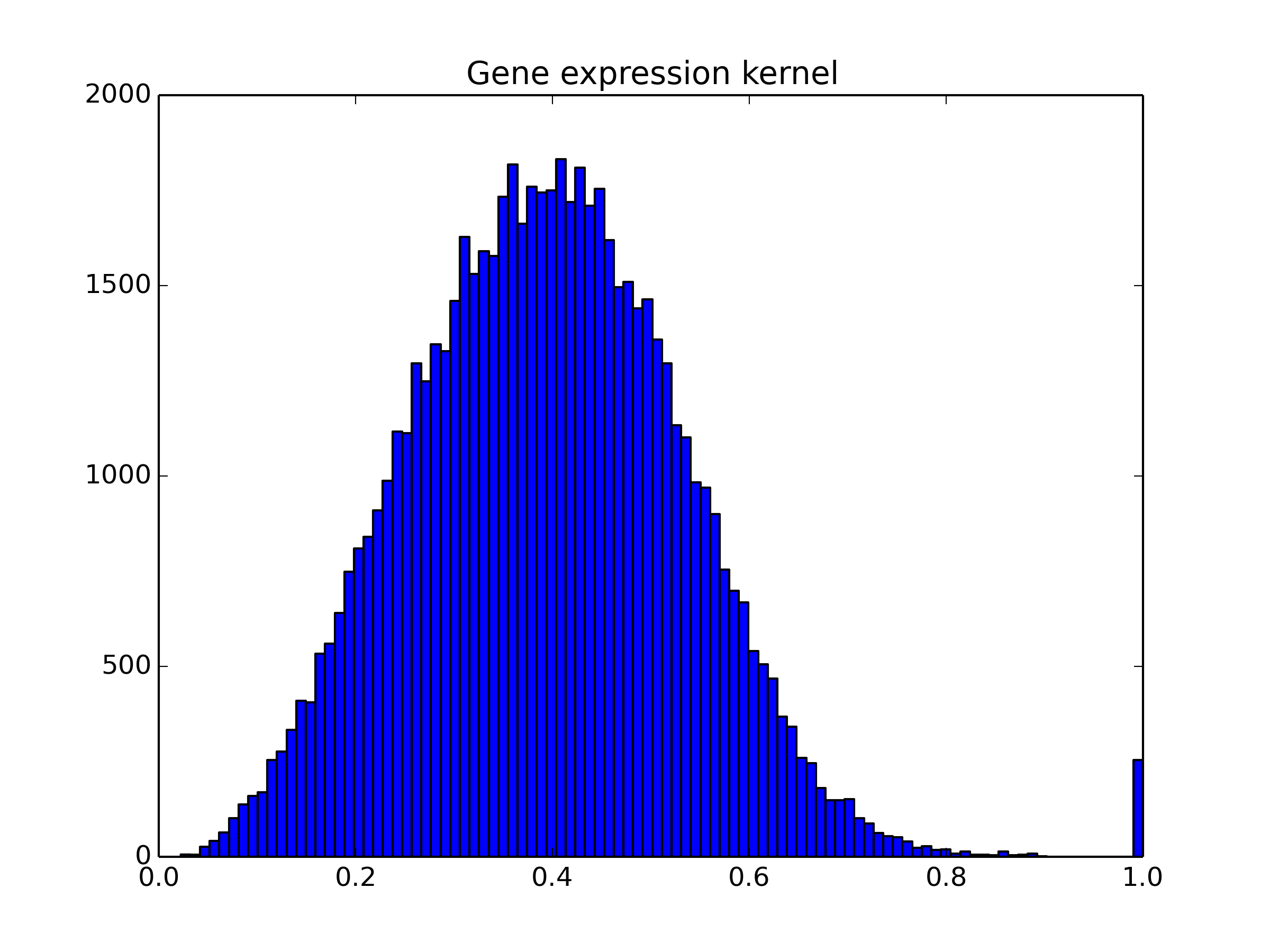}
				\captionsetup{width=1\columnwidth}
				\caption{Gene expression kernel} 
			\end{subfigure} %
			\begin{subfigure}{0.32 \columnwidth}
				\includegraphics[width=\columnwidth]{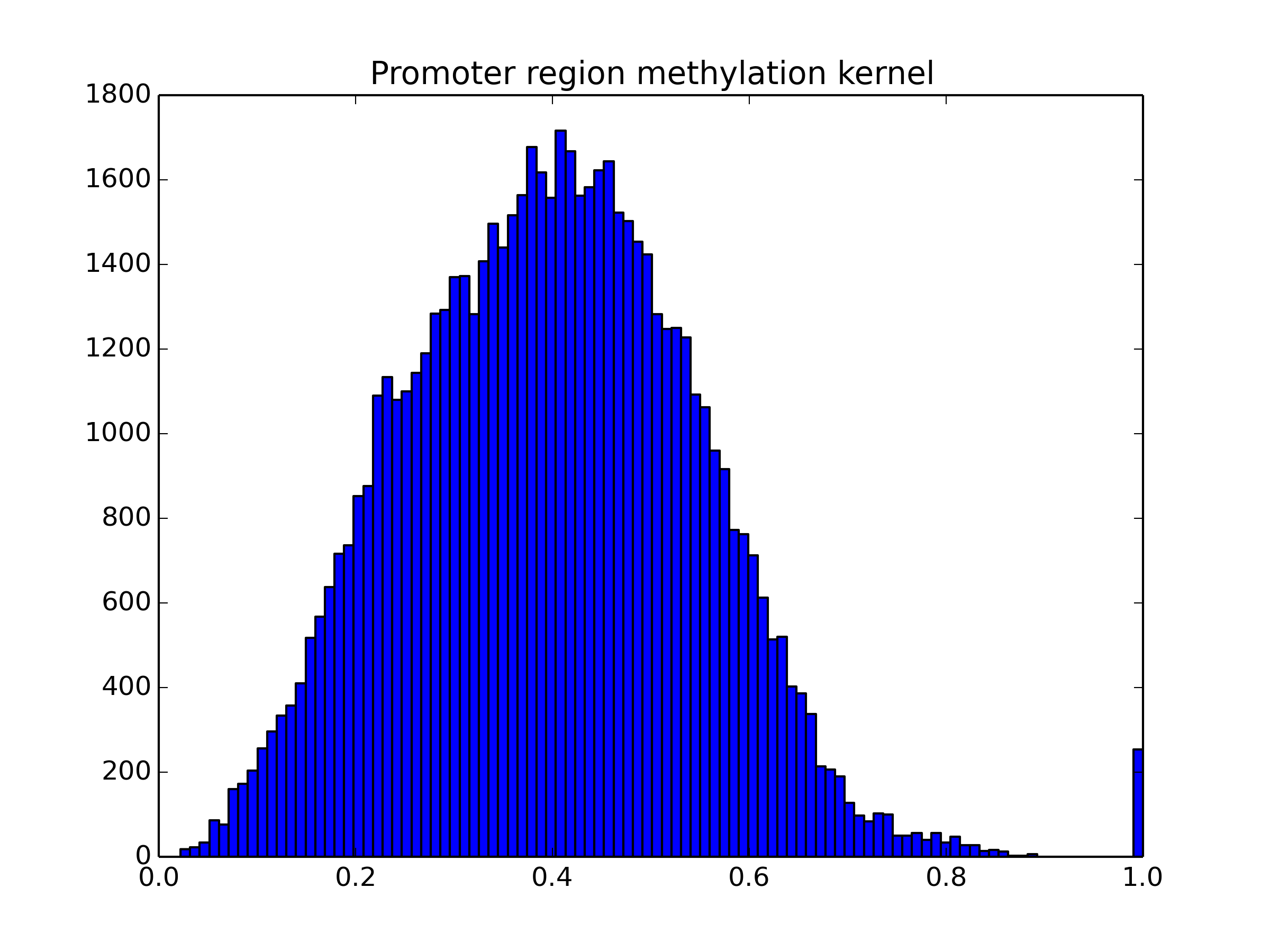}
				\captionsetup{width=1.1\columnwidth}
				\caption{Promoter-region methylation kernel} 
			\end{subfigure} %
			\begin{subfigure}{0.32 \columnwidth}
				\includegraphics[width=\columnwidth]{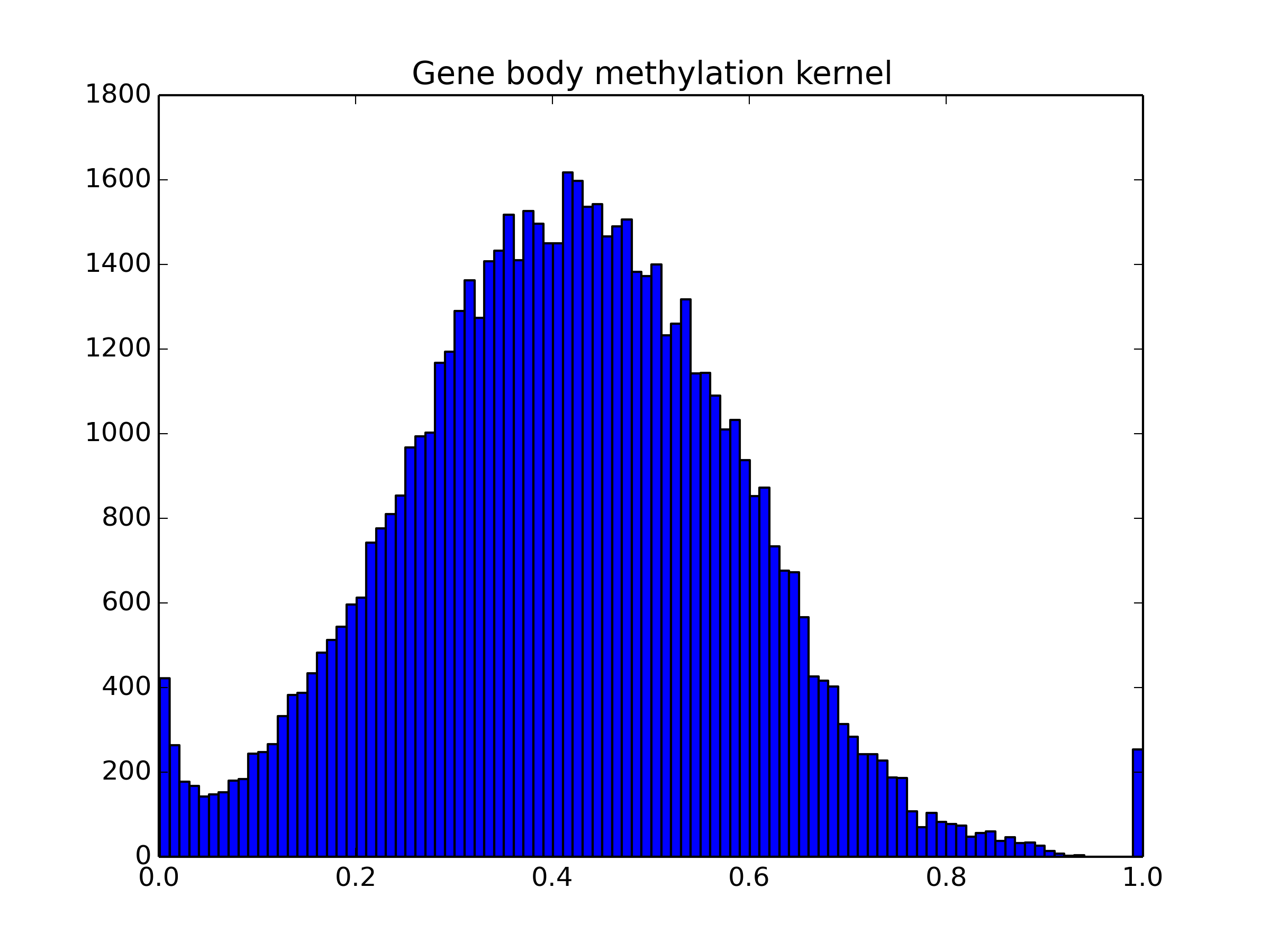}
				\captionsetup{width=1\columnwidth}
				\caption{Gene body methylation kernel} 
			\end{subfigure} %
			\caption{Plots of the distribution of the methylation kernels.}
			\label{distribution_methylation_kernels}
		\end{figure*}

		\noindent To construct the similarity kernels for the HMF S-MF model, giving the similarity of samples, we used a Gaussian kernel with $\sigma^2 = \text{no. genes}$. The kernel value distributions are plotted in Figure \ref{distribution_methylation_kernels}. Notice the Gaussian-like distribution of values between [0,1]. 
		If we generate values randomly from our HMF model's probabilistic definition (for example using value 1 for all hyperparameters, and then randomly sampling values from the prior distributions), we obtain a similar distribution of values. $\sigma^2$ was chosen in such a way to match the model definition more closely. 
		
\clearpage
\section{Additional experiments}
	We performed several additional experiments on the drug sensitivity and methylation datasets, to explore the advantages and limits of our models. In particular, we looked at: the run-time performance of the models; the effectiveness of automatic model selection using the Automatic Relevance Determination (ARD) Bayesian prior; the best initialisation approach; the best values to use for the dataset importances $\alpha$; and the trade-offs of our hybrid choices: effects of factorisation types and nonnegativity constraints on predictive performance.
	
	\subsection{Run-time comparison}
		In this section we give a rough indication of the run-time performances of the models we compared in our experiments. We compare the time it takes for most of the matrix factorisation methods (HMF D-MF, HMF D-MTF, BNMF, BNMTF, NMF, NMTF, and Multiple NMF) on the four drug sensitivity datasets. In particular, we give the number of iterations we used in our experiments for each model, the total run-time it took to train each model, and the time per iteration of the model updates. We use $K_t = 10, K = 10, L = 10$ for all models. \\
		
		\noindent The run-time performances on the drug sensitivity datasets are given in Table \ref{runtime_drug_sensitivity}. Often the matrix factorisation variants are faster than the matrix tri-factorisation versions (up to six times, in the case of NMF and NMTF). The non-probabilistic methods give the fastest run-time (both per iteration, and total), but our experiments show that their predictive performance is worse and they are more prone to overfitting. Our HMF models incur a slightly higher run-time than the other Bayesian models (BNMF, BNMTF), due to having to consider all four datasets at the same time, but only by a factor of roughly two (despite there being four datasets in total). \\
		
		\begin{table*}[b!]
			\captionsetup{width=0.975\columnwidth}
			\caption{Run-time performances of the matrix factorisation models on each of the four drug sensitivity datasets, giving the total number of iterations (It) used to train each model in our experiments (the same on each dataset), and then for each dataset the average number of seconds per iteration (s / it), and the total time in seconds to train a single model (Total).}
			\vspace{5pt}
			\label{runtime_drug_sensitivity}
			\centering
			\begin{tabular}{lccccccccc}
				\toprule
				& & \multicolumn{2}{c}{GDSC $IC_{50}$} & \multicolumn{2}{c}{CTRP $EC_{50}$} & \multicolumn{2}{c}{CCLE $IC_{50}$} & \multicolumn{2}{c}{CCLE $EC_{50}$}	\\
				\cmidrule(lr){3-4} \cmidrule(lr){5-6} \cmidrule(lr){7-8} \cmidrule(lr){9-10}
				Model & Iterations & s / it & Total & s / it & Total & s / it & Total & s / it & Total \\
				\midrule
				HMF D-MF     & 200  & 0.120 & 23.9 & 0.119 & 23.8 & 0.058 & 11.6 & 0.052 & 10.5 \\
				HMF D-MTF   & 200  & 0.303 & 60.6 & 0.293 & 58.6 & 0.148 & 29.5 & 0.148 & 29.6 \\
				BNMF            & 1000 & 0.061 & 61.2 & 0.059 & 58.8 & 0.034 & 34.3 & 0.037 & 37.0 \\
				BNMTF          & 500  & 0.112 & 56.1 & 0.107 & 53.5 & 0.052 & 26.1 & 0.056 & 28.1 \\
				NMF              & 1000 & 0.007 & 6.9 & 0.007 & 6.6 & 0.002 & 2.3 & 0.002 & 2.0 \\
				NMTF            & 1000 & 0.037 & 36.6 & 0.033 & 33.4 & 0.009 & 9.3 & 0.010 & 9.8 \\
				Multiple NMF & 1000 & 0.017 & 16.9 & 0.017 & 17.0 & 0.012 & 11.6 & 0.011 & 11.0 \\
				\bottomrule
			\end{tabular}
		\end{table*}
		
		\noindent Note that when we use cross-validation to measure the predictive performances, our HMF models do not need to run nested cross-validation to choose the dimensionality ($K_t$), whereas the other matrix factorisation models (BNMF, BNMTF, NMF, NMTF, Multiple NMF) do. This actually makes our models faster in cross-validation. \\
		
		\noindent All run-time experiments were conducted on a MacBook Pro laptop, with 2.2 GHz Intel Core i7 processor, 16 GB 1600 MHz DDR3 memory, and an Intel Iris Pro 1536 MB Graphics card.

	\subsection{Model selection}
		Our model employs the ARD prior to perform automatic model selection. In this experiment, we verify how effective this is. We repeat the in-matrix cross-validation experiments on the four drug sensitivity experiments, for the HMF D-MF (multiple matrix factorisation) and HMF D-MTF (multiple matrix tri-factorisation) models. We vary the values for $K_t$, using the values [1, 2, 3, 4, 5, 6, 7, 8, 9, 10, 12, 14, 16, 18, 20, 25, 30]. \\
		
		\noindent Usually when we add more factors to a matrix factorisation models, it gives the model more freedom to fit well to the data, eventually leading to overfitting. Hence, the average cross-validation performance should initially go down, and then go up again as it starts overfitting. If the ARD works as desired, and we add more factors ($K_t$ increases), the ARD will turn them off and use a similar number of factors to before. This should result in less overfitting than the equivalent model with no ARD, resulting in a flatter curve going back up as the number of factors increase. \\
		
\begin{figure}[t]
			\centering
			\begin{subfigure}[t]{0.48 \columnwidth}
				\includegraphics[width=1\columnwidth]{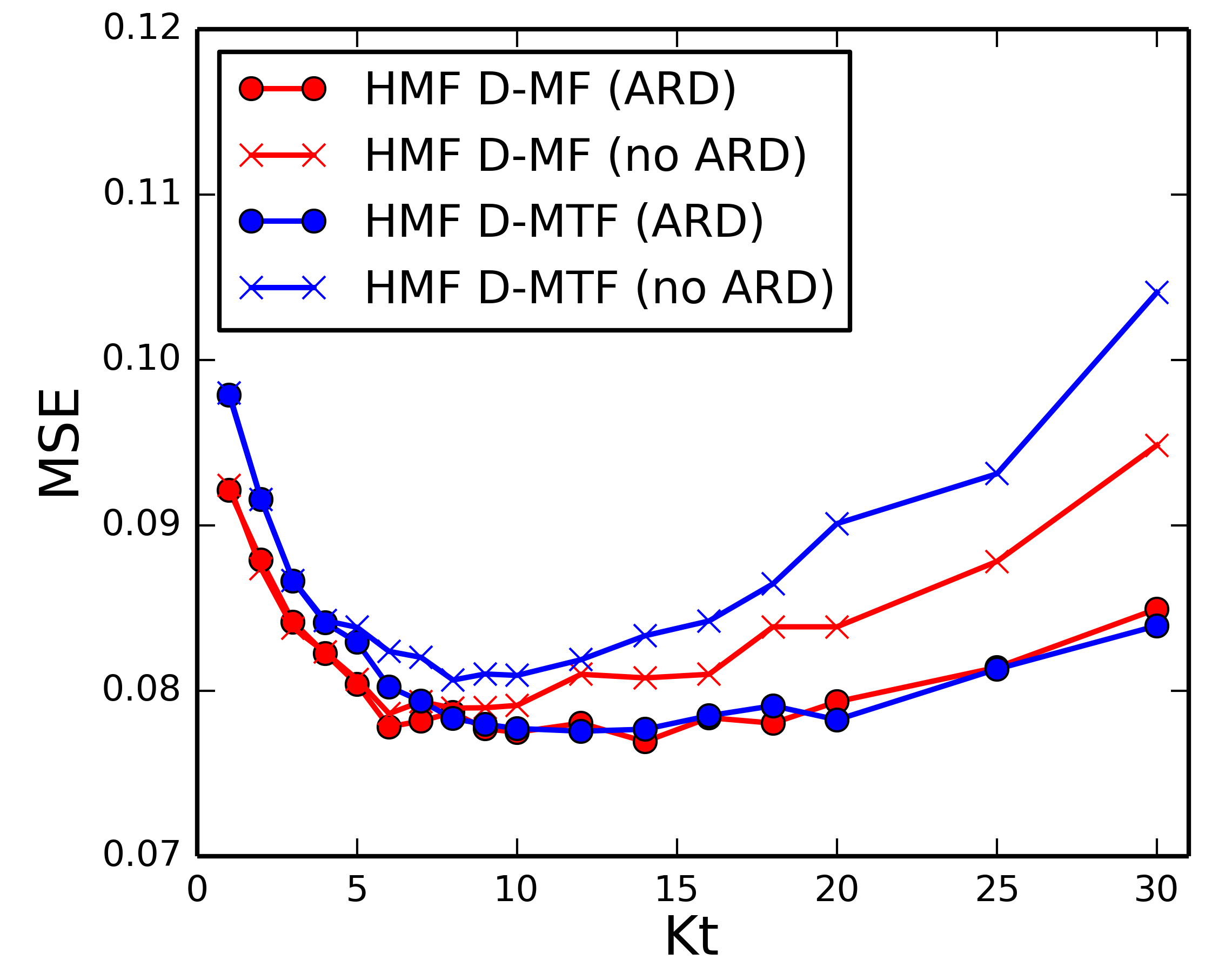}
				\captionsetup{width=0.95\columnwidth}
				\caption{GDSC $IC_{50}$} 
				\label{model_selection_gdsc}
			\end{subfigure}
			\begin{subfigure}[t]{0.48 \columnwidth}
				\includegraphics[width=1\columnwidth]{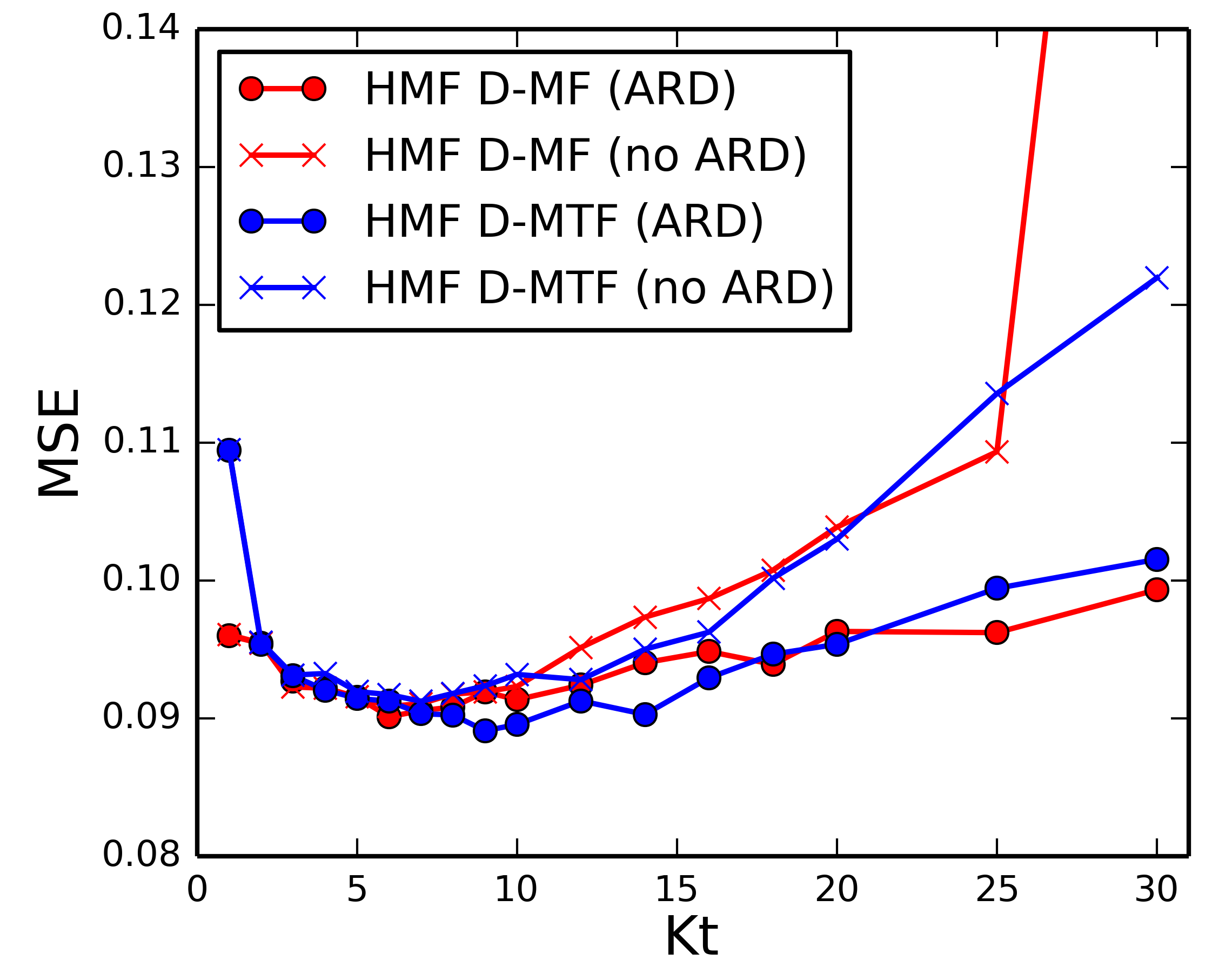}
				\captionsetup{width=0.95\columnwidth}
				\caption{CTRP $EC_{50}$} 
				\label{model_selection_ctrp}
			\end{subfigure}
			\begin{subfigure}[t]{0.48 \columnwidth}
				\includegraphics[width=1\columnwidth]{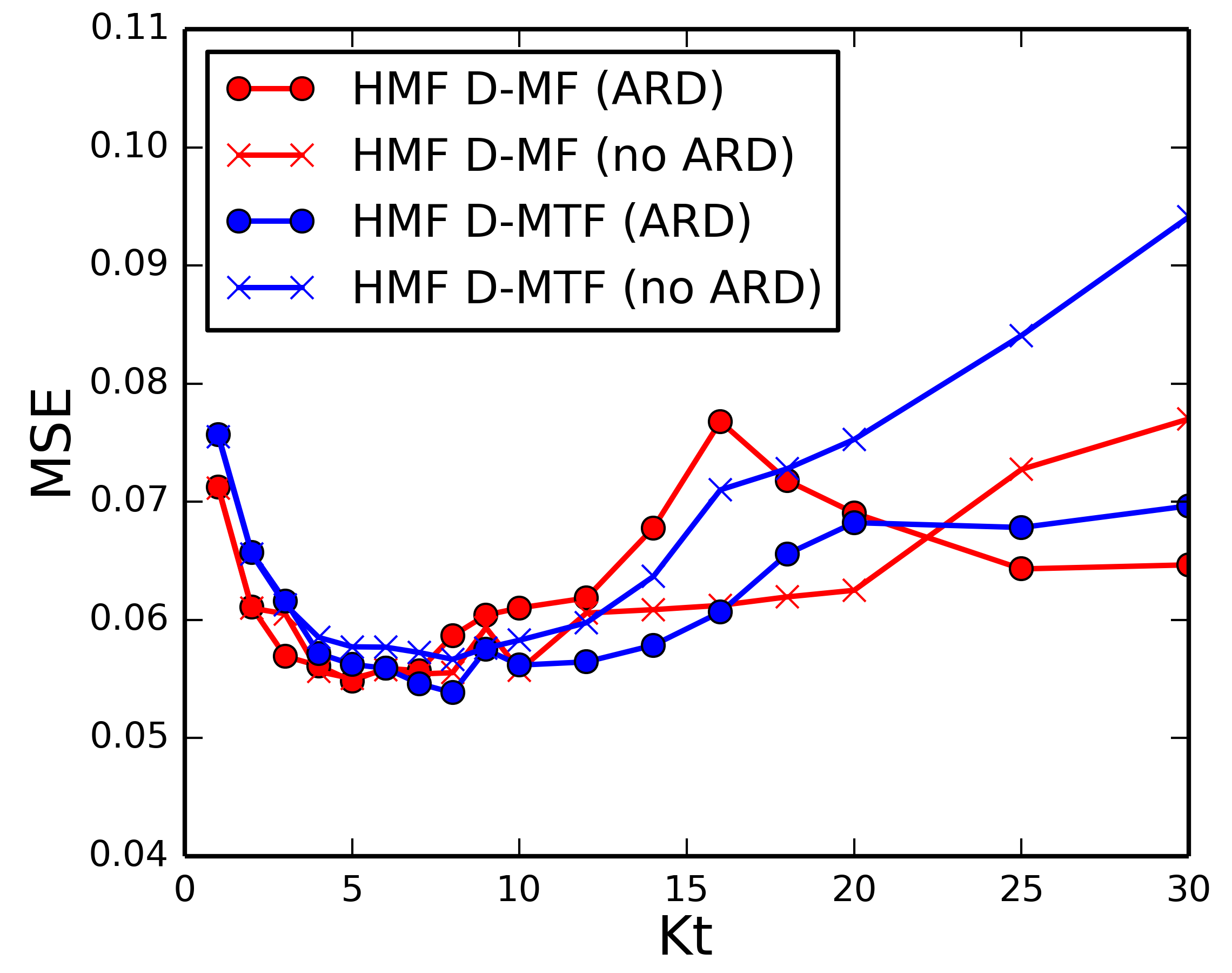}
				\captionsetup{width=0.95\columnwidth}
				\caption{CCLE $IC_{50}$} 
				\label{model_selection_ccle_ic}
			\end{subfigure}
			\begin{subfigure}[t]{0.48 \columnwidth}
				\includegraphics[width=1\columnwidth]{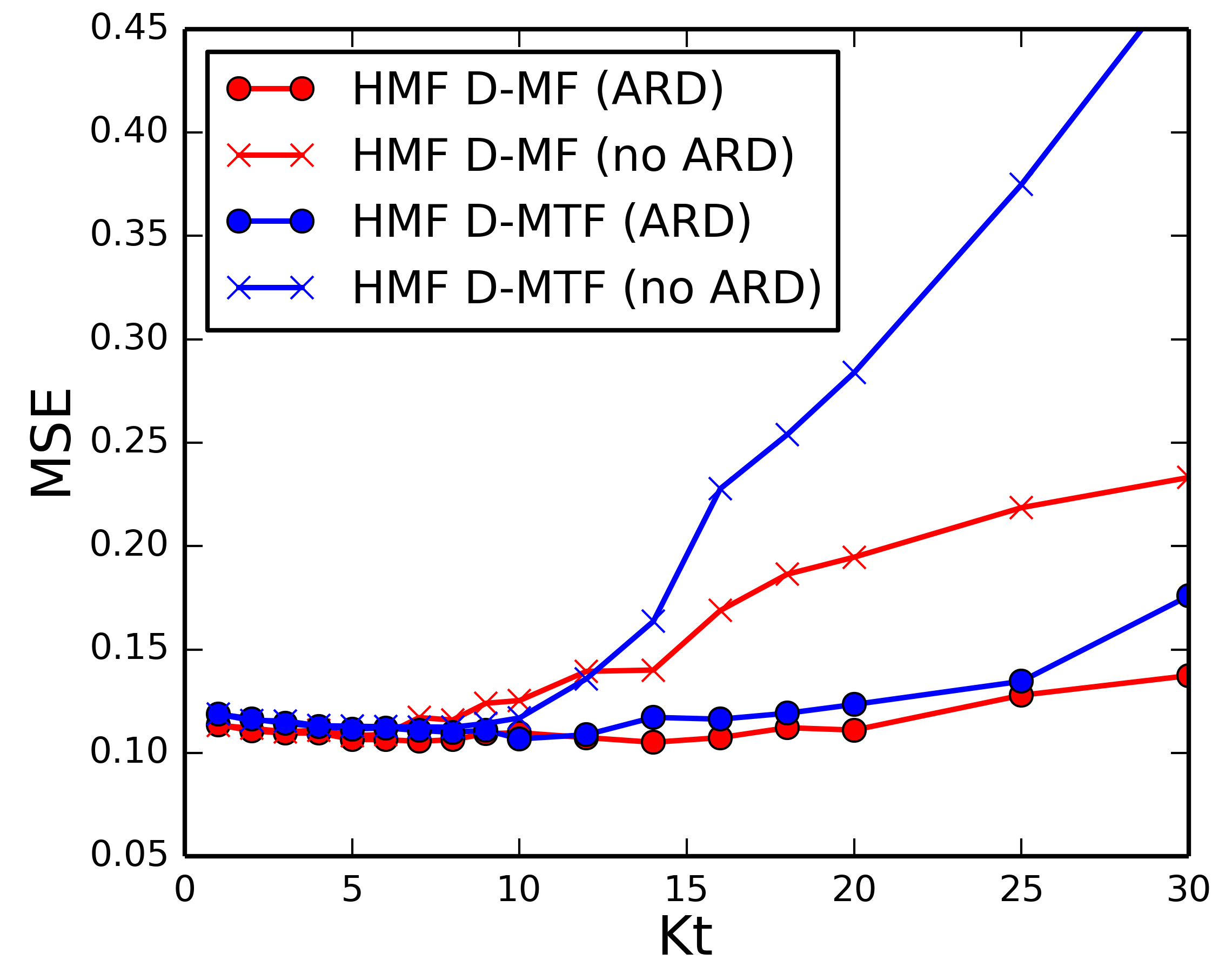}
				\captionsetup{width=0.95\columnwidth}
				\caption{CCLE $EC_{50}$} 
				\label{model_selection_ccle_ec}
			\end{subfigure}
			\captionsetup{width=0.98\columnwidth}
			\caption{Graphs showing the cross-validation performance of in-matrix predictions on the drug sensitivity datasets, where we vary the dimensionality $K_t$ for our HMF models (HMF D-MF in red, HMF D-MTF in blue), both for HMF with ARD (o), and without (x). Adding ARD clearly reduces overfitting as $K_t$ increases.}
			\label{model_selection}
		\end{figure}

		\noindent We compare the effects of adding ARD to the HMF model in Figure \ref{model_selection}. Here, we clearly see that adding ARD to the model consistently reduces overfitting on all four drug sensitivity datasets. ARD is not perfect, and we can still see that the curve goes up as the values for $K_t$ increase, but this overfitting is significantly less severe than the models without ARD.
		
		\paragraph{Usage recommendations} Always use ARD to reduce overfitting in the model. Even though ARD does not always entirely eliminate the need for model selection (there is still some overfitting as $K_t$ becomes very large), it generally makes it much less critical to try a large range of dimensionalities to find the best one. Instead, trying one or a couple will prove just as effective.
		
	\subsection{Initialisation} \label{Initialisation experiment}
		The initialisation method can have a huge impact on performance. Initialise too well, and it leads to overfitting very quickly. Initialise poorly, and your model may not converge to a good solution. \\
		
		\noindent As discussed in Section \ref{Initialisation}, there are several ways to initialise the Gibbs sampling parameter values. Here, we measure the convergence of 
		the HMF D-MF and HMF D-MTF models (nonnegative shared factors, real-valued private factors) on the drug sensitivity datasets. We try the following initialisation approaches:
		\begin{enumerate}
			\item \textbf{Exp:} All parameters initialised using expectation.
			\item \textbf{Random:} ARD initialised using expectation, all other parameters using random draws.
			\item \textbf{$\boldsymbol K$-means, exp:} Entity type factor matrices $F^t$ initialised using $K$-means, all other parameters using expectation.
			\item \textbf{$\boldsymbol K$-means, random:} Entity type factor matrices $F^t$ initialised using $K$-means, ARD initialised using expectation, and all other factor matrices using random draws.
			\item \textbf{Exp, least squares:} Entity type factor matrices $F^t$ and ARD initialised using expectation, and all other factor matrices using least squares.
			\item \textbf{Random, least squares:} Entity type factor matrices $F^t$ initialised using random draws, ARD initialised using expectation, and all other factor matrices using least squares.
			\item \textbf{$\boldsymbol K$-means, least squares:} Entity type factor matrices $F^t$ initialised using $K$-means, ARD initialised using expectation, and all other factor matrices using least squares.
		\end{enumerate}
		
		\begin{figure}[t!]
			\centering
			\begin{subfigure}[t]{0.35 \columnwidth}
				\includegraphics[width=1\columnwidth]{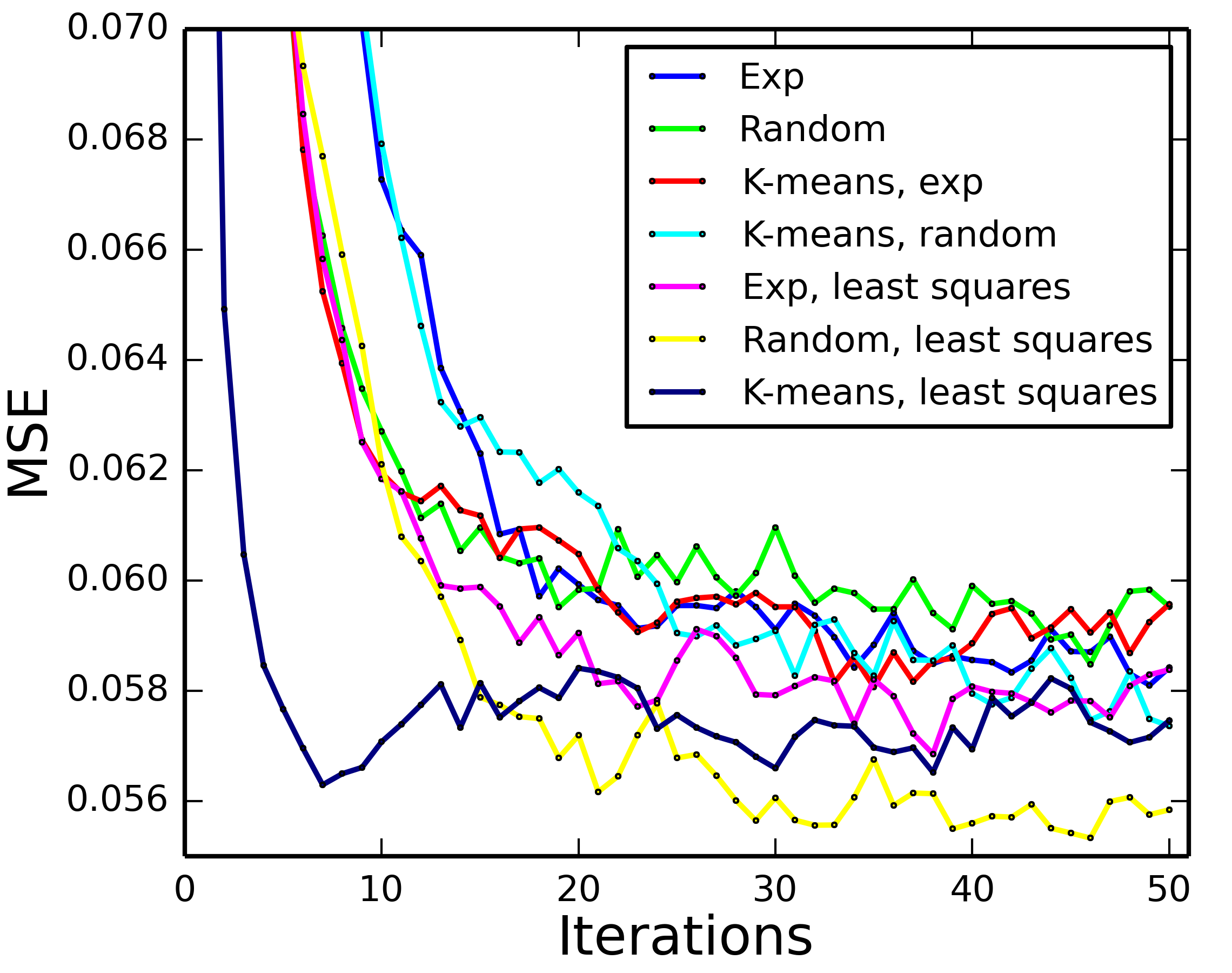}
				\captionsetup{width=\columnwidth}
				\caption{HMF D-MF, GDSC $IC_{50}$} 
			\end{subfigure}
			\begin{subfigure}[t]{0.35 \columnwidth}
				\includegraphics[width=1\columnwidth]{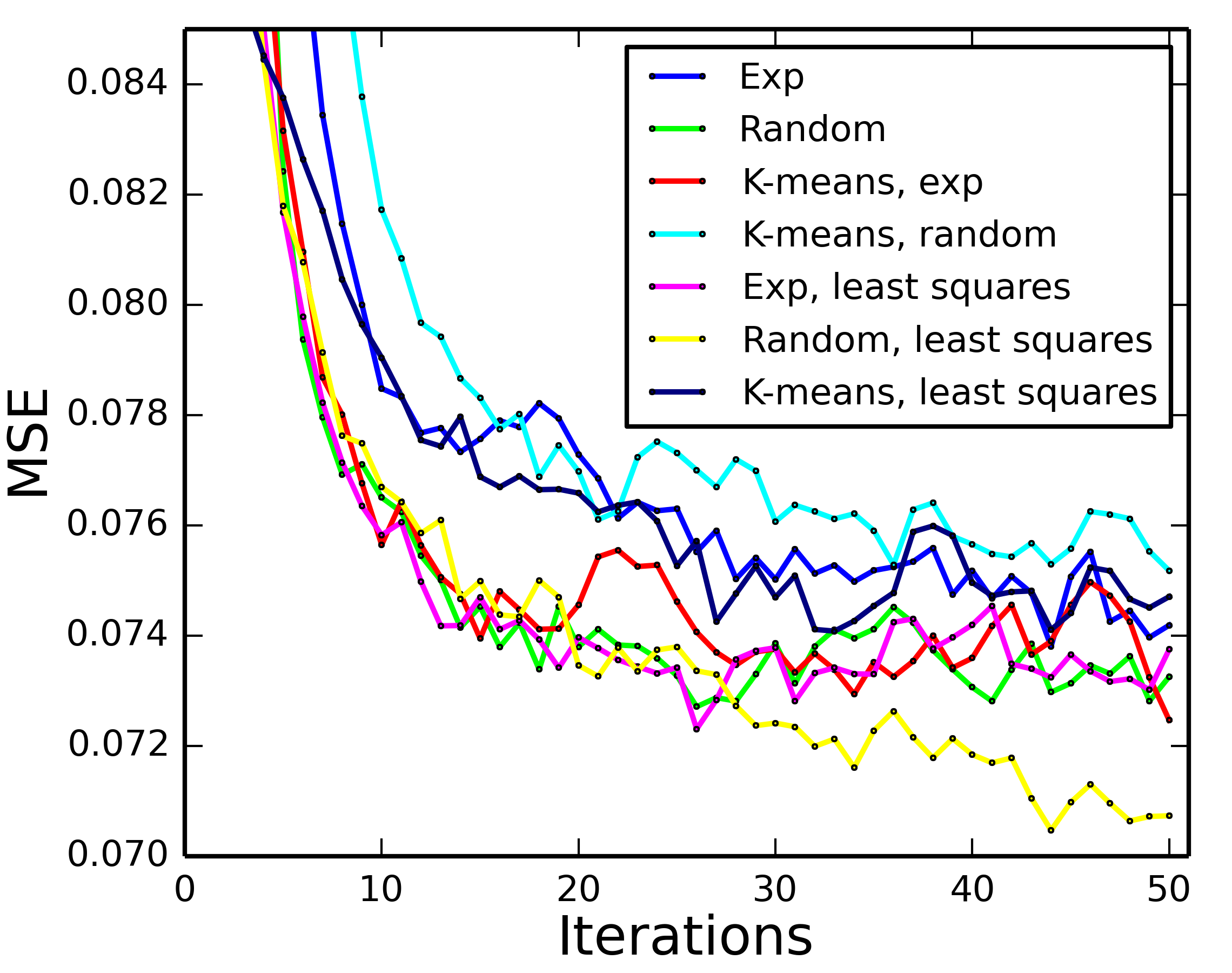}
				\captionsetup{width=\columnwidth}
				\caption{HMF D-MF, CTRP $EC_{50}$} 
			\end{subfigure}
			\begin{subfigure}[t]{0.35 \columnwidth}
				\includegraphics[width=1\columnwidth]{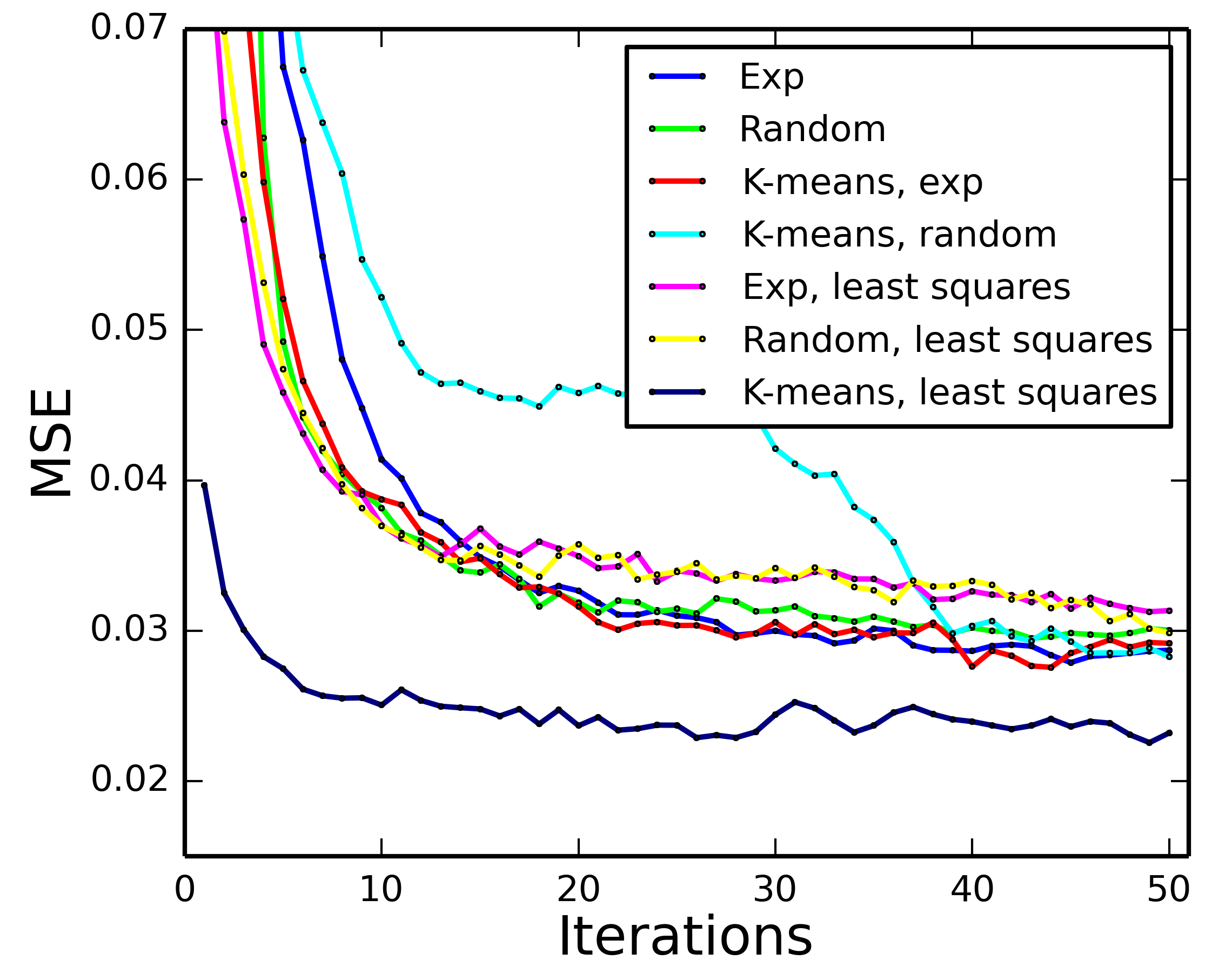}
				\captionsetup{width=\columnwidth}
				\caption{HMF D-MF, CCLE $IC_{50}$} 
			\end{subfigure}
			\begin{subfigure}[t]{0.35 \columnwidth}
				\includegraphics[width=1\columnwidth]{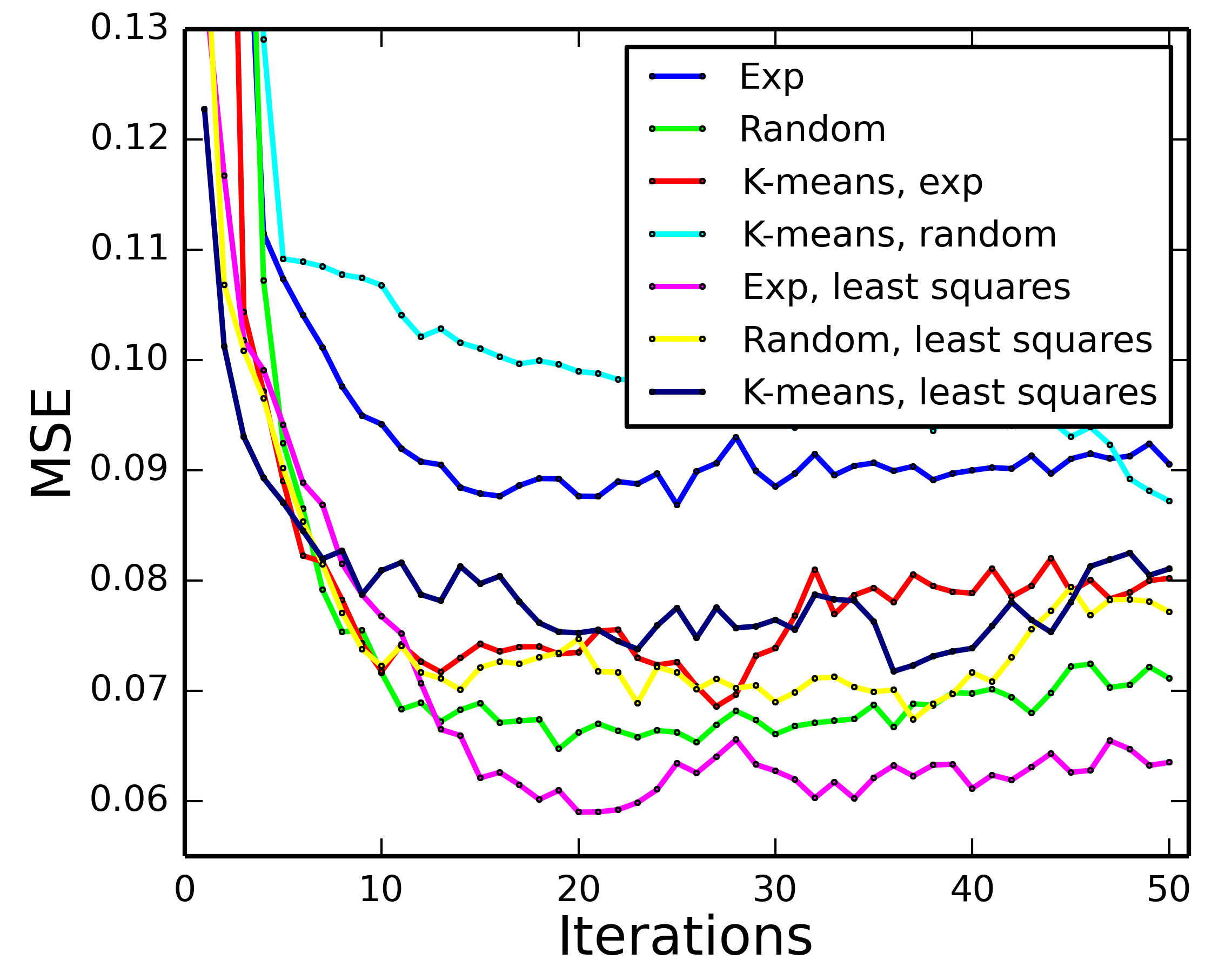}
				\captionsetup{width=\columnwidth}
				\caption{HMF D-MF, CCLE $EC_{50}$} 
			\end{subfigure}
			\begin{subfigure}[t]{0.35 \columnwidth}
				\includegraphics[width=1\columnwidth]{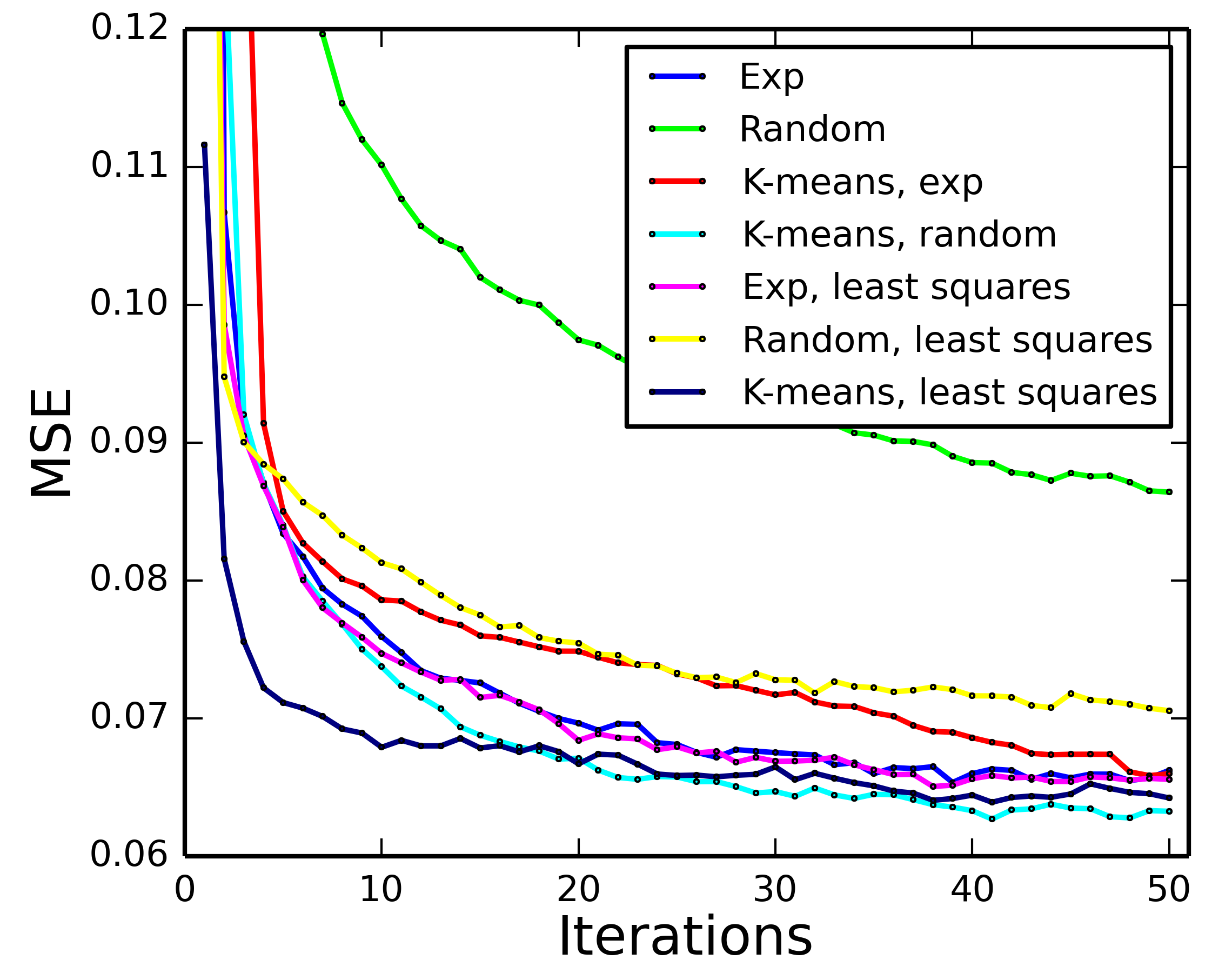}
				\captionsetup{width=\columnwidth}
				\caption{HMF D-MTF, GDSC $IC_{50}$} 
			\end{subfigure}
			\begin{subfigure}[t]{0.35 \columnwidth}
				\includegraphics[width=1\columnwidth]{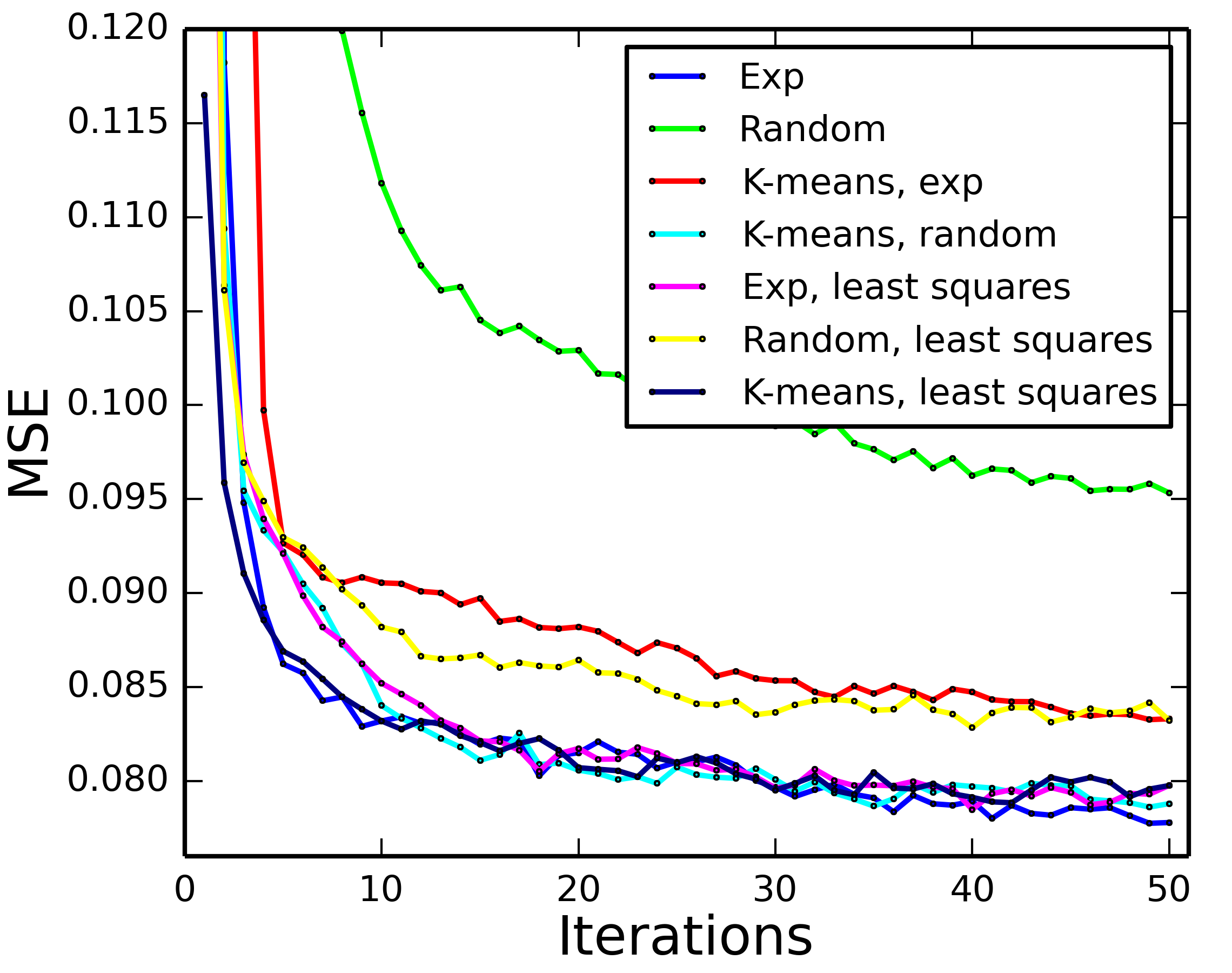}
				\captionsetup{width=\columnwidth}
				\caption{HMF D-MTF, CTRP $EC_{50}$} 
			\end{subfigure}
			\begin{subfigure}[t]{0.35 \columnwidth}
				\includegraphics[width=1\columnwidth]{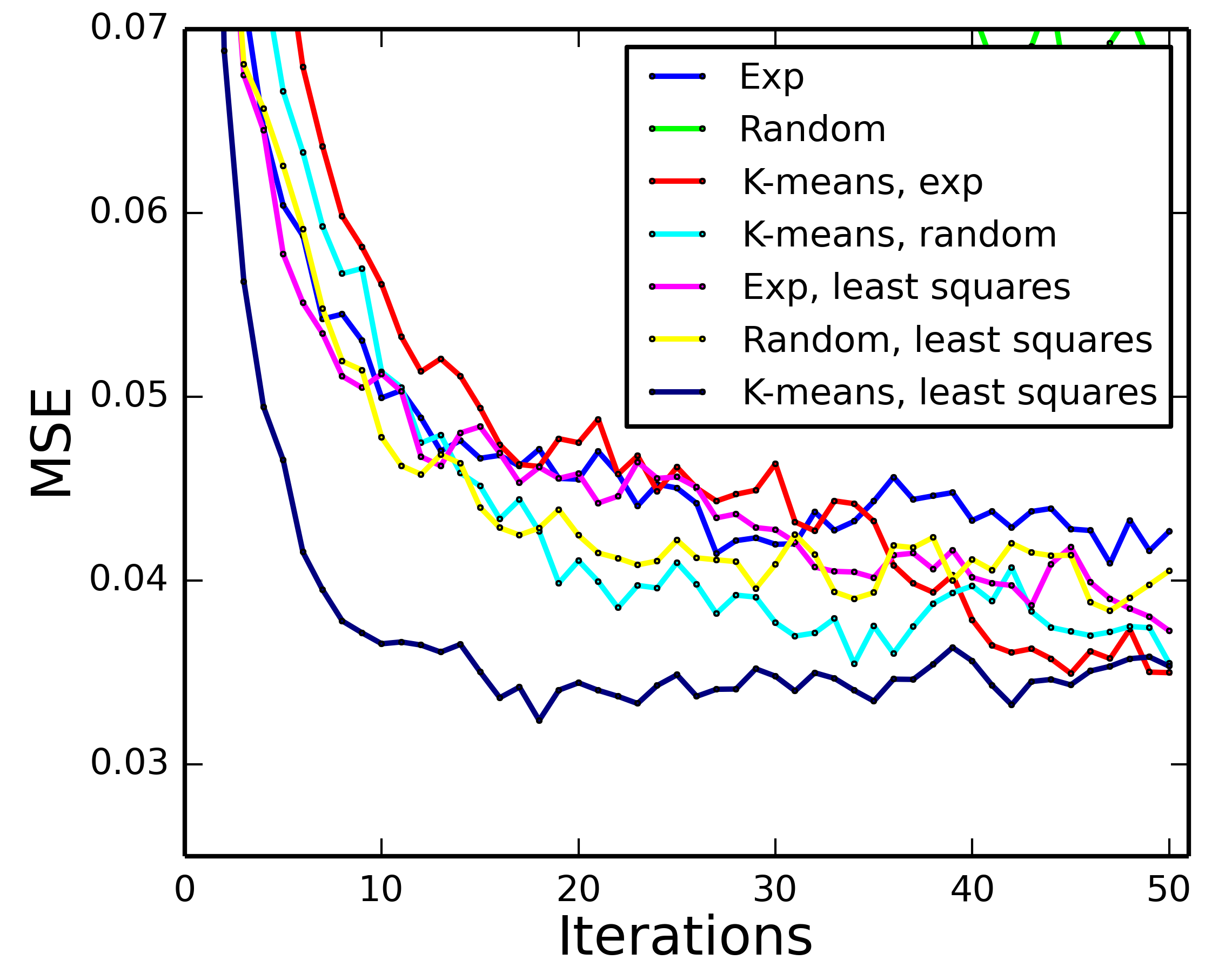}
				\captionsetup{width=\columnwidth}
				\caption{HMF D-MTF, CCLE $IC_{50}$} 
			\end{subfigure}
			\begin{subfigure}[t]{0.35 \columnwidth}
				\includegraphics[width=1\columnwidth]{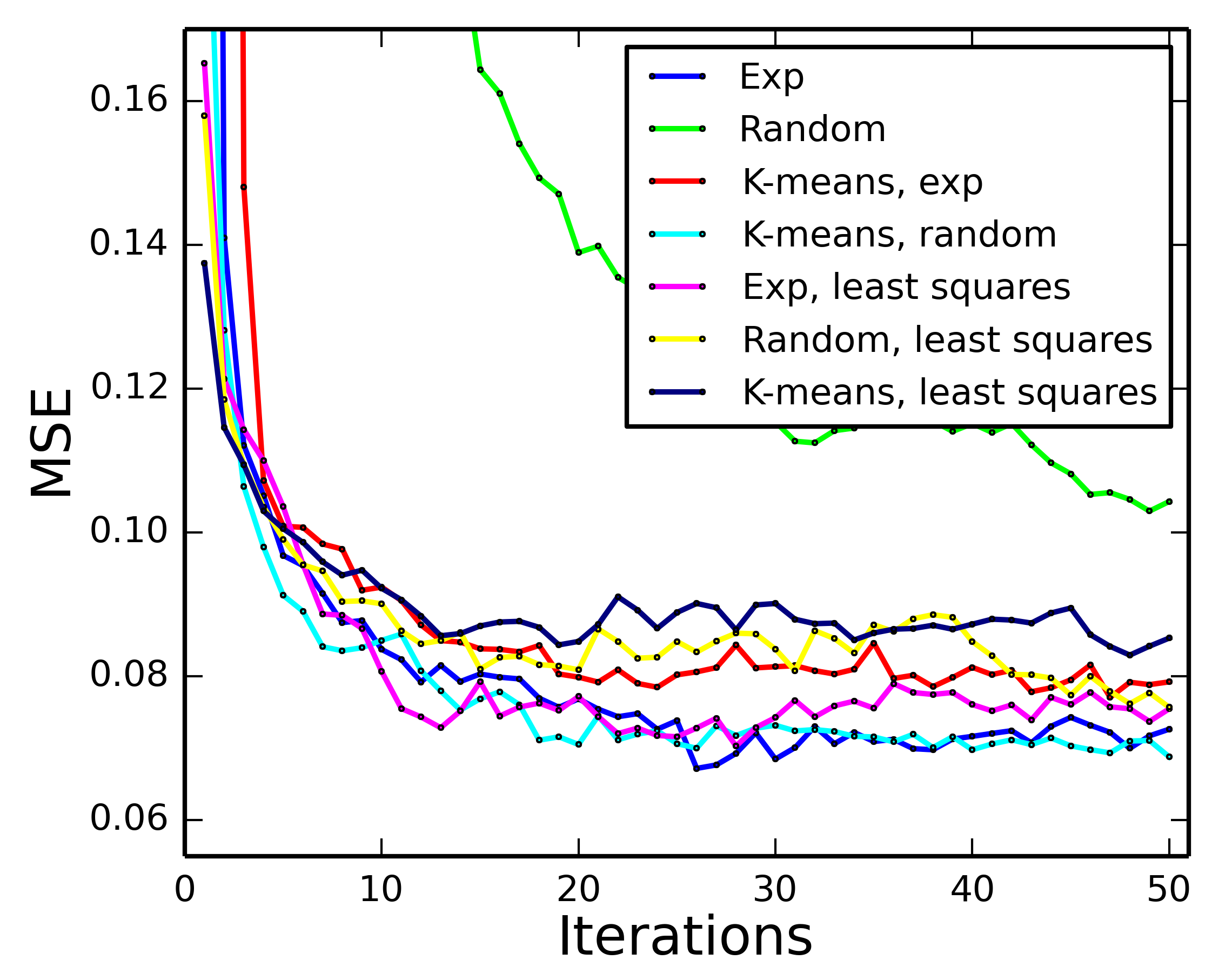}
				\captionsetup{width=\columnwidth}
				\caption{HMF D-MTF, CCLE $EC_{50}$} 
			\end{subfigure}
			\captionsetup{width=0.98\columnwidth}
			\caption{Graphs showing the convergence of the HMF D-MF (top two rows) and HMF D-MTF (bottom two rows) models on the four drug sensitivity datasets, for the seven different initialisation approaches.}
			\label{init_figure}
		\end{figure}
		
		\noindent The plots of convergence for HMF D-MF and HMF D-MTF are given in Figure \ref{init_figure}. We can see that the $K$-means with least squares initialisation strategy (dark blue) provides the fastest convergence on half of the datasets, oftentimes significantly faster. Random for $\F^t$ and least squares also performs well. The other strategies sometimes provide faster convergence, but none of them do so consistently.
		
		\paragraph{Recommendation} The fastest convergence is generally provided by combining $K$-means initialisation for $\F^t$ with least squares for the other factor matrices. Random initialisation for $\F^t$ with least squares also works well.

	\subsection{Importance value}
		We experimented with different values for the importance values $\alpha$, for the out-of-matrix prediction setting. Specifically, we consider the case where we predict one dataset (either gene expression, promoter-region methylation, or gene body methylation) using the other two datasets as additional datasets. \\
		
		\noindent We vary the value of $\alpha$ for the dataset we are trying to predict ($\alpha_0$) as well as for the two other datasets we are learning from ($\alpha_1, \alpha_2$), using the values $\left[ 0.25, 0.5, 1.0, 1.5, 2.0 \right]$. We use $K_t = 10$ for all experiments, non-negative factors for the shared factor matrices, and real-valued for the private ones. For initialisation we use $K$-means for the shared factor matrices, and least squares for the private (real-valued) ones. \\
		
		\noindent We perform 10-fold cross-validation, taking out 10\% of the samples each time for the dataset we are trying to predict, and then measuring the mean squared error (MSE) of predictions. The average performances can be found in Table \ref{varying_importance_results_hmf_d_mf_mtf}, for both HMF D-MF (multiple matrix factorisation) and HMF D-MTF (multiple matrix tri-factorisation). \\
		
		\noindent For the HMF D-MF model (left column) we see that datasets with low predictivity (such as GE and PM -- note the high MSE) have the best parameter values when the importance of the dataset to be predicted ($\alpha_0$) is low, and the importance of the datasets to learn from ($\alpha_1, \alpha_2$) is high. This is presumably because higher importance values lead to a better fit to the data, and if there is low predictivity, we should not fit to the data too much (otherwise we might overfit). 
		In contrast, when the predictivity is high (such as GM), the importance value for all datasets should not be too low, because this results in a poor fit to the data and hence poor predictions. \\
		
		\noindent For the HMF D-MTF model (right column) we see a similar effect, in that if the predictivity is better, the best values for the importance increase. However, for this approach all of the importance values should generally be set low if the datasets are different. \\
		
		\noindent We conducted the same experiment for the approach based on similarity kernels, where we use the same ones as used in the out-of-matrix predictions from the paper. For the dataset we are trying to predict we decompose it using matrix factorisation (HMF S-MF). Matrix tri-factorisation could have also been chosen, but because the third dataset is not shared in this case, it is equivalent to matrix factorisation (in fact, we get very similar tables as the ones shown for HMF S-MF). As before, we use nonnegative factors for the shared matrices, real-valued factors for the private ones, and $K$-means and least squares for initialisation. The results can be found in Table \ref{varying_importance_results_hmf_s_mf_mtf}. Here, we see that the importance value is much less important, as long as it is not lower than 1.0.
		
		\paragraph{Recommendation}
		If the datasets are very dissimilar and have low predictivity, use a low importance value for the main dataset for which we are trying to predict values (like 0.5). When using HMF D-MF, use a higher importance value for the other datasets (like 1.5), but when using HMF D-MTF, use a low importance value as well (like 0.5). If the datasets are more similar, use normal importance values (1.0) for all datasets. Finally, when using similarity kernels (HMF S-MF), use the normal importance value for the kernels (1.0).
		
		\begin{table*}[h!]
			\captionsetup{width=0.975\columnwidth}
			\caption{Performances of out-of-matrix cross-validation results for HMF D-MF (left column) and D-MTF (right column), where we vary the importance value for the dataset we are trying to predict ($\alpha_0$), and for the other two datasets we are learning from ($\alpha_1, \alpha_2$). We have three different datasets (gene expression, GE; gene body methylation, GM; and promoter region methylation, PM). We therefore have three different prediction settings. We have highlighted the most promising parameter value areas in green, and the least promising in red.}
			\vspace{5pt}
			\label{varying_importance_results_hmf_d_mf_mtf}
			\centering
			\begin{tabular}{cccccc}
				\toprule
				\multicolumn{6}{c}{HMF D-MF, GM + PM $\rightarrow$ GE}	\\
				\midrule
				& \multicolumn{5}{c}{GM ($\alpha_1$), PM ($\alpha_2$)}	\\
				\cmidrule(lr){2-6}
				GE ($\alpha_0$) & 0.25 & 0.5 & 1.0 & 1.5 & 2.0	\\
				\cmidrule(lr){1-1} \cmidrule(lr){2-6}
				0.25	& 	\cellcolor{yellow!25}0.878	&	\cellcolor{yellow!25}0.843	& 	\cellcolor{green!25}0.835	&	\cellcolor{green!25}0.831	& 	\cellcolor{green!25}0.832	\\
				0.5		& 	\cellcolor{yellow!25}0.845	&	\cellcolor{yellow!25}0.849	& 	\cellcolor{green!25}0.829	&	\cellcolor{green!25}0.832	& 	\cellcolor{green!25}0.829	\\
				1.0 	&  	\cellcolor{yellow!25}0.848	&	\cellcolor{red!25}0.916	& 	\cellcolor{red!25}1.195	&	\cellcolor{red!25}1.322	& 	\cellcolor{red!25}0.831	\\
				1.5		& 	\cellcolor{yellow!25}0.871	&	\cellcolor{red!25}0.948	& 	\cellcolor{red!25}1.288	&	\cellcolor{red!25}1.407	& 	\cellcolor{red!25}1.470	\\
				2.0		&	\cellcolor{yellow!25}0.896	&	\cellcolor{red!25}0.966	& 	\cellcolor{red!25}1.392	&	\cellcolor{red!25}1.711	& 	\cellcolor{red!25}1.782	\\
				\bottomrule
				&&&&&\\
				&&&&&\\
			\end{tabular}
			\quad
			\begin{tabular}{cccccc}
				\toprule
				\multicolumn{6}{c}{HMF D-MTF, GM + PM $\rightarrow$ GE}	\\
				\midrule
				& \multicolumn{5}{c}{GM ($\alpha_1$), PM ($\alpha_2$)}	\\
				\cmidrule(lr){2-6}
				GE ($\alpha_0$) & 0.25 & 0.5 & 1.0 & 1.5 & 2.0	\\
				\cmidrule(lr){1-1} \cmidrule(lr){2-6}
				0.25	& 	\cellcolor{green!25}0.866	&	\cellcolor{yellow!25}0.906	& 	\cellcolor{yellow!25}0.941	&	\cellcolor{yellow!25}0.950	& 	\cellcolor{yellow!25}0.954	\\
				0.5		& 	\cellcolor{green!25}0.874	&	\cellcolor{green!25}0.870	& 	\cellcolor{yellow!25}0.914	&	\cellcolor{yellow!25}0.941	& 	\cellcolor{yellow!25}0.945	\\
				1.0 	&  	\cellcolor{yellow!25}0.933	&	\cellcolor{red!25}0.985	& 	\cellcolor{yellow!25}0.940	&	\cellcolor{yellow!25}0.942	& 	\cellcolor{yellow!25}0.961	\\
				1.5		& 	\cellcolor{yellow!25}0.959	&	\cellcolor{yellow!25}0.958	& 	\cellcolor{red!25}1.065	&	\cellcolor{red!25}1.026	& 	\cellcolor{red!25}1.000	\\
				2.0		&	\cellcolor{yellow!25}0.957	& 	\cellcolor{red!25}1.080	&	\cellcolor{red!25}1.211	& 	\cellcolor{red!25}1.145	&	\cellcolor{red!25}1.147	\\
				\bottomrule
				&&&&&\\
				&&&&&\\
			\end{tabular}
			\begin{tabular}{cccccc}
				\toprule
				\multicolumn{6}{c}{HMF D-MF, GE + GM $\rightarrow$ PM}	\\
				\midrule
				& \multicolumn{5}{c}{GE ($\alpha_1$), GM ($\alpha_2$)}	\\
				\cmidrule(lr){2-6}
				PM ($\alpha_0$) & 0.25 & 0.5 & 1.0 & 1.5 & 2.0	\\
				\cmidrule(lr){1-1} \cmidrule(lr){2-6}
				0.25	& 	\cellcolor{yellow!25}0.859	&	\cellcolor{green!25}0.799	& 	\cellcolor{green!25}0.795	&	\cellcolor{green!25}0.799	& 	\cellcolor{green!25}0.798	\\
				0.5		& 	\cellcolor{yellow!25}0.811	&	\cellcolor{yellow!25}0.812	& 	\cellcolor{green!25}0.783	&	\cellcolor{green!25}0.783	& 	\cellcolor{green!25}0.784	\\
				1.0 	&  	\cellcolor{green!25}0.789	&	\cellcolor{yellow!25}0.814	& 	\cellcolor{red!25}0.987	&	\cellcolor{green!25}0.788	& 	\cellcolor{green!25}0.783	\\
				1.5		& 	\cellcolor{green!25}0.784	&	\cellcolor{yellow!25}0.809	& 	\cellcolor{red!25}1.070	&	\cellcolor{red!25}1.247	& 	\cellcolor{yellow!25}0.870	\\
				2.0		&	\cellcolor{green!25}0.798	&	\cellcolor{yellow!25}0.835	& 	\cellcolor{red!25}1.111	&	\cellcolor{red!25}1.280	& 	\cellcolor{red!25}1.255	\\
				\bottomrule
				&&&&&\\
				&&&&&\\
			\end{tabular}
			\quad
			\begin{tabular}{cccccc}
				\toprule
				\multicolumn{6}{c}{HMF D-MTF, GE + GM $\rightarrow$ PM}	\\
				\midrule
				& \multicolumn{5}{c}{GE ($\alpha_1$), GM ($\alpha_2$)}	\\
				\cmidrule(lr){2-6}
				PM ($\alpha_0$) & 0.25 & 0.5 & 1.0 & 1.5 & 2.0	\\
				\cmidrule(lr){1-1} \cmidrule(lr){2-6}
				0.25	& 	\cellcolor{green!25}0.862	&	\cellcolor{yellow!25}0.893	& 	\cellcolor{red!25}0.943	&	\cellcolor{red!25}0.946	& 	\cellcolor{red!25}0.947	\\
				0.5		& 	\cellcolor{green!25}0.841	&	\cellcolor{yellow!25}0.885	& 	\cellcolor{yellow!25}0.898	&	\cellcolor{red!25}0.943	& 	\cellcolor{red!25}0.944	\\
				1.0 	&  	\cellcolor{green!25}0.852	&	\cellcolor{green!25}0.848	& 	\cellcolor{red!25}1.012	&	\cellcolor{yellow!25}0.885	& 	\cellcolor{red!25}0.904	\\
				1.5		& 	\cellcolor{green!25}0.876	&	\cellcolor{green!25}0.866	& 	\cellcolor{red!25}0.900	&	\cellcolor{red!25}1.080	& 	\cellcolor{red!25}0.904	\\
				2.0		&	\cellcolor{yellow!25}0.884	&	\cellcolor{yellow!25}0.881	& 	\cellcolor{red!25}0.900	&	\cellcolor{red!25}1.096	& 	\cellcolor{red!25}1.099	\\
				\bottomrule
				&&&&&\\
				&&&&&\\
			\end{tabular}
			\begin{tabular}{cccccc}
				\toprule
				\multicolumn{6}{c}{HMF D-MF, GE + PM $\rightarrow$ GM}	\\
				\midrule
				& \multicolumn{5}{c}{GE ($\alpha_1$), PM ($\alpha_2$)}	\\
				\cmidrule(lr){2-6}
				GM ($\alpha_0$) & 0.25 & 0.5 & 1.0 & 1.5 & 2.0	\\
				\cmidrule(lr){1-1} \cmidrule(lr){2-6}
				0.25	& 	\cellcolor{red!25}0.790	&	\cellcolor{red!25}0.708	& 	\cellcolor{red!25}0.703	&	\cellcolor{red!25}0.697	& 	\cellcolor{red!25}0.701	\\
				0.5		& 	\cellcolor{red!25}0.724	&	\cellcolor{yellow!25}0.670	& 	\cellcolor{yellow!25}0.687	&	\cellcolor{yellow!25}0.688	& 	\cellcolor{yellow!25}0.685	\\
				1.0 	&  	\cellcolor{red!25}0.698	&	\cellcolor{green!25}0.657	& 	\cellcolor{green!25}0.659	&	\cellcolor{yellow!25}0.670	& 	\cellcolor{yellow!25}0.685	\\
				1.5		& 	\cellcolor{red!25}0.698	&	\cellcolor{green!25}0.655	& 	\cellcolor{green!25}0.667	&	\cellcolor{green!25}0.657	& 	\cellcolor{green!25}0.666	\\
				2.0		&	\cellcolor{yellow!25}0.689	&	\cellcolor{green!25}0.668	& 	\cellcolor{yellow!25}0.671	&	\cellcolor{yellow!25}0.676	& 	\cellcolor{green!25}0.665	\\
				\bottomrule
			\end{tabular}
			\quad
			\begin{tabular}{cccccc}
				\toprule
				\multicolumn{6}{c}{HMF D-MTF, GE + PM $\rightarrow$ GM}	\\
				\midrule
				& \multicolumn{5}{c}{GE ($\alpha_1$), PM ($\alpha_2$)}	\\
				\cmidrule(lr){2-6}
				GM ($\alpha_0$) & 0.25 & 0.5 & 1.0 & 1.5 & 2.0	\\
				\cmidrule(lr){1-1} \cmidrule(lr){2-6}
				0.25	& 	\cellcolor{yellow!25}0.773	&	\cellcolor{yellow!25}0.775	& 	\cellcolor{red!25}0.851	&	\cellcolor{red!25}0.854	& 	\cellcolor{red!25}0.861	\\
				0.5		& 	\cellcolor{yellow!25}0.761	&	\cellcolor{green!25}0.746	& 	\cellcolor{yellow!25}0.774	&	\cellcolor{red!25}0.818	& 	\cellcolor{red!25}0.850	\\
				1.0 	&  	\cellcolor{yellow!25}0.782	&	\cellcolor{green!25}0.757	& 	\cellcolor{green!25}0.754	&	\cellcolor{yellow!25}0.786	& 	\cellcolor{yellow!25}0.779	\\
				1.5		& 	\cellcolor{red!25}0.832	&	\cellcolor{green!25}0.752	& 	\cellcolor{green!25}0.745	&	\cellcolor{yellow!25}0.783	& 	\cellcolor{red!25}0.795	\\
				2.0		&	\cellcolor{red!25}0.836	&	\cellcolor{red!25}0.811	& 	\cellcolor{red!25}0.802	&	\cellcolor{red!25}0.791	& 	\cellcolor{red!25}0.805	\\
				\bottomrule
			\end{tabular}
		\end{table*}

		\begin{table*}[h!]
			\captionsetup{width=\columnwidth}
			\caption{Performances of out-of-matrix cross-validation results for HMF S-MF, where we vary the importance value for the dataset we are trying to predict ($\alpha_0$), and for the other two datasets we are learning from ($\alpha_1, \alpha_2$). We have three different datasets (gene expression, GE; gene body methylation, GM; and promoter region methylation, PM). We therefore have three different prediction settings. We have highlighted the most promising parameter value areas in green, and the least promising in red.}
			\vspace{5pt}
			\label{varying_importance_results_hmf_s_mf_mtf}
			\centering
			\begin{tabular}{cccccc}
				\toprule
				\multicolumn{6}{c}{HMF S-MF, GM + PM $\rightarrow$ GE}	\\
				\midrule
				& \multicolumn{5}{c}{GM ($\alpha_1$), PM ($\alpha_2$)}	\\
				\cmidrule(lr){2-6}
				GE ($\alpha_0$) & 0.25 & 0.5 & 1.0 & 1.5 & 2.0	\\
				\cmidrule(lr){1-1} \cmidrule(lr){2-6}
				0.25	&	\cellcolor{red!25}0.873	& 	\cellcolor{red!25}0.870	&	\cellcolor{red!25}0.871	&	\cellcolor{red!25}0.880	&	\cellcolor{red!25}0.879	\\
				0.5		& 	\cellcolor{yellow!25}0.852	&	\cellcolor{yellow!25}0.850	&	\cellcolor{yellow!25}0.853	&	\cellcolor{yellow!25}0.852	&	\cellcolor{green!25}0.849	\\
				1.0 	& 	\cellcolor{green!25}0.839	&	\cellcolor{green!25}0.835	&	\cellcolor{green!25}0.846	&	\cellcolor{green!25}0.843	&	\cellcolor{green!25}0.842	\\
				1.5		& 	\cellcolor{red!25}0.872	&	\cellcolor{green!25}0.839	&	\cellcolor{green!25}0.837	&	\cellcolor{green!25}0.842	&	\cellcolor{green!25}0.840	\\
				2.0		& 	\cellcolor{red!25}0.945	&	\cellcolor{green!25}0.849	&	\cellcolor{green!25}0.834	&	\cellcolor{green!25}0.845	&	\cellcolor{green!25}0.833	\\
				\bottomrule
				&&&&&\\
				&&&&&\\
			\end{tabular}
			\quad
			\begin{tabular}{cccccc}
				\toprule
				\multicolumn{6}{c}{HMF S-MF, GE + GM $\rightarrow$ PM}	\\
				\midrule
				& \multicolumn{5}{c}{GM ($\alpha_1$), PM ($\alpha_2$)}	\\
				\cmidrule(lr){2-6}
				GE ($\alpha_0$) & 0.25 & 0.5 & 1.0 & 1.5 & 2.0	\\
				\cmidrule(lr){1-1} \cmidrule(lr){2-6}
				0.25	& 	\cellcolor{red!25}0.858	&	\cellcolor{red!25}0.858	&	\cellcolor{red!25}0.866	&	\cellcolor{red!25}0.859	&	\cellcolor{red!25}0.854	\\
				0.5		& 	\cellcolor{yellow!25}0.829	&	\cellcolor{yellow!25}0.830	&	\cellcolor{yellow!25}0.823	&	\cellcolor{yellow!25}0.832	&	\cellcolor{yellow!25}0.828	\\
				1.0 	& 	\cellcolor{yellow!25}0.824	&	\cellcolor{green!25}0.813	&	\cellcolor{green!25}0.815	&	\cellcolor{green!25}0.814	&	\cellcolor{green!25}0.814	\\
				1.5		& 	\cellcolor{red!25}0.853	&	\cellcolor{green!25}0.804	&	\cellcolor{green!25}0.806	&	\cellcolor{green!25}0.812	&	\cellcolor{green!25}0.808	\\
				2.0		& 	\cellcolor{red!25}0.898	&	\cellcolor{red!25}0.858	&	\cellcolor{green!25}0.810	&	\cellcolor{green!25}0.807	&	\cellcolor{green!25}0.809	\\
				\bottomrule
				&&&&&\\
				&&&&&\\
			\end{tabular}
			\begin{tabular}{cccccc}
				\toprule
				\multicolumn{6}{c}{HMF S-MF, GE + PM $\rightarrow$ GM}	\\
				\midrule
				& \multicolumn{5}{c}{GM ($\alpha_1$), PM ($\alpha_2$)}	\\
				\cmidrule(lr){2-6}
				GE ($\alpha_0$) & 0.25 & 0.5 & 1.0 & 1.5 & 2.0	\\
				\cmidrule(lr){1-1} \cmidrule(lr){2-6}
				0.25	& 	\cellcolor{red!25}0.762	&	\cellcolor{red!25}0.771	&	\cellcolor{red!25}0.782	&	\cellcolor{red!25}0.796	&	\cellcolor{red!25}0.797	\\
				0.5		& 	\cellcolor{yellow!25}0.722	&	\cellcolor{yellow!25}0.724	&	\cellcolor{yellow!25}0.740	&	\cellcolor{yellow!25}0.742	&	\cellcolor{red!25}0.753	\\
				1.0 	& 	\cellcolor{green!25}0.701	&	\cellcolor{green!25}0.706	&	\cellcolor{green!25}0.707	&	\cellcolor{yellow!25}0.718	&	\cellcolor{yellow!25}0.721	\\
				1.5		& 	\cellcolor{red!25}0.756	&	\cellcolor{green!25}0.703	&	\cellcolor{green!25}0.702	&	\cellcolor{green!25}0.709	&	\cellcolor{green!25}0.710	\\
				2.0		& 	\cellcolor{red!25}0.757	&	\cellcolor{green!25}0.710	&	\cellcolor{green!25}0.702	&	\cellcolor{green!25}0.697	&	\cellcolor{green!25}0.701	\\
				\bottomrule
				&&&&&\\
				&&&&&\\
			\end{tabular}
		\end{table*}
	
	\clearpage
	\subsection{Negativity constraints}
		Negativity constraints have two advantages. Firstly, they make it easier to analyse the factor values after performing matrix factorisation, and for example identify clusters in the data. With real-valued factors this can be a lot more complicated. Secondly, the nonnegativity constraints can reduce overfitting on sparse or noisy datasets. \\
	
		\noindent We measured the effects of nonnegativity constraints on our HMF models. On the drug sensitivity datasets we tried three variants: nonnegative (all factor matrices are nonnegative), real-valued (all factor matrices are real-valued), and semi-nonnegative (shared factor matrices are nonnegative, dataset-specific ones are real-valued). On the gene expression and methylation datasets we tried the real-valued and semi-nonnegative versions. For the variants containing real-valued factor matrices, we also see whether there is a difference between row-wise and column-wise posterior draws. \\
	
		\noindent We ran 10-fold cross-validation for in-matrix predictions of the drug sensitivity datasets. We use $K_t = 10$, $K$-means initialisation for the $\F^t$ matrices, and least squares initialisation for the other matrices. Results are given in Table \ref{varying_nonnegativity_sensitivity_hmf_d_mf_mtf}, where we see that the entirely real-valued models consistently outperform all other versions, for both HMF D-MF and D-MTF. In addition, row-wise draws perform better than column-wise ones. \\
		
		\noindent Similarly, we ran 10-fold cross-validation for out-of-matrix predictions of the drug sensitivity datasets. We ran the HMF D-MF, HMF D-MTF, and HMF S-MF models, again with $K_t = 10$, and $K$-means and least squares initialisation. For HMF D-MF we used importance value 0.5 for all three datasets. For HMF D-MTF, 1.5 for the main dataset we are trying to predict, and 0.5 for the others. For HMF S-MF we used 1.0 for all datasets. These results are given in Table \ref{varying_nonnegativity_methylation_hmf_d_mf_mtf}, showing again that the real-valued version performs better for HMF D-MF, D-MTF, and S-MF. Column draws sometime do best, but generally row-wise draws are the best options. \\
		
		\begin{table*}[t!]
			\captionsetup{width=1\columnwidth}
			\caption{Performances of in-matrix cross-validation results for HMF D-MF (top table) and D-MTF (bottom table) on the drug sensitivity datasets, where we vary the nonnegativity of the matrices. We try nonnegative, semi-nonnegative, and real-valued variants, and for the real-varied variants try both row-wise draws and column-wise draws. The best performances are highlighted in bold.}
			\vspace{5pt}
			\label{varying_nonnegativity_sensitivity_hmf_d_mf_mtf}
			\centering
			\begin{tabular}{llcccc}
				\multicolumn{2}{c}{\textbf{HMF D-MF}} & \multicolumn{4}{c}{MSE}	\\
				\cmidrule(lr){1-2} \cmidrule(lr){3-6}
				$\F_t$ & $\G_l$ & GDSC $IC_{50}$ & CTRP $EC_{50}$ & CCLE $IC_{50}$ & CCLE $EC_{50}$ \\
				\midrule
				Nonnegative 			& Nonnegative 			& 0.0783 & 0.0907 & 0.0544 &  0.1083 \\
				Nonnegative 			& Real-valued (row) 	& 0.0764 & 0.0899 & 0.0541 & 0.1069 \\
				Nonnegative 			& Real-valued (column) 	& 0.0760 & 0.0903 & 0.0572 & 0.1050 \\
				Real-valued (row) 		& Real-valued (row) 	& \textbf{0.0758} & \textbf{0.0878} & \textbf{0.0519} & \textbf{0.1028} \\
				Real-valued (column) 	& Real-valued (column) 	& 0.0761 & 0.0889 & 0.0521 & 0.1029 \\
				\bottomrule
				&&&&&\\
			\end{tabular}
			\begin{tabular}{llcccc}
				\multicolumn{2}{c}{\textbf{HMF D-MTF}} & \multicolumn{4}{c}{MSE}	\\
				\cmidrule(lr){1-2} \cmidrule(lr){3-6}
				$\F_t$ & $\S_n$ & GDSC $IC_{50}$ & CTRP $EC_{50}$ & CCLE $IC_{50}$ & CCLE $EC_{50}$ \\
				\midrule
				Nonnegative 			& Nonnegative 			& 0.0802 & 0.0916 & 0.0573 & 0.1133 \\
				Nonnegative 			& Real-valued (row) 	& 0.0773 & 0.0899 & 0.0558 & 0.1110 \\
				Nonnegative 			& Real-valued (column) 	& 0.0780 & 0.0904 & 0.0549 & 0.1116 \\
				Real-valued (row) 		& Real-valued (row) 	& \textbf{0.0770} & \textbf{0.0883} & \textbf{0.0535} & \textbf{0.1011} \\
				Real-valued (column) 	& Real-valued (column) 	& 0.0799 & 0.0904 & 0.0572 & 0.1101 \\
				\bottomrule
				&&&&&\\
			\end{tabular}
		\end{table*}
		\begin{table*}[t!]
			\captionsetup{width=1\columnwidth}
			\caption{Performances of out-of-matrix cross-validation results for HMF D-MF (top table), D-MTF (middle table), and S-MF (bottom table), on the gene expression and methylation datasets, where we vary the nonnegativity of the matrices. We try semi-nonnegative and real-valued variants, and also row-wise draws and column-wise draws. The best performances are highlighted in bold.}
			\vspace{5pt}
			\label{varying_nonnegativity_methylation_hmf_d_mf_mtf}
			\centering
			\begin{tabular}{llccc}
				\multicolumn{2}{c}{\textbf{HMF D-MF}} & \multicolumn{3}{c}{MSE}	\\
				\cmidrule(lr){1-2} \cmidrule(lr){3-5}
				$\F_t$ & $\G_l$ & GM, PM to GE & GE, GM to PM & GE, PM to GM \\
				\midrule
				Nonnegative 			& Real-valued (row) 	& 0.880 & 0.792 & 0.670 \\
				Nonnegative 			& Real-valued (column) 	& 0.851 & 0.803 & 0.672 \\
				Real-valued (row) 		& Real-valued (row) 	& \textbf{0.834} & 0.775 & 0.651 \\
				Real-valued (column) 	& Real-valued (column) 	& 0.839 & \textbf{0.769} & \textbf{0.647} \\
				\bottomrule
				&&&&\\
			\end{tabular}
			\begin{tabular}{llccc}
				\multicolumn{2}{c}{\textbf{HMF D-MTF}} & \multicolumn{3}{c}{MSE}	\\
				\cmidrule(lr){1-2} \cmidrule(lr){3-5}
				$\F_t$ & $\S_n$ & GM, PM to GE & GE, GM to PM & GE, PM to GM \\
				\midrule
				Nonnegative 			& Real-valued (row) 	& 0.959 & 0.864 & 0.755 \\
				Nonnegative 			& Real-valued (column) 	& 0.985 & 0.869 & 0.777 \\
				Real-valued (row) 		& Real-valued (row) 	& \textbf{0.893} & \textbf{0.822} & 0.756 \\
				Real-valued (column) 	& Real-valued (column) 	& 0.909 & 0.837 & \textbf{0.738} \\
				\bottomrule
				&&&&\\
			\end{tabular}
			\begin{tabular}{llccc}
				\multicolumn{2}{c}{\textbf{HMF S-MF}} & \multicolumn{3}{c}{MSE}	\\
				\cmidrule(lr){1-2} \cmidrule(lr){3-5}
				$\F_t$ & $\S_n$ & GM, PM to GE & GE, GM to PM & GE, PM to GM \\
				\midrule
				Nonnegative 			& Real-valued (row) 	& 0.846 & 0.818 & 0.713 \\
				Nonnegative 			& Real-valued (column) 	& 0.836 & 0.811 & 0.723 \\
				Real-valued (row) 		& Real-valued (row) 	& \textbf{0.815} & \textbf{0.805} & \textbf{0.697} \\
				Real-valued (column) 	& Real-valued (column) 	& 0.849 & 0.823 & 0.721 \\
				\bottomrule
			\end{tabular}
		\end{table*}
		
		\noindent Although nonnegativity can reduce the chance for overfitting, it comes at the cost of worse fitting to the data, as the nonnegativity makes convergence harder. This probably explains the lower predictive performance in this experiment for the nonnegative and semi-nonnegative models. The Bayesian nature of the models already reduces overfitting, potentially making the nonnegativity redundant. However, if the data is very sparse, and hence overfitting is more likely, the nonnegativity could be a great option. \\
		
\begin{figure}[t]
			\centering
			\begin{subfigure}[t]{0.48 \columnwidth}
				\includegraphics[width=1\columnwidth]{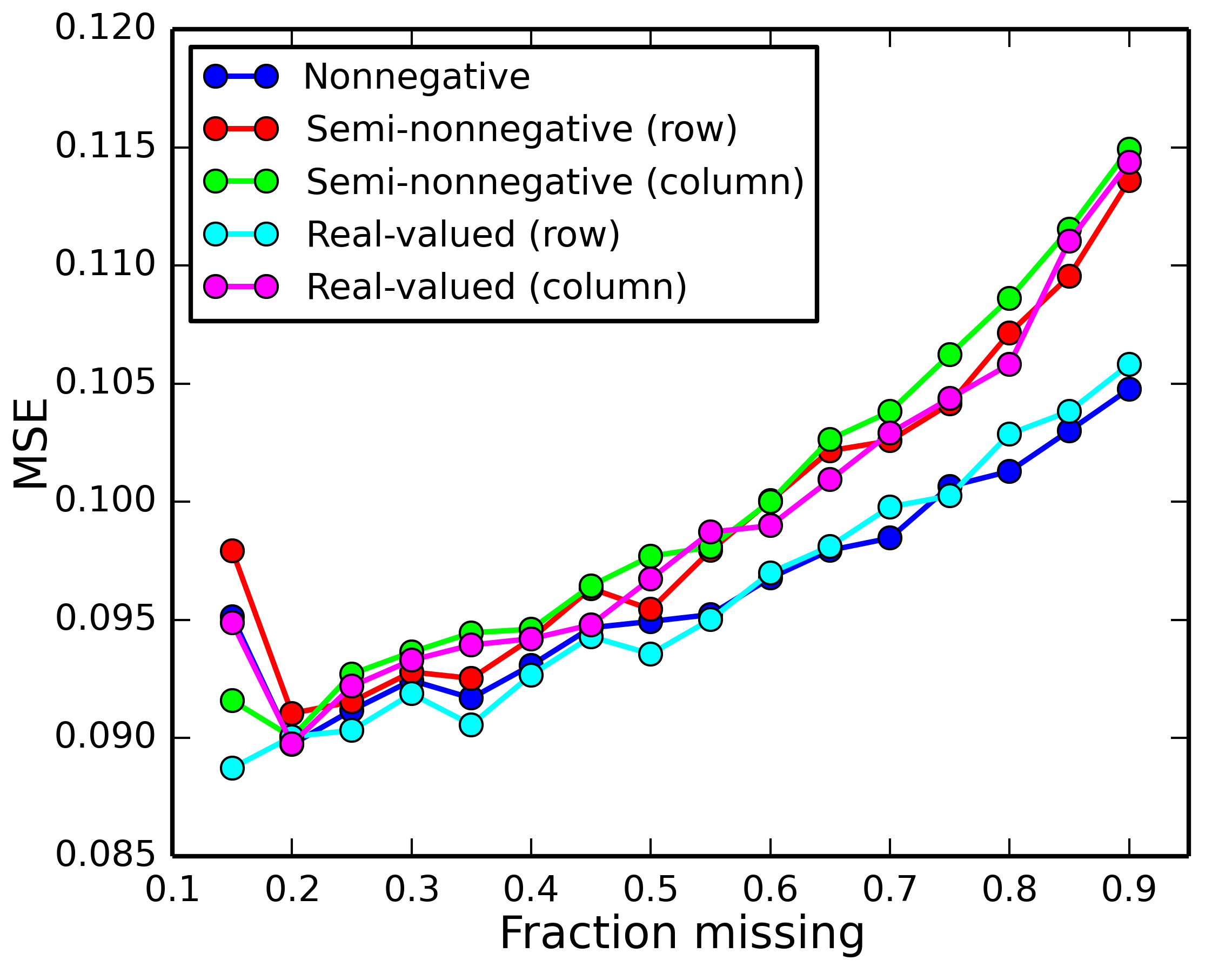}
				\captionsetup{width=0.6\columnwidth}
				\caption{HMF D-MF} 
			\end{subfigure}
			\begin{subfigure}[t]{0.48 \columnwidth}
				\includegraphics[width=1\columnwidth]{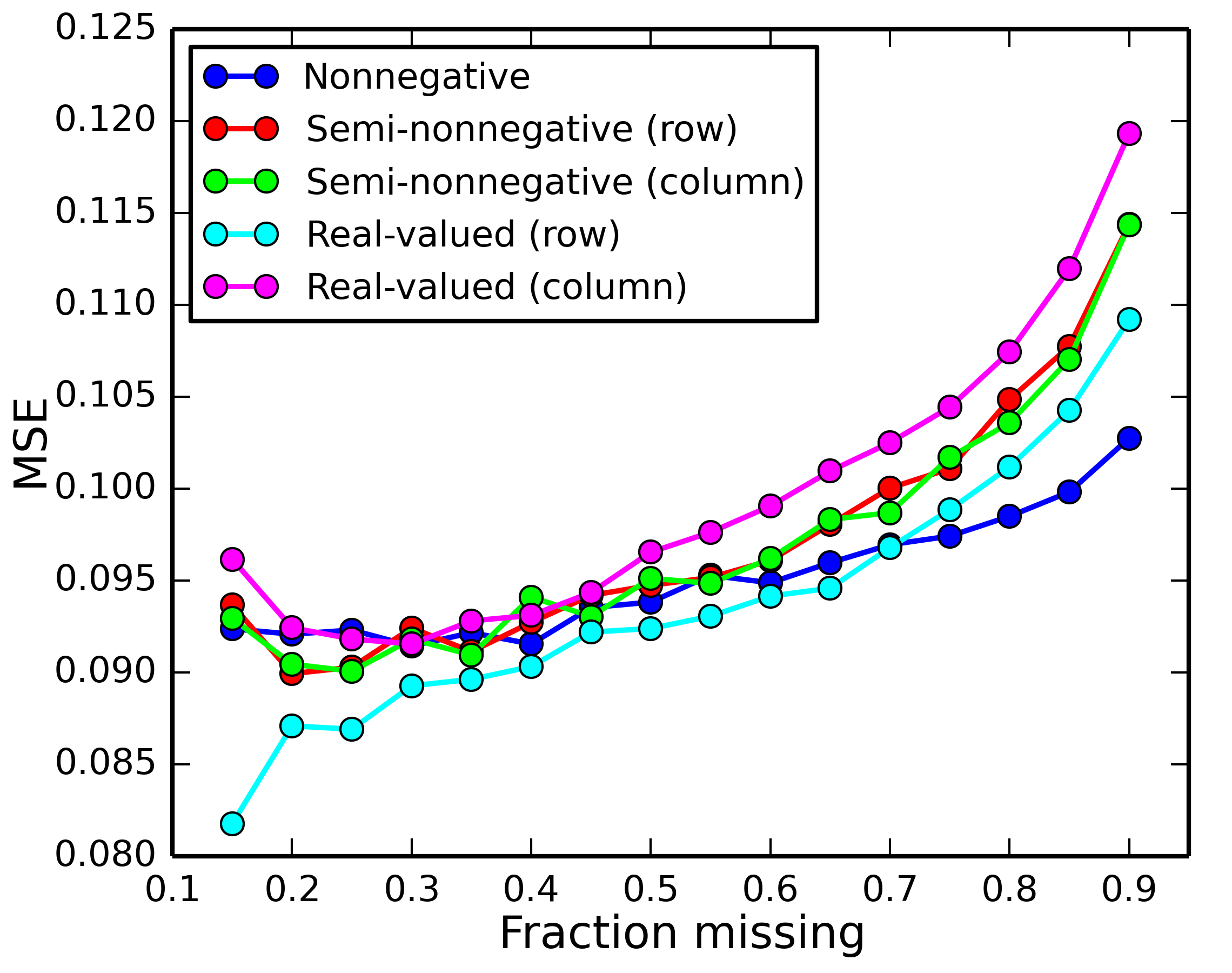}
				\captionsetup{width=0.6\columnwidth}
				\caption{HMF D-MTF} 
			\end{subfigure}
			\captionsetup{width=0.98\columnwidth}
			\caption{Graphs showing the performance for the different negativity options for the HMF D-MF (left) and HMF D-MTF (right) models, as the sparsity of the CTRP dataset increases. We plot the average mean squared error (MSE) across 10 random training and data splits.}
			\label{sparse_negativity}
		\end{figure}

		\noindent To explore the advantages of nonnegativity in sparse settings, we repeat the experiments for sparse data predictions that were performed in the main paper on the CTRP drug sensitivity dataset, comparing the five different versions of HMF D-MF and D-MTF from the previous section. We vary the fraction of observed data, splitting the data randomly into train and test 10 times, and taking the average performance of predictions. This is shown for both methods in Figure \ref{sparse_negativity}. We do see that when the sparsity increases to very high levels like 90\%, the nonnegative model (in dark blue) outperforms the real-valued model with row draws (in light blue). This is especially true for the HMF D-MTF model.
		
		\clearpage
		
		\paragraph{Recommendation} It is generally best use Gaussian priors for all factor matrices, resulting in entirely real-valued models. Row-wise draws most often give better performances than column-wise draws. In the case of very sparse matrices, nonnegative models can reduce overfitting.
		
	\subsection{Factorisation types}
		Finally, we experiment with the choice of factorisationt types. Recall that each dataset can be factorised either using matrix factorisation ($\D_l$) or matrix tri-factorisation ($\R_n$). For the drug sensitivity dataset, there are therefore a number of hybrid factorisation possibilities: using matrix factorisation for all (as used in the main paper; HMF D-MF), using matrix tri-factorisation for all (HMF D-MTF), using matrix factorisation on one dataset and tri-factorisation for the other three, or using matrix factorisation on two datasets and tri-factorisation on two as well. Note that applying matrix tri-factorisation on only one dataset is equivalent to using matrix factorisation on all four (since the second factor matrix is not shared with any other dataset). In this section we explore some of these choices. 
		
		\stoptocentries \subsubsection*{Methylation data} \starttocentries
		We firstly consider the methylation datasets (GE, GM, and PM). We computed the Spearman correlation of values in each pair of these datasets, as given in Table \ref{correlation_methylation}. Here we see that GM and PM are (weakly) positively correlated, and GE and PM are (weakly negatively) correlated. We then performed 10-fold out-of-matrix cross-validation experiments (as before), varying the factorisation types on the three datasets. For simplicity we used $K_t = 10$ and $\alpha = 0.5$ for all datasets, $K$-means initialisation for the shared factor matrices ($\F_t$) and least squares for the private ones ($\G_l, \S_n$). \\
		
		\noindent The results are given in Table \ref{varying_factorisation_methylation}, where the left three columns indicate the factorisation types, and the right three give the predictive performances on each of the datasets (each column corresponds to an experiment where we predict missing rows in that dataset). As can be seen, using multiple matrix factorisation ($\D$ for all matrices) gives the best performance most of the time, which is unsurprising since the three datasets are so weakly correlated. The best performance for GE ($R,R,D$) is most likely due to random variations of performance in the cross-validation procedure (the splitting of data into train and test sets is done randomly each time).
		
		\begin{table*}[t!]
			\captionsetup{width=1\columnwidth}
			\caption{Spearman correlation between values of each of the dataset pairs.}
			\label{correlation_methylation}
			\centering
			\begin{tabular}{c | c c c}
				& GE & GM & PM	\\
				\hline
				GE & - & -0.07 & -0.12 \\
				GM & -0.07 & - & 0.14 \\
				PM & -0.12 & 0.14 & - \\
			\end{tabular}
		\end{table*}
		
		\begin{table*}[t!]
			\captionsetup{width=1\columnwidth}
			\caption{Performances of out-of-matrix cross-validation results on the methylation datasets, where we vary the factorisation types on each of the three datasets. The best performances are highlighted in bold.}
			\label{varying_factorisation_methylation}
			\centering
			\begin{tabular}{c c c c c c}
				\toprule
				\multicolumn{3}{c}{Factorisation type} & \multicolumn{3}{c}{Performance}	\\
				\cmidrule(lr){1-3} \cmidrule(lr){4-6}
				GE & GM & PM & GE & GM & PM \\
				\cmidrule(lr){1-3} \cmidrule(lr){4-6}
				$R$ & $R$ & $R$ & 0.876 & 0.744 & 0.864 \\
				$D$ & $R$ & $R$ & 0.877 & 0.717 & 0.949 \\
				$R$ & $D$ & $R$ & 1.128 & 0.698 & 0.960 \\
				$R$ & $R$ & $D$ & \textbf{0.860} & 0.692 & 0.834 \\
				$D$ & $D$ & $D$ & 0.869 & \textbf{0.663} & \textbf{0.799} \\
				\bottomrule
			\end{tabular}
		\end{table*}
	
		\stoptocentries \subsubsection*{Drug sensitivity data} \starttocentries
			Similarly, we do this experiment for the four drug sensitivity datasets, for in-matrix predictions. These four datasets have much higher correlation (as they are repeated experiments -- the same experiment conducted by different biological labs), as shown in Table \ref{correlation_drug_sensitivity}, giving the Spearman correlation between the overlapping observed entries in each pair of the datasets. You can also find the overlaps between the datasets in Table \ref{Drug sensitivity datasets}, from the main paper. 
			Here we see that:
			\begin{itemize}
				\item CCLE $IC_{50}$ and CCLE $EC_{50}$ are highly correlated, and have a big overlap.
				\item GDSC $IC_{50}$ and CCLE $IC_{50}$ are highly correlated, but few GDSC entries are also in CCLE. In contrast, many CCLE entries are in GDSC.
				\item GDSC $IC_{50}$ and CCLE $EC_{50}$ are (relatively) weakly correlated, but have a very small overlap. Overlap wise the same applies as for CCLE $IC_{50}$.
				\item Very few CTRP $EC_{50}$ entries are in CCLE $IC_{50}$ or $EC_{50}$, but many are in GDSC $IC_{50}$.
			\end{itemize}
			For the experiments we used $K_t = 10$ and $\alpha = 1.0$ for all datasets,  $K$-means initialisation for the shared factor matrices ($\F_t$) and least squares for the private ones ($\G_l, \S_n$, and if we used matrix factorisation we shared the row factors (corresponding to drugs). The results are given in Table \ref{varying_factorisation_drug_sensitivity}. There are a lot of results, so we will consider them one column at a time. 
			
			\begin{table*}[t!]
				\captionsetup{width=1\columnwidth}
				\caption{Spearman correlation between values of each of the dataset pairs.}
				\label{correlation_drug_sensitivity}
				\centering
				\begin{tabular}{c | c c c c}
					& GDSC $IC_{50}$ & CTRP $EC_{50}$ & CCLE $IC_{50}$ & CCLE $EC_{50}$	\\
					\hline
					GDSC $IC_{50}$ & - & 0.47 & 0.59 & 0.39 \\
					CTRP $EC_{50}$ & 0.47 & - & 0.44 & 0.45 \\
					CCLE $IC_{50}$ & 0.59 & 0.44 & - & 0.65 \\
					CCLE $EC_{50}$ & 0.39 & 0.45 & 0.65 & - \\
				\end{tabular}
			\end{table*}
		
			\begin{table*}[t]
				\caption{Overview of the four drug sensitivity dataset after preprocessing and filtering.} 
				\label{Drug sensitivity datasets}
				\centering
				\begin{tabular}{lccccccc}
					\toprule
					& Number & Number & Fraction & \multicolumn{4}{l}{Overlap with other datasets} \\
					\cmidrule(lr){5-8}
					Dataset & cell lines & drugs & observed & GDSC $IC_{50}$ & CTRP $EC_{50}$ & CCLE $IC_{50}$ & CCLE $EC_{50}$ \\
					\midrule
					GDSC $IC_{50}$ 		& 399 		& 48 		& 73.57\%	& - 	& 52.25\%	& 9.34\%	& 6.00\% \\
					CTRP $EC_{50}$		& 379 		& 46 		& 86.03\%	& 57.39\%	& - 	& 11.96\% 	& 7.37\% \\
					CCLE $IC_{50}$	 	& 253 		& 16  		& 96.42\% 	& 44.19\% 	& 51.51\% 	& - 	& 55.06\% \\
					CCLE $EC_{50}$		& 252 		& 16 		& 58.88\% 	& 28.52\% 	& 31.87\% 	& 55.28\% 	& - \\
					\bottomrule
				\end{tabular}
			\end{table*}
		
			\begin{table*}[t!]
				\captionsetup{width=1\columnwidth}
				\caption{Performances of in-matrix cross-validation results on the drug sensitivity datasets, where we vary the factorisation types on each of the four datasets. The best performances are highlighted in bold.}
				\label{varying_factorisation_drug_sensitivity}
				\centering
				\begin{tabular}{c c c c c c c c}
					\toprule
					\multicolumn{4}{c}{Factorisation type} & \multicolumn{4}{c}{Performance}	\\
					\cmidrule(lr){1-4} \cmidrule(lr){5-8}
					GDSC $IC_{50}$ & CTRP $EC_{50}$ & CCLE $IC_{50}$ & CCLE $EC_{50}$ & GDSC $IC_{50}$ & CTRP $EC_{50}$ & CCLE $IC_{50}$ & CCLE $EC_{50}$ \\
					\cmidrule(lr){1-4} \cmidrule(lr){5-8}
					$R$ & $R$ & $R$ & $R$ & 0.0767 & \textbf{0.0889} & \textbf{0.0537} & 0.1071 \\
					$D$ & $R$ & $R$ & $R$ & 0.0765 & 0.0901 & 0.0546 & 0.1098 \\
					$R$ & $D$ & $R$ & $R$ & \textbf{0.0758} & 0.0909 & 0.0543 & 0.1079 \\
					$R$ & $R$ & $D$ & $R$ & 0.0765 & 0.0890 & 0.0584 & \textbf{0.1055} \\
					$R$ & $R$ & $R$ & $D$ & 0.0768 & 0.0893 & \textbf{0.0537} & 0.1078 \\
					$D$ & $D$ & $R$ & $R$ & 0.0763 & 0.0899 & 0.0569 & 0.1079 \\
					$D$ & $R$ & $D$ & $R$ & 0.0766 & 0.0901 & 0.0553 & 0.1090 \\
					$D$ & $R$ & $R$ & $D$ & 0.0769 & 0.0898 & 0.0554 & 0.1064 \\
					$R$ & $D$ & $D$ & $R$ & 0.0765 & 0.0901 & 0.0566 & 0.1060 \\
					$R$ & $D$ & $R$ & $D$ & 0.0772 & 0.0906 & 0.0543 & 0.1082 \\
					$R$ & $R$ & $D$ & $D$ & 0.0771 & 0.0892 & 0.0542 & 0.1076 \\
					$D$ & $D$ & $D$ & $D$ & 0.0776 & 0.0910 & 0.0562 & 0.1064 \\
					\bottomrule
				\end{tabular}
			\end{table*}
		
			\begin{itemize}
				\item \textbf{GDSC $IC_{50}$} (first column) -- multiple matrix tri-factorisation (first row) achieves better performance than multiple matrix factorisation (last row), but a slight improvement can be achieved using a hybrid approach. 
				\item \textbf{CTRP $EC_{50}$} (second column) -- the best performance is achieved when using multiple matrix tri-factorisation on all datasets. Note that the CTRP dataset has a very similar correlation to all three other datasets.
				\item \textbf{CCLE $IC_{50}$} (third column) --  the best performance is achieved when using multiple matrix tri-factorisation on all datasets, or when only CCLE $EC_{50}$ is not decomposed as an $R$ matrix, but using matrix factorisation ($D$) instead.
				\item \textbf{CCLE $EC_{50}$} (last column) -- the best performances are achieved when the CCLE $IC_{50}$ and $EC_{50}$ are not both decomposed as an $R$ matrix, but instead one or either is decomposed as a $D$ matrix. 
			\end{itemize}
			The last two items seem to imply that having a large overlap of entries and high correlation (such as CCLE $IC_{50}$ and $EC_{50}$), does not necessarily mean we should use matrix tri-factorisation on both datasets. The best hybrid combination of factorisations is not so obvious, but by trying out multiple candidates a suitable hybridity can be found.
			
		\paragraph{Recommendation} For dissimilar datasets, with low correlation (like the methylation data), it is best to use multiple matrix factorisation. When the datasets are very similar, with high correlation (like the drug sensitivity data), matrix tri-factorisation can give better results. These two models will generally give very good performance already. One of the hybrid combinations of matrix factorisation and tri-factorisation can sometimes lead to even better results. Nested cross-validation can be used to find the best hybrid combination.

 
\clearpage
\section*{Bibliography} 
\bibliography{bibliography}
\bibliographystyle{abbrvnat}